\documentclass[10pt,letterpaper]{article}
\providecommand{\MateriaFontPath}{fonts/}
\usepackage{amsmath}
\usepackage[letter]{materia}
\usepackage{graphicx,booktabs,array}
\usepackage{framed}
\AtBeginDocument{\colorlet{shadecolor}{MateriaRose!6}}\renewenvironment{shaded}{\MakeFramed{\advance\hsize-\width\FrameRestore}}{\endMakeFramed}%
\usepackage[numbers,sort&compress]{natbib}
\usepackage{authblk}
\usepackage{hyperref}
\graphicspath{{figures/}}

\title{Where the Tokens Go: Heat, Entropy and Error in Agentic Coding}
\author[1,2,3,4,5,*]{Al\'an Aspuru-Guzik}
\author[1,2,3,*]{Varinia Bernales}
\affil[1]{Department of Computer Science, University of Toronto, Toronto, Canada}
\affil[2]{Department of Chemistry, University of Toronto, Toronto, Canada}
\affil[3]{Acceleration Consortium, Toronto, Canada}
\affil[4]{Vector Institute for Artificial Intelligence, Toronto, Canada}
\affil[5]{NVIDIA, Toronto, Canada}
\affil[*]{Corresponding authors: alan@aspuru.com, varinia@bernales.org}
\date{November 2026}
\usepackage{colortbl}%
\usepackage{tikz}%
\DeclareRobustCommand{\status}[1]{\tikz[baseline={([yshift=-0.12ex]s.base)}]\node[circle,draw=MateriaRose,line width=0.45pt,inner sep=0pt,minimum size=1.75ex,text=MateriaRose,font=\fontsize{5.6}{5.6}\selectfont\bfseries](s){\smash{#1}\vphantom{x}};}
\DeclareCaptionFont{dictsize}{\fontsize{7.5}{9.3}\selectfont}\newlength{\dictpad}\newcommand{\dictbeg}{\vspace{\dictpad}}\newcommand{\dictend}{\par\vspace{\dictpad}}
\begin{document}
\justifying\captionsetup{justification=justified,font+=footnotesize}\renewcommand\bibfont{\small\justifying}%
\makeatletter
\twocolumn[\begin{@twocolumnfalse}
\maketitle
\begin{abstract}
Coding agents are billed by the token, yet software engineering says nothing about what a token actually buys.
Here we introduce a theory of software thermodynamics.
The theory draws analogues from classical and stochastic thermodynamics, extends them, and applies them to modern agentic coding.
We count the tokens an agent generates as its cost and divide that cost into work and heat.
The test suite acts as a software Maxwell's daemon that judges every coding change.
We test the theory with an analogue of the experiments in which a single
molecule is pulled open and let fold again: an agent makes a change to a repository (coding) and then undoes it
(uncoding), many times over, and we count the tokens of every model call.
The cycle returns the requested edits to where they began, with each direction started from a stored reference tree, and the spread of its cost over repeated cycles defines a software temperature, which rose with the sampling temperature of the model.
About 80 per cent of the tokens of a successful run are generated in model calls that leave the requested change
where it was, and we call these tokens heat.
The free energy of a change, which we estimate by analogy with the equality of Jarzynski, is 14 to 22 per cent of the mean cost of a successful coding run.
Coding costs 1.1 to 1.5 times as much as uncoding.
The daemon is imperfect: the verdict of a test suite varies with its environment, and a suite that accepts a broken
change has a cost in tokens.
The practical implication of the proposed theory could be new tools for analysing and optimising coding agents
through the lens of thermodynamics.
\end{abstract}
\vspace{0.6em}
\begin{shaded}
\noindent A model proposes changes to a repository, one after another, and every
word it proposes costs tokens; that is the engine, and the tokens are its fuel. The daemon at the gate cannot write code. She runs the tests on each change and opens the door for the changes that pass.
Maxwell's daemon sorted fast molecules from slow ones by looking; ours sorts working changes from broken ones the
same way, and like his, she pays for looking, and sometimes she makes mistakes herself! A change made and then unmade passes
her twice and leaves the requested change where it began, so everything the round trip cost was lost. This
article asks four things of the engine and its daemon: what a round trip costs, what is the least that a change can cost and where the rest goes, how reliably she reads the errors, and what her wrong readings cost.
\end{shaded}
\vspace{0.8em}
\end{@twocolumnfalse}]
\makeatother

\section{Introduction}\label{sec:intro}
In one survey of 2025, 90 per cent of technology professionals reported that they use AI at work~\cite{dora2025report}. In another, about 31 per cent of respondents reported that they use AI agents at work, some of them infrequently~\cite{stackoverflow2025survey}. Therefore, it is important to understand the efficiency of the tools that these professionals employ. A coding agent is a language model that acts on a repository through a harness, the program that gives the model its tools~\cite{yang2024sweagent}. It edits a repository toward a stated goal, and in the setting we study it edits until the change passes a set of tests, which we call the gate. Code that passes the gate is kept. We ask how
many tokens go into the code that is kept and how many are spent elsewhere. Borrowing a word from thermodynamics,
we call the latter \emph{heat}.

\begin{figure*}[!t]\begin{shaded}\noindent\textbf{One change and its bill.} An agent renames a module, updates every reference to it and runs the tests. The run costs about 2{,}300 tokens. About 1{,}900 of them go to reading, searching and testing, which we call heat, and about 400 go to the edits, which we call work. We estimate that the change could cost as little as 350 tokens, its free energy.\footnotemark\ This article is about where the rest goes.\end{shaded}\end{figure*}\setcounter{dbltopnumber}{1}

\begin{figure*}[t]\centering\includegraphics[width=\textwidth]{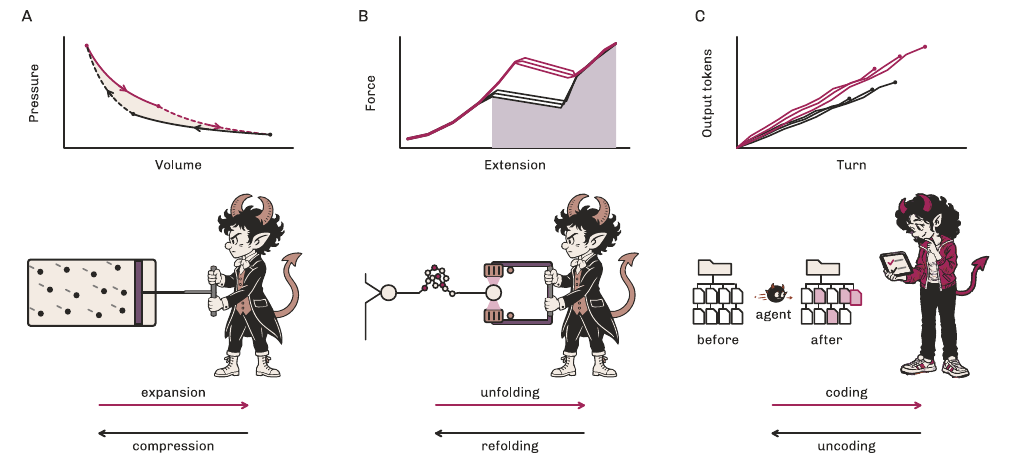}
\caption{The three experiments described in this article, shown schematically. In every panel the forward direction is drawn in rose and the reverse direction in black. (A) Carnot's engine. A Maxwell daemon moves the piston, so that the gas expands and is then compressed. The cycle has two isotherms (solid) and two adiabats (dashed), and the area that they enclose is the work delivered. (B) The pulling of a single molecule. A Maxwell daemon holds optical tweezers and pulls an RNA molecule with their optical trap. Moving the trap away unfolds the molecule, and moving it back lets the molecule refold~\cite{collin2005verification}. The force against the extension differs from one repetition to the next, and the shaded area under the refolding curve is the work that the molecule returns as it refolds. (C) Coding and uncoding (this work). An agent changes a repository (coding, forward) and then undoes the change (uncoding, reverse). The gate daemon stands for the test suite and judges every change. The plot shows the output tokens that three runs in each direction had spent after each turn. Both directions cost tokens.}
\label{fig:three}\end{figure*}

\begin{table*}[p]\centering\fontsize{7.5}{9.3}\selectfont\setlength{\tabcolsep}{5pt}\renewcommand{\arraystretch}{0.8}\setlength{\dictpad}{1.0pt}
\captionsetup{justification=raggedright,singlelinecheck=false,font+=dictsize}\caption{Correspondence between classical thermodynamics and the software thermodynamics proposed in this work, in four groups. The first column names the idea, the second gives its symbol or symbols, the third its meaning in classical thermodynamics, either as a concept or as an example, and the fourth its counterpart in software. The last two columns give the section where the line is defined, set or chosen, and the section where it is measured or estimated. The results test how far the definitions behave like the originals. {\color{MateriaRose}Status} of each software entry: \status{d}~defined; \status{e}~estimated from runs; \status{a}~analogy; \status{h}~untested hypothesis.}
\label{tab:dictionary}
\begin{tabular}{>{\dictbeg\raggedright\arraybackslash}m{0.105\textwidth}<{\dictend}l>{\dictbeg\raggedright\arraybackslash}m{0.20\textwidth}<{\dictend}>{\dictbeg\raggedright\arraybackslash}m{0.335\textwidth}<{\dictend}cll}
\toprule
Idea & Symbol(s) & Classical thermodynamics & Software thermodynamics & Status & Defined & Measured \\
\midrule
\rowcolor{MateriaWash}\multicolumn{7}{l}{\rule{0pt}{2.4ex}\textbf{\color{MateriaRose}The system, work and heat}} \\
System &  & A gas in a cylinder & A \emph{tree}: a code repository at one exact version & \status{d} & \S\ref{sec:defsystem} &  \\
\rowcolor{MateriaWash!65}Process &  & One stroke of the piston & A \emph{run}: an agent edits the tree, call by call, until the tests pass & \status{d} & \S\ref{sec:defrun} &  \\
Reaction coordinate & $q$ & The extension of a pulled molecule & The fraction of the change's blocks that are in place in the tree & \status{d} & \S\ref{sec:defsystem} & \S\ref{sec:m-system} \\
\rowcolor{MateriaWash!65}Cost & $C$ & The fuel burned & The tokens that the model generates in a run, whether or not the run lands & \status{d} & \S\ref{sec:defrun} & \S\ref{sec:cycle} \\
Work & $W$ & The energy that the piston puts in & The tokens of the turns that add target edits to the tree; turns that remove them count negative (Eq.~\ref{eq:work}) & \status{d} & \S\ref{sec:defwork} & \S\ref{sec:free} \\
\rowcolor{MateriaWash!65}Heat & $h$ & The energy that leaves as disorder & The tokens of the turns that leave the target edits unchanged & \status{d} & \S\ref{sec:defwork} & \S\ref{sec:cycle} \\
\rowcolor{MateriaWash!65} & $Q$ &  & The sum of all the tokens of a cycle & \status{d} & \S\ref{sec:defcycle} & \S\ref{sec:free} \\
Energy & $E$ & The energy of one microstate & The minimum average work that builds one exact tree from the empty repository (Eq.~\ref{eq:partition}) & \status{d} & \S\ref{sec:defenergy} &  \\
\rowcolor{MateriaWash!65}Free energy & $F$ & The least work that takes a system from one state to another & The minimum average work that a change can cost (Eqs.~\ref{eq:secondlaw} and~\ref{eq:jarzynski}) & \status{e} & \S\ref{sec:deffree} & \S\ref{sec:free} \\
Dissipation &  & The work beyond the reversible minimum, lost as heat & The mean work above the free energy, $\langle W\rangle - \Delta F$ & \status{e} & \S\ref{sec:deffree} & \S\ref{sec:free} \\
\addlinespace
\rowcolor{MateriaWash}\multicolumn{7}{l}{\rule{0pt}{2.4ex}\textbf{\color{MateriaRose}Temperature, entropy and information}} \\
Software temperature & $1/\beta$ & How hard the molecules jostle, $k_BT$ & The tokens that one nat costs the model (Eqs.~\ref{eq:betacyc} and~\ref{eq:betawork}) & \status{e} & \begin{tabular}[t]{@{}l@{}}\S\ref{sec:defcycle}\\\S\ref{sec:free}\end{tabular} & \begin{tabular}[t]{@{}l@{}}\S\ref{sec:engine}\\\S\ref{sec:free}\end{tabular} \\
\rowcolor{MateriaWash!65}Hinton temperature~\cite{ackley1985learning} & $T_{\rm H}$ & No counterpart: a dial on the apparatus & The sampling setting of the LLM's API (lower samples less, higher samples more) & \status{d} & \begin{tabular}[t]{@{}l@{}}\S\ref{sec:intro}\\\S\ref{sec:methods}\end{tabular} &  \\
Entropy & $S$ & The logarithm of the number of microstates compatible with what one can see & The logarithm of the number of distinct trees that count as done; our count is a lower bound & \status{e} & \S\ref{sec:engine} & \S\ref{sec:engine} \\
\rowcolor{MateriaWash!65}Unit of information &  & The nat and the bit; one bit is $\ln 2$ nats & The same; a verdict of the gate is at most one bit, and a fork with $k$ branches carries at most $\log_2 k$ bits & \status{d} & \begin{tabular}[t]{@{}l@{}}\S\ref{sec:engine}\\\S\ref{sec:methods}\end{tabular} &  \\
Price of a bit &  & $k_BT\ln 2$ of heat to erase it (Landauer, Eq.~\ref{eq:landauer}) & By analogy, $(1/\beta)\ln 2$ tokens to settle one fork between two ways & \status{h} & \S\ref{sec:engine} &  \\
\addlinespace
\rowcolor{MateriaWash}\multicolumn{7}{l}{\rule{0pt}{2.4ex}\textbf{\color{MateriaRose}The cycle and its landscape}} \\
Cycle &  & \emph{Unfolding and refolding}: a molecule is pulled open, then let fold & \emph{Coding and uncoding}: a change is made, then unmade, and its edits return to where they began & \status{d} & \S\ref{sec:defcycle} & \begin{tabular}[t]{@{}l@{}}\S\ref{sec:cycle}\\\S\ref{sec:engine}\end{tabular} \\
\rowcolor{MateriaWash!65}Two tasks &  & Two strokes of the same engine & Task A and Task B: each renames one module, and the program behaves the same before and after & \status{d} & \begin{tabular}[c]{@{}l@{}}\S\ref{sec:cycle}\\\S\ref{sec:methods}\end{tabular} &  \\
Reversibility &  & A process that is slow enough to produce no entropy. When run backwards, it returns its work & Not observed: every cycle cost tokens. Our reference is weaker, a change that costs the same to undo as to make & \status{e} & \S\ref{sec:cycle} &  \\
\rowcolor{MateriaWash!65}Visit potential & $\Phi$ & The potential of mean force, $-k_BT\ln\rho$~\cite{kirkwood1935statistical}; out of equilibrium, the quasi-potential~\cite{freidlin1984random,wang2008potential} & $\Phi = -\ln\rho$, with $\rho$ the density of the repository states that the runs visited; a pure number, and a different quantity from $F$ & \status{e} & \begin{tabular}[t]{@{}l@{}}\S\ref{sec:wells}\\\S\ref{sec:m-map}\end{tabular} &  \\
Basin & $\mathcal{M}$ & The basin of a minimum of the potential energy surface~\cite{stillinger1982hidden}, or a metastable state when the system stays in it for long~\cite{prinz2011markov} & The repository states around one minimum of $\Phi$ & \status{e} & \begin{tabular}[t]{@{}l@{}}\S\ref{sec:wells}\\\S\ref{sec:m-map}\end{tabular} &  \\
\rowcolor{MateriaWash!65}Efficiency &  & The share of the heat drawn that an engine turns into work, bounded by Carnot (Eq.~\ref{eq:carnot}) & Tokens needed over tokens spent, 14 to 22 per cent here & \status{e} & \S\ref{sec:discussion} &  \\
\addlinespace
\rowcolor{MateriaWash}\multicolumn{7}{l}{\rule{0pt}{2.4ex}\textbf{\color{MateriaRose}The gate (the software's test suite)}} \\
Two reservoirs &  & A hot and a cold reservoir, with an engine between them & The model, sampled through its harness, as the hot reservoir and the gate as the cold one & \status{a} & \S\ref{sec:intro} &  \\
\rowcolor{MateriaWash!65}Clausius's version of the second law &  & Heat does not pass by itself from a colder body to a hotter one & A gate colder than the agent takes disorder out of the repository, and a gate hotter than the agent lets disorder in & \status{a} & \S\ref{sec:intro} & \S\ref{sec:mistake} \\
Measurement device &  & Maxwell's daemon, reading a molecule & The gate, reading a tree & \status{d} & \S\ref{sec:defgate} & \S\ref{sec:gate} \\
\rowcolor{MateriaWash!65}Measurement error &  & The daemon misreads & A flaky test serves a wrong verdict (Eq.~\ref{eq:pfalse}) & \status{e} &  & \S\ref{sec:mistake} \\
\bottomrule
\end{tabular}
\end{table*}

\begin{table*}[!t]\centering\footnotesize\setlength{\tabcolsep}{10pt}\renewcommand{\arraystretch}{1.08}
\captionsetup{justification=raggedright,singlelinecheck=false}\caption{Three experiments: Carnot's cycle, the unfolding and refolding of a single molecule, and the coding and uncoding of a single repository, which is the experiment of this work. Our experiment is of the second kind. It has one bath, its cycle delivers no work, and what it yields is the free energy of a change, read from the work of many runs.}\label{tab:experiments}
\begin{tabular}{@{}>{\raggedright\arraybackslash}p{0.105\textwidth}>{\raggedright\arraybackslash}p{0.215\textwidth}>{\raggedright\arraybackslash}p{0.255\textwidth}>{\raggedright\arraybackslash}p{0.30\textwidth}@{}}
\toprule
 & Carnot's cycle & Unfolding and refolding a single molecule~\cite{liphardt2002equilibrium,collin2005verification} & Coding and uncoding a single repository (this work) \\
\midrule
The system & A gas in a cylinder & A single RNA molecule held in optical tweezers & A code repository edited by a coding agent \\
\rowcolor{MateriaWash!65}The baths & Two, one hot and one cold & One, the solution around the molecule & One, the model sampled through its harness, at the software temperature $1/\beta$; the gate is read as a second, colder one (\S\ref{sec:intro}) \\
What is driven & The piston & The distance between the two ends of the molecule & The instruction given to the agent: make the change, or undo it \\
\rowcolor{MateriaWash!65}The cycle & The gas expands at the hot bath and is compressed at the cold one, back to its start & The molecule is pulled open (unfolding, forward) and then let fold again (refolding, reverse) & The agent makes a change (coding, forward) and then undoes it (uncoding, reverse) (\S\ref{sec:defcycle}) \\
What the cycle delivers & Work, taken from the heat of the hot bath & Nothing: opening costs more work than folding returns, and the difference is lost as heat & Nothing: every token of the cycle is heat (\S\ref{sec:cycle}) \\
\rowcolor{MateriaWash!65}What is recorded & The heat drawn and the work delivered & The force against the extension, and from it the work of each pull & The tokens of each turn and the progress toward the goal, and from them the work of each run (Eq.~\ref{eq:work}, \S\ref{sec:defwork}) \\
Repetitions & One slow cycle & Many pulls; the work differs from pull to pull & Many runs; the work differs from run to run \\
\rowcolor{MateriaWash!65}The relation used & Carnot's bound on the efficiency (Eq.~\ref{eq:carnot}, \S\ref{sec:intro}) & Jarzynski's equality and Crooks' relation~\cite{jarzynski1997nonequilibrium,crooks1999entropy} & Jarzynski's equality, by analogy and as an estimator (Eq.~\ref{eq:jarzynski}, \S\ref{sec:free}) \\
What it yields & The largest share of the heat that can become work & The free energy of folding & The free energy of the change, in tokens, and the work dissipated above it (\S\ref{sec:free}) \\
\rowcolor{MateriaWash!65}The instrument & Thermometers and gauges & The tweezers, which read force and extension & The gate, which reads the tree; we measure its noise and its errors (\S\ref{sec:gate}, \S\ref{sec:mistake}) \\
\bottomrule
\end{tabular}
\end{table*}

Software engineering measures correctness, coverage and velocity, but it has no theory, and no instrument, for the value or the efficiency of a token in agentic coding. Thermodynamics is the science that divides the energy a machine uses into work and heat. We apply it to the tokens that a coding agent generates and call the resulting theory software thermodynamics.

Figure~\ref{fig:three} shows the three experiments we compare: a Carnot engine (A), the pulling of a single molecule (B), and coding and uncoding (C).

\footnotetext{The numbers of the box are means over the landed coding runs of Task A at a Hinton temperature of 0.5 (\S\ref{sec:m-free}): 2{,}305 tokens, of which 1{,}890 are heat and 415 are work, and an estimated free energy of 351 tokens.}The best-known machine of thermodynamics is the Carnot engine (Fig.~\ref{fig:three}A). It works between two reservoirs, one hot and one cold. In each cycle it draws heat from the hot reservoir, converts part of that heat into work and gives the rest to the cold one. The two temperatures set the largest efficiency $\eta$ that the engine can reach, the share of the heat that becomes work,
\begin{equation}
\eta \le 1 - \frac{T_{\rm cold}}{T_{\rm hot}} .
\label{eq:carnot}
\end{equation}

In this work, we employ the same idea that Carnot employed: a cycle that goes from one state to another and back. Here the two states are the repository before a change and after it: the change is made (coding, the forward direction) and then unmade (uncoding, the reverse direction). An analogy with Carnot's engine would be appealing, but it would be wrong: in software, coding and uncoding are run at a single temperature, that of the model, and therefore the cycle delivers no work.

\setcounter{dbltopnumber}{2}Luckily, there are classical experiments on molecules in which the pulling and refolding of a single molecule is a much more direct analogue of the coding cycle that we study (Fig.~\ref{fig:three}B). Liphardt, Collin and their co-workers held single RNA molecules in optical tweezers, pulled each open and let it fold again, many times, in a bath at one temperature~\cite{liphardt2002equilibrium,collin2005verification}. Opening the molecule cost more work than folding it returned, and the work differed from pull to pull. The least work that takes the molecule from one state to the other is its free energy. From that spread Liphardt and co-workers recovered the free energy of the stretching with Jarzynski's equality~\cite{jarzynski1997nonequilibrium}, and Collin and co-workers recovered free energies of folding with Crooks' relation~\cite{crooks1999entropy}.

We carry out an analogue of the unfolding and refolding experiment, but with code (Fig.~\ref{fig:three}C). A change to a code repository (coding) takes the place of the unfolding, and undoing that change (uncoding) takes the place of the refolding. Tokens take the place of energy, and the spread of the work over many coding and uncoding runs gives an estimate of the free energy of the change. We call this experiment \emph{coding and uncoding}.

Our comparison asks the reader to hold two vocabularies at once. Hoffmann faced the same task when he explained the language of solid-state physics to chemists, and he solved it by drawing parallels to the language of molecular orbitals and bonds~\cite{hoffmann1987chemistry}. Inspired by his comparison, we set out ours in two tables. Table~\ref{tab:dictionary} translates the terms of classical thermodynamics into those of software thermodynamics. Table~\ref{tab:experiments} places Carnot's cycle, the unfolding and refolding of a molecule and our coding and uncoding side by side.

Even though the agent does not run a Carnot cycle, we keep one more idea of Carnot's: his two reservoirs. The similarity is that the gate is at a different temperature from the agent. The model, sampled through its harness, is the hot side. The test suite at the gate is the cold side: a perfect gate would catch every faulty change and would sit at zero temperature. We therefore read the model as the hot reservoir and the gate as the cold one, and we call this the \emph{two-reservoir reading}. Two reservoirs bring with them Clausius's version of the second law: heat does not pass by itself from a colder body to a hotter one~\cite{clausius1854,clausius1856}. A refrigerator obeys this law by spending work to carry heat out of what it holds. In our reading the gate plays the refrigerator. It takes out of the repository the disorder that the agent puts in, and it can do so only while it is the colder of the two. Tests are never perfect in software engineering, so a real gate is not at zero. We make the gate err at a known rate, which gives it a temperature we can read.

So far, we have spoken loosely of temperature, and this is the place to define what we mean by it in this article. We employ two different temperatures. The one that we measure we call the software temperature. As in thermodynamics, it is defined through the constant $\beta$ of the Boltzmann weight: it is $1/\beta$, in units of tokens per nat. A nat is a unit of information, the analogue of the bit when logarithms are natural: one bit is $\ln 2 = 0.693$ nats. The software temperature is therefore the number of tokens that makes an outcome $e$ times rarer.

The second temperature that we employ is a setting and has no thermodynamic meaning of its own. It is the sampling temperature of the model's API, a dimensionless parameter. Its origins hark back to the work of Hinton and co-workers, whose Boltzmann machine carries a temperature of this kind~\cite{ackley1985learning}. Hinton shared the Nobel Prize in Physics in 2024 for this line of work. Therefore, we call the sampling temperature the Hinton temperature, $T_{\rm H}$.

The two temperatures interact as follows. We use $T_{\rm H}$ as a setting that moves the model from cold, where sampling is more repetitive and explores less, to hot, where it explores more. Language models are usually run at a Hinton temperature $T_{\rm H}$ close to 1.0. Many of our experiments therefore measure the software temperature, or quantities derived from it, as a function of the Hinton temperature, which, we recall, is only a setting. \textbf{The software temperature is a measurement of the model, the task and the harness together, and nothing forces it to agree with the Hinton temperature.}

The units of the software temperature are important, because they say how many tokens are spent for a given amount of information. As we describe in \S\ref{sec:defcycle}, these units let us read thermodynamic quantities as probabilities and tokens as information.

With the two temperatures defined, we can name the remaining parts of the picture. The gate plays the part of Maxwell's daemon (see the box): it sorts the proposed changes and keeps those that pass. The tokens spent are the cost, which
divides into work and heat, and the code tree that results is the thermodynamic state. We aim for a theory that is both predictive and useful for understanding and optimising
coding processes.

We first describe the experiment and define \textbf{\emph{coding}} and \textbf{\emph{uncoding}}. We gave a coding agent a change to make in a repository and counted the tokens of every model call, which we call a \textbf{\emph{turn}}, until the tests passed. We then gave it the changed repository and asked it to undo the change. We call the first direction coding (forward) and the second uncoding (reverse). We repeated both directions many times, for two changes and at several settings of the Hinton temperature $T_{\rm H}$. We call one change at one setting a \textbf{\emph{condition}}. A run \textbf{\emph{lands}} when it reaches its goal and the tests pass there.

To track the progress of a run, we record the fraction of the requested change that is in place. It is zero before the change and one when the change is complete. The tokens of a turn that advances this fraction are \textbf{\emph{work}}. The tokens that are spent without changing it are \textbf{\emph{heat}}. For example, heat is spent reading a file, searching, running the tests or editing lines that do not advance the fraction.

Table~\ref{tab:dictionary} is a translation guide: it gives the correspondences used throughout the article in plain language. The equations, their derivations and the experimental controls are in \S\ref{sec:methods}.

Now that we have the background of the theory and an overview of the experiment, we are ready to discuss four distinct measurements that quantify the predictions of our theory. We describe each of them below, each with a short preview of its results.

\subsubsection*{1. The round trip: a hairpin experiment in code}
We measured the cycle: a coding run followed by an uncoding run, which returns the requested edits to where they started. It is the counterpart of the pulling experiment of Table~\ref{tab:experiments}, in which a single molecule is unfolded and refolded (Fig.~\ref{fig:cycle}). The cycle delivers no work, so every token of it is lost. The spread of the cycle's cost over repetitions defines the software temperature.

\subsubsection*{2. Where the tokens go: work, heat and free energy}
In the second experiment, we track how a run spends its tokens. We sort the tokens of each run into work and heat, as defined above: work advances the change and heat does not.

Two runs of the same change never spend the same work, because some runs take a direct route and others wander. The change, however, is the same in every run, and therefore there is a least amount of work that it requires on average. We call that least amount the \textbf{\emph{free energy}} of the change. The work that is spent above that minimal work is lost to dissipation.

No single run gives us the free energy, because we cannot know whether that run took the most direct route. The same problem arises when a single molecule is pulled open: every pull costs a different amount of work. Jarzynski showed that the free energy can still be obtained from many pulls, with an average that gives the most weight to the cheapest ones~\cite{jarzynski1997nonequilibrium}.

Arguably, Jarzynski's result is one of the most consequential recent results of modern stochastic thermodynamics, because nonequilibrium trajectories can still reproduce equilibrium information in a practical and elegant way. In this experiment, we apply Jarzynski's approach to estimate the free energy of our coding changes, in analogy to the classical stochastic thermodynamics approach.

Armed with this toolkit, we tested whether uncoding is the true reverse of coding. The refolding of a molecule is the true reverse of its unfolding: the same path, travelled in the opposite direction. For such a pair, Crooks showed that the free energy and the temperature predict how much more one direction costs than the other~\cite{crooks1999entropy}. We asked whether the same prediction holds for coding and uncoding.

\subsubsection*{3. The daemon's eyesight: the noise of the gate}
In the third experiment, we turn from the agent to the test suite. The test suite is the daemon of our picture: it reads each change and decides whether the change lands.

Maxwell imagined a daemon with perfect eyesight. A real daemon is made of matter, and therefore it has a temperature of its own and it trembles. Feynman's ratchet is the classical example: when the pawl is as warm as the gas that it sorts, it opens at random and sorts nothing~\cite{parrondo1996criticism}. Sagawa and Ueda made this quantitative: a daemon that sometimes misreads gains less than one bit from each reading, and the work that it can extract falls accordingly~\cite{sagawa2010generalized}.

A gate (a test suite) is a daemon at a non-zero temperature. Its thermal noise is the tests that fail from time to time for reasons that have nothing to do with the change, such as timing or the state of the machine. In this experiment, we ask the thermodynamic question: is the signal of a defect larger than the noise of the daemon that must read it?

We measured the noise by running the same tests several times on the same unchanged code and counting the tests that failed. We measured the signal by planting known defects in the code and counting how many of them the suite caught. Because of its noise, a test suite is given a pass mark: a change lands when no more than a set number of tests fail. We set the pass mark in two ways: from a single run on unchanged code, as it is set in practice, and from the noise measured over many runs. We also compared the whole suite with a smaller gate, the tests of the changed file alone.

\subsubsection*{4. When the daemon errs: a gate with a temperature}
In the fourth experiment, we give the noise of the daemon a temperature. In the third experiment, we observed the noise that a test suite has by itself. Here we controlled it: we made the gate give the wrong verdict at a rate that we set, and we counted the broken changes that it let through.

The thermodynamic idea is Boltzmann's. A system at temperature $T$ crosses a barrier of height $\Delta$ with a probability proportional to $e^{-\Delta/T}$. A cold system almost never crosses, and a warm one crosses often, because its thermal noise carries it over. For a gate, the barrier is the repair: a broken change should land only after the agent has spent the tokens that fix it, about a thousand in our runs. When the gate errs, a broken change lands with no repair. The noise of the gate has carried it over the barrier, as thermal noise carries a particle.

We therefore define the temperature of the gate from two quantities that we measure: the share of landed changes that are broken, and the cost of the repair in tokens. A gate that never errs has zero temperature, and a gate that errs more often is hotter. This temperature has the same unit as the software temperature of the model, tokens a nat, and therefore the daemon and the engine can be compared on one scale. Clausius's version of the second law then says what to expect: the gate removes disorder only while it is colder than the model.

We did this on two systems: a toy in which most attempts were broken, and our own repository with a one-line defect planted, in which about half of the attempts were broken.

\paragraph*{Preview of the results.} Coding cost more than uncoding, and the software temperature rose with the Hinton temperature. About 80 per cent of the tokens of a landed run were heat, and the free energy was 14 to 22 per cent of the mean cost of a landed coding run. The whole test suite caught no more planted defects than the tests of the changed file alone. The more often the gate was wrong, the more broken changes landed.

\subsubsection*{Where this work comes from}
This work draws on four earlier lines of work. The first is stochastic thermodynamics. It is the thermodynamics of small systems, in which work, heat and entropy are defined for a single trajectory and differ from one trajectory to the next~\cite{sekimoto1998langevin,seifert2012stochastic}. Its two central results are the ones that we use in the second experiment. Jarzynski's equality gives the free energy of a change from the work of many trajectories, and Crooks's relation compares the work of a process with the work of its reverse~\cite{jarzynski1997nonequilibrium,crooks1999entropy}. 

Both results are theorems for a small system in contact with a heat bath. They assume that every trajectory starts from thermal equilibrium, that the bath sets the temperature, and that the system keeps no memory of its path. It would be convenient if a coding agent met these conditions, but it does not: a run does not start from equilibrium, no bath sets its temperature, and the agent remembers every turn that it has taken. The cost of our runs also has a long tail, because a few heavy runs cost far more than the rest. Therefore, we cannot take the two results as theorems for our setting, and we use them by analogy.

The analogy also gives us our definition of temperature. When the work of a process follows a Gaussian distribution, the two results give the temperature as the variance of the work divided by twice its mean. We take this expression as the definition of the software temperature, and we let the data say whether it behaves like a temperature in our more complex situation. How to define a temperature for work distributions with a long tail, such as ours, is a research question that we leave for future work.

The second line of work we build on concerns the price of a measurement. Maxwell imagined a small being, later called a daemon, who watches
the molecules of a gas and opens a door only for the fast ones. The daemon sorts hot from cold without doing any work, and the second law forbids this~\cite{maxwell1871theory,canales2020bedeviled}. Szilard reduced the puzzle to its smallest case, a single molecule in a box at
temperature $T$. Knowing which half of the box the molecule is in is one bit of information, and it lets the daemon draw
\begin{equation}
W_{\rm bit} = k_BT\ln 2
\end{equation}
of work by letting the molecule push a piston. Szilard concluded that the measurement must itself cost at least that much, or the second law fails~\cite{szilard1929entropieverminderung}. Landauer then focused on the cost of forgetting. Erasing one bit of memory, whatever the physical device, dissipates at least
\begin{equation}
Q_{\rm erase} \ge k_BT\ln 2
\label{eq:landauer}
\end{equation}
of heat~\cite{landauer1961}. Bennett closed the argument. A measurement can in principle be made without dissipation, but the daemon must clear its memory before it can act again, and it pays for that erasure. Therefore, over a full cycle
\begin{equation}
W_{\rm gained} - Q_{\rm erase} \le 0 ,
\end{equation}
and the daemon breaks even at best~\cite{bennett1982}.

Two later results concern daemons that are closer to real devices. Sagawa and Ueda, and Horowitz and
Parrondo, extended the bound to a daemon whose measurement sometimes errs. The work that such a daemon can draw falls with its error, and it is set by the mutual information that the measurement still carries~\cite{sagawa2010generalized,horowitz2011thermodynamic}.
Parrondo and Espa\~nol showed that Feynman's ratchet, a mechanical daemon, never reaches the bound, because it
touches both baths at once~\cite{parrondo1996criticism}. Parrondo, Horowitz and Sagawa review the modern accounting, which treats information as a physical resource and memories as information reservoirs on an equal footing with thermal and chemical baths~\cite{parrondo2015thermodynamics}.

A gate
of tests is a daemon of this kind. It reads a verdict, green or red, and acts on it: a green change lands and a red one does not. A verdict carries at most one bit, and it carries the whole bit only when green and red are equally likely before the gate is read. When the gate's verdict is sometimes wrong, we price it as Sagawa and Ueda price an erring daemon's measurement, by the mutual information between the true state of the change and the verdict that the agent receives. For a verdict that is wrong with probability $P$, when half of the attempts are truly red, this information is $1-h(P)$ bits, where $h(P) = -P\log_2 P - (1-P)\log_2(1-P)$ is the binary entropy. The expression for any share of red attempts is in \S\ref{sec:m-daemon} (Eq.~\ref{eq:info}). The price we put on a verdict is the price of its use, in tokens. We make no claim about a physical cost of measuring or of erasing in the machine that runs the tests.

The third line of work we build on measures free energies on single molecules that are driven far from equilibrium. Liphardt and co-workers pulled one RNA molecule open again and again and tested Jarzynski's equality on the work of the pulls~\cite{liphardt2002equilibrium}. Collin and co-workers then pulled and released single RNA molecules~\cite{collin2005verification}. They held one RNA hairpin between two beads in optical tweezers, moved the beads apart until the hairpin opened, and brought them back until it folded, hundreds of times. Figure~\ref{fig:cycle}a shows five such pulls and five releases. The work of a pull is the area under its curve of force against extension. Unfolding cost more work than refolding returned, and both amounts changed from one repetition to the next. Crooks's fluctuation theorem says that the distributions of the two works cross at the free energy of the change, whatever the pulling speed~\cite{crooks1999entropy}. In the experiment of Collin and co-workers, the two distributions crossed at the same free energy at each of three pulling speeds.

Our coding experiment follows the same design. Each run corresponds to one pull of the molecule: the agent receives the same instruction on each repetition and does a different amount of work. Figure~\ref{fig:cycle}d shows the two work distributions, of coding and of uncoding.

Two refinements of these experiments matter for our estimates, because we have tens of runs in each condition and some of them never land. First, the exponential average over a small number of pulls overestimates the free energy, by an amount that falls as pulls are added~\cite{gore2003bias}. Second, when only a part of the pulls ends in the state of interest, the estimate carries a term that depends on the size of that part~\cite{alemany2012experimental}. A related group of results, the thermodynamic uncertainty relations, ties the regularity of a driven process to the dissipation that pays for it, first for currents and then for the time to reach a target~\cite{barato2015thermodynamic,gingrich2017fundamental,pal2021thermodynamic}.

The fourth line of work we build on is the evaluation of coding agents. These studies count the tasks that an agent resolves on fixed test suites and report the token cost~\cite{yang2024sweagent}. Entropy has been defined for software before, in two ways. The first is the Shannon entropy of how the changes of a period are spread over the files~\cite{hassan2009predicting}. The second is the logarithm of the number of programs that pass a given set of tests, which is estimated in practice from the mutants of one program that survive the tests~\cite{fotinos2026software}. To our knowledge, no study has defined an energy for a state of the code, or measured a temperature, a work or a fluctuation relation for software. In this work, each analogy is a definition tied to a quantity that we measure, and the measurements show how far the quantity behaves like its thermodynamic original.

\section{Results}
\label{sec:results}

\subsection{Overview and definitions}
Before the measurements, we define the quantities of Table~\ref{tab:dictionary} in terms of what we recorded. We measured model calls, file edits and test outcomes from the agent's \emph{records}. We describe how we computed each quantity in \S\ref{sec:methods}.

\subsubsection{The system and its state}\label{sec:defsystem}
The system is a code repository, and its files at one exact version are a \emph{tree}. A \emph{task} is a goal
stated in words together with two trees: the base tree, where the change has not been made, and the target tree,
where it has. We describe the change by a fixed list of $N$ \emph{blocks}. Each block is one piece of the difference between the two trees: the renaming of a file, or a contiguous group of changed lines. A block is \emph{in place} in a tree when the tree already has that change: the file carries its new name, or the changed lines are there. The \emph{reaction coordinate} of a tree is the fraction of the blocks that are in place,
\begin{equation}
q = \frac{n}{N},
\label{eq:q}
\end{equation}
where $n$ is the number of blocks in place. It is zero at the base
tree and one when every block is in place, whatever else the tree contains. The reaction coordinate plays the part of the extension of the molecule in a pulling experiment~\cite{liphardt2002equilibrium}: it says how far the change has been made, as the extension says how far the molecule has been pulled open. We call the value of $q$ the \emph{state} of the tree. Many trees share one state, as many configurations of a molecule share one extension. The definitions in this section describe each tree by $q$ alone, except the energy, which belongs to one exact tree (\S\ref{sec:defenergy}).

The \emph{agent} is a language model that edits the tree through its harness, one model call at a time. One call is a \emph{turn}. The \emph{gate} is the repository's test suite, run on the agent's edits. It answers pass or fail, which we call \emph{green} and \emph{red}, and its
purpose is to stop a change that breaks the code from being accepted.

\subsubsection{A run and its cost}\label{sec:defrun}
A \emph{run} is what the agent does to the system: one attempt at a task in one direction. In a \emph{coding} run (the forward direction), the agent starts from the base tree with instructions to make the change. In an \emph{uncoding} run (the reverse direction), the agent starts from the target tree with instructions to undo the change. A run \emph{lands}
when it reaches its goal and the gate passes there. A run that ends in any other way, because the agent stops short of
the goal or reaches the limit on turns, is \emph{unfinished}. The \emph{cost} $C$ of a run is the sum of the tokens that the model generated across its turns. A cost is never negative, and it is paid whether or not the run lands.

\subsubsection{Work and heat}\label{sec:defwork}
We divide the cost of a run into two parts, according to whether each turn changes the state $q$. The tokens of
a turn that leaves $q$ unchanged are \emph{heat}. In such a turn, the agent read a file, searched, ran the tests or edited outside the
goal's blocks, and the state did not move. The tokens of a turn that changes $q$ are \emph{work}, and they carry the
sign of the step, positive when $q$ rises and negative when it falls. With $c_i$ the tokens of turn $i$ and $q_i$
the state after it, the work and the heat of a run are
\begin{equation}
W = \sum_{i} c_i\,\mathrm{sgn}(q_i - q_{i-1}), \qquad h = \sum_{i\,:\,q_i = q_{i-1}} c_i .
\label{eq:work}
\end{equation}
The work is zero for a run that never touches the goal's blocks, and it does not depend on how long the agent read
before it wrote. A run that never steps back has $C = |W| + h$, in either direction. No landed run in this study stepped back.

An uncoding run takes $q$ from one to zero, so
its work is negative. The sign records the direction of the step. No token is returned, and an uncoding run pays for every token it generates. We keep the sign because it lets us compare the two directions as one compares
the stretching of a molecule with its refolding: the work put in on the way out against the work given back on the
way in~\cite{liphardt2002equilibrium,collin2005verification}.

\subsubsection{The cycle}\label{sec:defcycle}
A coding run followed by an uncoding run brings the state back to where it began, and we call the pair a
\emph{cycle}. Nothing is stored in the state at the end of a cycle, so we call its whole cost the cycle's
\emph{heat}, $Q$. It is the mean cost of a landed coding run plus that of a landed uncoding run. The cycle's \emph{asymmetry} $\mathcal{A}$ is the mean cost of a landed coding run minus the mean cost of a landed uncoding run. We measured the two directions separately. Every coding run started from the same stored base tree and every uncoding run from the same stored target tree, as the unfolding and the refolding of a molecule are measured in separate pulls from prepared states~\cite{collin2005verification}. What returns to its start is the state $q$, which goes from zero to one and back to zero. A landed tree may differ from its start in lines outside the goal's blocks (\S\ref{sec:m-cost}).

We define the \emph{software temperature of the cycle}, $1/\beta_{\rm cyc}$, from the mean and variance of the cost around the
cycle, by the relation that Gaussian work obeys under the fluctuation theorems (\S\ref{sec:m-cost}, Eq.~\ref{eq:betacyc}). This relation
is a definition and involves no fit. The cycle has this one temperature. It carries no heat from a hotter reservoir to a colder one and delivers no work. It is therefore a \emph{hysteresis loop}, like that of a molecule pulled open and folded again~\cite{collin2005verification}, and its heat $Q$ is all lost.

We also measure the \emph{arrow of time} of a run. A process that produces entropy can be told from its own record played backwards, and Prigogine placed this irreversibility at the centre of the thermodynamics of systems far from equilibrium~\cite{prigogine1978time}. Let $x$ be the record of a run, its tokens per turn in order, and $\tilde{x}$ the same record read backwards. With $P(x)$ the probability of a record among the runs of one condition, the arrow of time is the relative entropy between the records and their reverses,
\begin{equation}
D = \sum_{x} P(x)\,\ln\frac{P(x)}{P(\tilde{x})} .
\label{eq:arrow}
\end{equation}
$D$ is zero when a record and its reverse are equally likely. In thermodynamics, $D$ is a lower bound on the mean entropy that a process produces, in nats~\cite{parrondo2009entropy,roldan2012kl}. We estimate $D$ from how well a classifier tells the recorded runs from the same runs reversed (\S\ref{sec:m-arrow}).

\subsubsection{The free energy of a change}\label{sec:deffree}
The \emph{free energy difference} $\Delta F$ of a change is the minimum average work that takes the state from $q = 0$ to
$q = 1$, in tokens. It is zero for a change with no blocks, and it belongs to a task for a given model and harness. An
agent that writes the same blocks in fewer tokens assigns the change a lower free energy. No single run shows it, because runs of one task do different amounts of work. The minimum is taken over ways of making the change and the average over runs, so a single run can do less work than $\Delta F$, as a single pull of a molecule can. In thermodynamics, the mean work of a process is never less
than the free energy difference it crosses,
\begin{equation}
\langle W\rangle \ge \Delta F ,
\label{eq:secondlaw}
\end{equation}
and the excess $\langle W\rangle - \Delta F$ is the \emph{dissipated work}. In \S\ref{sec:free}, we estimate $\Delta F$
from the spread of the work over many runs in both directions, and we compare the result with the ratio of the landing rates of the two directions.

\subsubsection{The energy of a tree}\label{sec:defenergy}
The free energy of the last subsection belongs to a goal, and many trees meet one goal. We also define a quantity that belongs to one exact tree. The \emph{energy} $E$ of a tree is the minimum average work that builds exactly that tree from the empty repository, in tokens, for a given model and harness. The empty repository has zero energy. The energy of a change from one exact tree to another is the difference of their energies, $\Delta E$. In statistical mechanics, the free energy of a set of microstates follows from their energies,
\begin{equation}
F = -\frac{1}{\beta}\ln\sum_i e^{-\beta E_i} ,
\label{eq:partition}
\end{equation}
and we take the same relation for the trees $i$ that meet a goal. When $N$ trees of equal energy meet the goal, the relation gives $\Delta F = \Delta E - (1/\beta)\ln N$: the more trees meet the goal, the less work the goal costs. We define the energy in this work and we do not measure it. The mean work of runs that are asked for one exact tree bounds its energy from above (Eq.~\ref{eq:secondlaw}), and we leave that experiment for future work.

\subsubsection{The gate as an instrument}\label{sec:defgate}
The quantities defined so far depend on the gate, because a run lands only when the gate passes. The gate is therefore our instrument, and we study it in two ways.

First, the gate is a \emph{thermometer} with noise of its own. On a clean tree, which is a tree with no defect, the test suite still fails some tests, and the number of failures $N$ changes from one run of the suite to the next. The gate answers red when more than a set number of tests fail, and we call this number the \emph{threshold}. A threshold is therefore a quantile of the distribution of $N$: it fixes how often a clean tree is called red. A tree with a defect can still pass the gate. The share of landed trees that carry a defect follows from how often this happens (\S\ref{sec:m-thermo}, Eq.~\ref{eq:survivor}).

Second, the gate makes \emph{mistakes}. We call the verdict that the gate gives the agent the \emph{served} verdict, and the verdict that the tree deserves the true verdict. When the served verdict is wrong with probability $P$, some trees that are truly red land, and we call each of them a \emph{false landing}. The share of landings that are false exceeds $P$ once more than half of the attempts are truly red (\S\ref{sec:m-daemon}, Eqs.~\ref{eq:pfalse} and \ref{eq:gain}).

\begin{figure*}[!tp]\centering\includegraphics[width=\textwidth]{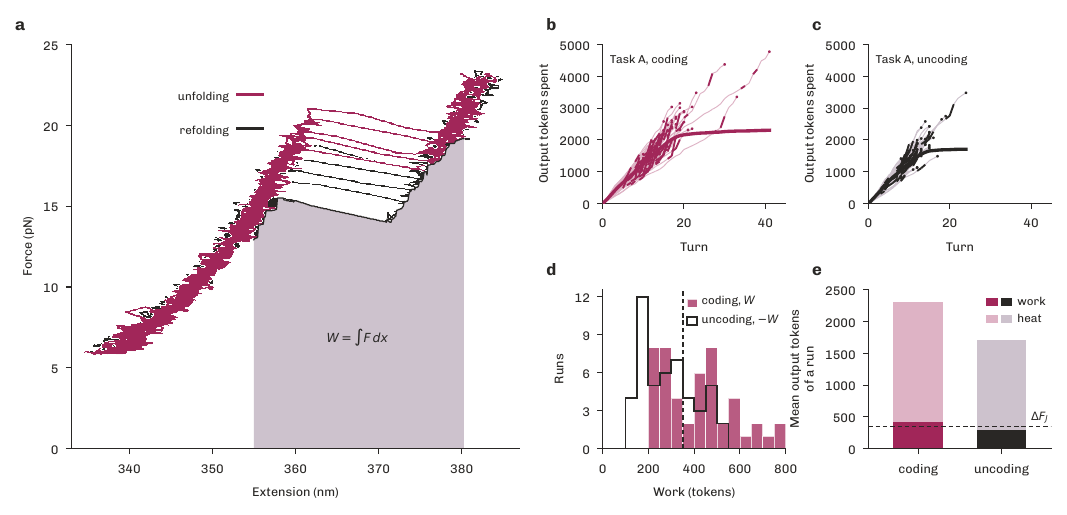}
\caption{A molecule pulled open and folded again, and a change made and undone. (a) Force against extension for one RNA hairpin held in optical tweezers and pulled at a loading rate of 7.5 pN/s: five unfolding curves (rose) and five refolding curves (black), redrawn from the data of Fig.~1 of ref.~\cite{collin2005verification} (\S\ref{sec:m-curves}). The shaded area under one refolding curve is the work returned as the molecule folds, 392 pN nm between 355 and 380 nm of extension, or 95 $k_BT$ at 298 K. (b--e) The coding agent on Task A at Hinton temperature 0.5. (b) The 48 landed coding runs (rose), in which the agent made the change (forward). Each thin line is one run: the output tokens it had spent after each turn, ending in a dot at the turn where it landed. A turn that changed the state $q$ is drawn dark, and its tokens are work. A turn that left $q$ unchanged is drawn pale, and its tokens are heat. The thick line is the mean over the runs, each run held at its final cost after it landed. (c) The 48 landed uncoding runs (black), in which it undid the change (reverse), on the same axes. (d) The work of the runs of b and c, in bins of 50 tokens: $W$ for coding (filled) and $-W$ for uncoding (outline). The dashed line is the free energy of the change, $\Delta F_J$, 351 tokens. (e) The mean output tokens of a run in each direction, divided into work (dark) and heat (pale). The dashed line is $\Delta F_J$. Supplementary Fig.~\ref{fig:cycleB} gives panels b to e for Task B.}
\label{fig:cycle}\end{figure*}

\subsection{Coding costs more than uncoding}
\label{sec:cycle}
Coding cost more than uncoding. We studied two changes, Task A and Task B. Each task renames a working module in place and updates every file that refers to it, much as relabelling a book requires correcting every catalogue card that points to it. We describe what the two modules do in \S\ref{sec:m-system}. Their functions do not affect this comparison. Task A is the longer of the two: its cycle cost about 1.5 times the tokens of the cycle of Task B (\S\ref{sec:extended}, Table~\ref{tab:cycle}). Neither task changes what the code does, and each can be undone exactly.

At three Hinton temperatures, we ran each task in both directions, coding (forward) and uncoding (reverse). This gives six conditions, and for each one we measured the heat and the asymmetry (\S\ref{sec:extended}, Table~\ref{tab:cycle}). The asymmetry is zero when the two directions cost the same. The asymmetry was positive in all
six conditions, and its 97.5 per cent interval excluded zero in four. For Task A, the intervals included zero at the two lower Hinton temperatures. Note that symmetry is different from reversibility. Symmetry means that the two directions cost the same, however much they cost. Reversibility means that the whole cycle costs nothing.

Some runs did not land, and the cost of a landing depends on how we count them. In thermodynamics, a mean is taken over all the trajectories of an experiment. A molecule that fails to unfold when it is pulled still counts in the mean work. We therefore report the cost of one landing under three approximations, which differ only in what they assume about the unfinished runs (\S\ref{sec:m-unfinished}). Let $C_i$ be the cost of run $i$ and $n_{\rm L}$ the number of landed runs of one direction.

The first way counts the landed runs alone. We call it VIVA, for victors in, vanquished absent:
\begin{equation}
\langle C\rangle_{\rm VIVA} = \frac{1}{n_{\rm L}} \sum_{i\ \rm landed} C_i .
\label{eq:viva}
\end{equation}
The second way counts every token of every run, with each unfinished run cut where it stopped, and divides by the number of landings that these tokens bought. We call it CORTA, the cut-off run truncation approximation:
\begin{equation}
\langle C\rangle_{\rm CORTA} = \frac{1}{n_{\rm L}} \sum_{i\ \rm all} C_i .
\label{eq:corta}
\end{equation}
The third way follows each run that the limit stopped until the context window of the model is full, because a run cannot continue beyond that point. We call it LOCA, the limit of context approximation. The window holds $M = 1{,}048{,}576$ tokens. A stopped run $i$ had a prompt of $p_i$ tokens at the limit. Over the last two thirds of its turns the prompt grew by $a_i$ tokens a turn, and the model generated $g_i$ tokens a turn. At these rates the window fills after $(M - p_i)/a_i$ more turns, and the run generates $g_i\,(M - p_i)/a_i$ more tokens. The mean cost of a landing is then
\begin{equation}
\langle C\rangle_{\rm LOCA} = \frac{1}{n_{\rm L}} \left[\, \sum_{i\ \rm all} C_i + \sum_{i\ \rm stopped} g_i\,\frac{M - p_i}{a_i} \right] .
\label{eq:loca}
\end{equation}
VIVA leaves out tokens that were spent, and it is a lower bound on the cost of a landing under CORTA. CORTA is what we spent, and it depends on our limit. LOCA is the cost when no stopped run ever lands. The tokens of the unfinished runs were heat: we show in \S\ref{sec:free} that they change the free energy of the change by at most 13 per cent.

\begin{figure*}[!tp]\centering\includegraphics[width=0.9\textwidth]{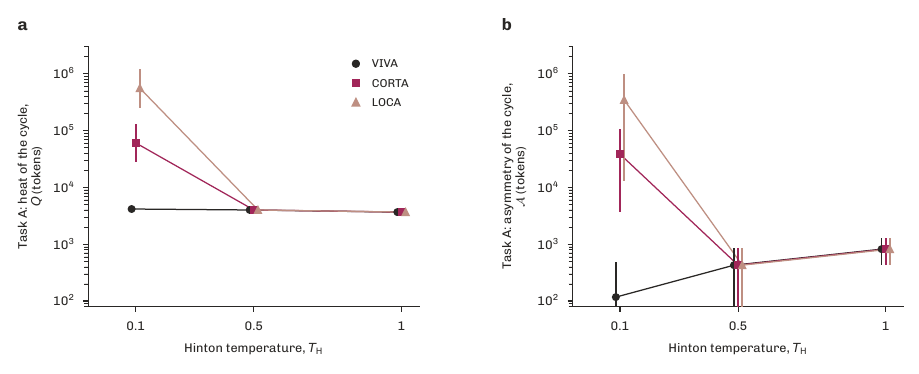}
\caption{The heat and the asymmetry of the cycle of Task A, counted in three ways. (a) The heat of the cycle, $Q$, against the Hinton temperature, $T_{\rm H}$, at 0.1, 0.5 and 1. (b) The asymmetry of the cycle, $\mathcal{A}$. Black circles employ the VIVA approximation, which means that they are the landed runs alone (Eq.~\ref{eq:viva}). Rose squares employ the CORTA approximation, which means that all runs are counted, each up to the call at which it stopped (Eq.~\ref{eq:corta}). Tan triangles employ the LOCA approximation, in which a run stopped by the limit on turns is followed until its context window is full (Eq.~\ref{eq:loca}). The vertical axes are logarithmic. Bars are 95 per cent bootstrap intervals over runs, and a bar that reaches the bottom of an axis extends below it. We offset the three counts horizontally within each condition so readers can distinguish the bars. The three conditions of Task A hold 144 runs, of which 131 landed. Supplementary Fig.~\ref{fig:threewaysB} shows Task B.}
\label{fig:threeways}\end{figure*}

The three approximations agreed where every run landed, and they differed by up to two orders of magnitude where runs were unfinished (Fig.~\ref{fig:threeways} for Task A; Supplementary Fig.~\ref{fig:threewaysB} for Task B). On Task A at $T_{\rm H} = 0.1$, the cycle heat was about 4{,}000 tokens under VIVA, 61{,}000 under CORTA and 556{,}000 under LOCA (\S\ref{sec:m-cost}). The asymmetry was positive in all six conditions under each of the three approximations (\S\ref{sec:extended}, Table~\ref{tab:cycle}).

Survivor selection does not explain the asymmetry at the middle Hinton temperature. In a separate sample with twenty runs a direction at that temperature, the asymmetry was positive on both tasks, and on Task B it grew when the unfinished coding runs were counted (\S\ref{sec:m-cost}). Survivor selection remains a possible contributor to the asymmetry at the coldest Hinton temperature, where the interval for the asymmetry included zero when all runs were included.

\subsection{The cycle is a hysteresis loop}
\label{sec:engine}
Every token spent on the cycle was lost, and most of them were heat. A cycle that returns the state to its start at a single temperature stores nothing and delivers nothing. Its classical counterpart is the hysteresis loop of a molecule that is pulled open and allowed to fold again~\cite{liphardt2002equilibrium,collin2005verification}: opening costs more work than folding returns, and the difference is lost as heat. Figure~\ref{fig:cycle} sets such a molecule (a) beside our two directions, one line a run, for Task A at Hinton temperature 0.5 from the runs of \S\ref{sec:free} (b--e). Supplementary Fig.~\ref{fig:cycleB} shows Task B at the same Hinton temperature. On Task A a landed coding run cost about 2{,}300 tokens on average and a landed uncoding run about 1{,}700 (Fig.~\ref{fig:cycle}b,c). On both tasks, 74 to 83 per cent of these tokens were generated in turns that left the state unchanged (Fig.~\ref{fig:cycle}e; \S\ref{sec:m-free}). The mean work of coding exceeded the free energy of the change by 64 tokens on Task A and by 111 on Task B (Fig.~\ref{fig:cycle}d; Supplementary Fig.~\ref{fig:cycleB}c).

The token record of a run carried a weak arrow of time. We asked how well a run's sequence of tokens per turn can be told from the same sequence read backwards, which measures the asymmetry of its time order. In a physical system this asymmetry is a lower bound on the entropy produced. Here we use it by analogy, and only on the runs that landed. On runs kept out of the fit, the estimate was about 0.3 nats a run for coding on Task A and 0.4 for uncoding, and its interval included zero on Task B (\S\ref{sec:m-arrow}). On Task A the estimate, multiplied by the software temperature, was about 1 per cent of the cycle heat. The estimate uses only the tokens of each turn, so the asymmetry of the full record can be larger.

\begin{figure}[!htbp]\centering\includegraphics[width=\columnwidth]{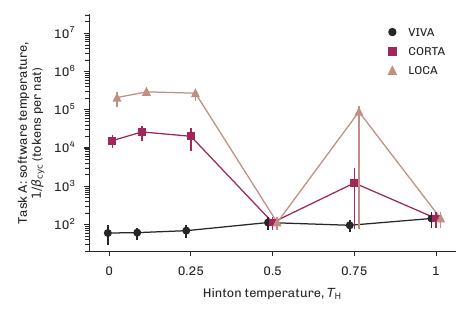}
\caption{The software temperature of the cycle, $1/\beta_{\rm cyc}$, in tokens per nat, against the Hinton temperature, $T_{\rm H}$, for Task A, counted in three ways: the VIVA approximation (black circles, Eq.~\ref{eq:viva}), the CORTA approximation (rose squares, Eq.~\ref{eq:corta}) and the LOCA approximation (tan triangles, Eq.~\ref{eq:loca}), as in Fig.~\ref{fig:threeways}. We measure the software temperature and set the Hinton temperature. Both quantities are defined in \S\ref{sec:intro} and \S\ref{sec:m-units}. The vertical axis is logarithmic. Bars are 95 per cent bootstrap intervals over runs. The six Hinton temperatures are 0.01, 0.1, 0.25, 0.5, 0.75 and 1. The six conditions of Task A hold 598 runs, of which 546 landed. Supplementary Fig.~\ref{fig:softTB} shows Task B.}
\label{fig:softT}\end{figure}

In the six conditions of Table~\ref{tab:cycle} (\S\ref{sec:extended}) the software temperature of the cycle rose with the Hinton temperature
on Task A, and rose then levelled on Task B. Over the landed runs of the twelve conditions (VIVA) it was 62 to 147 tokens per nat on Task A and 48 to 80 on Task B, higher at the higher Hinton temperatures (Fig.~\ref{fig:softT}; Supplementary Fig.~\ref{fig:softTB}). When the unfinished runs are counted, the cost varies far more from run to run, and the same formula gives 11{,}000 to 27{,}000 tokens per nat under CORTA and 210{,}000 to 370{,}000 under LOCA in the conditions below $T_{\rm H} = 0.5$. A rise with the Hinton temperature is consistent with a temperature interpretation. The run-length null of Supplementary Note~1 also predicts this rise whenever run lengths vary more at a higher Hinton temperature. The trend therefore cannot distinguish these explanations.

The cycle involves one temperature, the model's. The gate has a temperature of its own when it errs. We measure it in \S\ref{sec:mistake}, on the same repository with a known error
injected.

A software temperature of about a hundred tokens per nat has a plain meaning. Runs of the same task differ in cost
and end on different code, and the software temperature ties those two facts together as temperature in physics
ties energy to disorder. The values measured on Task A, 128 to 153 tokens per nat (\S\ref{sec:free}), are about 100 tokens per bit, a nat
being 1.44 bits. The number has two equivalent readings. As a Boltzmann factor, it says that if one way of doing
the task costs 100 tokens more than another, the agent takes it about half as often; we define the number this way, from the spread of work over attempts, and it measures the model's willingness to spend. Read as a price, by analogy with Landauer's bound, it would say that settling one choice between two ways costs at least about 100 tokens. We did not measure such a price. We call a fork any place where the code could have gone more than one way:
which file, which name, where the new line goes, whether to run the tests now. A fork with $k$ branches carries $\log_2 k$ bits, so a fork between two ways carries one bit. This price has the form of Landauer's bound, $k_BT\ln 2$ for each bit, with tokens in place of energy~\cite{landauer1961}. We use the bound by analogy, as we use the fluctuation relations.

The two readings meet in the conditions above. A landed run of Task A at Hinton temperature 0.25 cost about 1,950 tokens at the median, which at this price would pay for at most about twenty choices between two ways. The coding runs of Task A ended on 19 to
39 distinct trees across the five Hinton temperatures of Table~\ref{tab:trees} (\S\ref{sec:extended}), four to five bits of spread; its
uncoding runs ended on 2 to 10, one to three bits; and Task B ended on 2 to 4 a side. On Task A, then, the runs left four to five bits open, at forks that the gate did not decide. We did not count the forks that every run settled the same way. What remains of the twenty bits would be an upper limit on them, because the cost of a run includes heat. A count of distinct trees over a few
dozen runs is a lower bound on the number of trees. The software temperature is the exchange rate between those tokens and
those bits.

\begin{figure*}[!tp]\centering\includegraphics{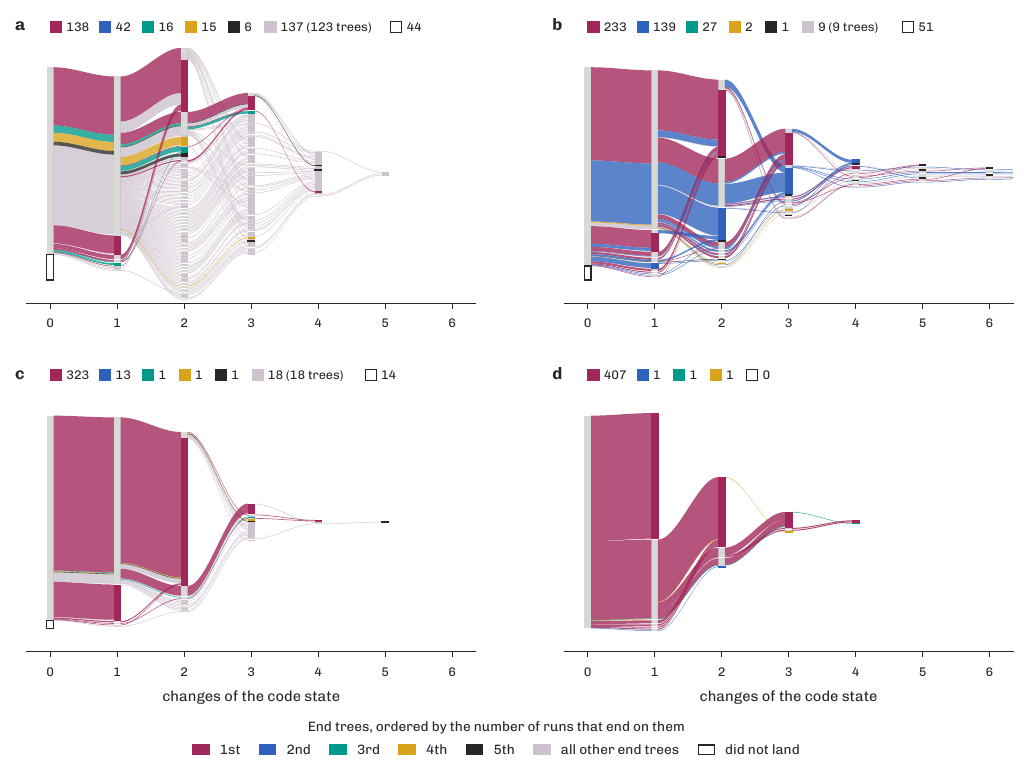}
\caption{Where the runs go. Each panel follows all the runs of one task in one direction, from the tree on which they start (step 0) through each change of the code. A grey bar is a tree, and a band joins one tree to the next: the wider the band, the more runs took that step. A run is coloured by the tree on which it ends. In each panel the end trees are ordered by the number of runs that end on them: rose is the first, followed by blue, teal, ochre and black, and pale mauve gathers all the others. The row beside each panel letter gives the number of runs of each colour. The hollow square in that row gives how many of these runs did not land, and such a run ends on a hollow bar, whatever its colour. (a) Task A, coding. (b) Task B, coding. (c) Task A, uncoding. (d) Task B, uncoding. The coding runs fan out: their stream splits at almost every step, and on Task A many runs end on a tree that few others reach. The uncoding runs stay together, and nearly all of them end on one tree. Making a change leaves many forks open, and undoing it leaves almost none. The six Hinton temperatures are pooled, the vertical scale is the same in the four panels, and the counts of each panel are in \S\ref{sec:m-paths}.}
\label{fig:flow}\end{figure*}

The forks show when the runs are followed one change at a time (Fig.~\ref{fig:flow}). We identified the tree after each code change. For each task and direction, we drew all runs as a stream from their common starting tree. The stream splits wherever two runs part, and each split is a fork. The trees on which the stream ends
are the ones the entropy $S$ of Table~\ref{tab:dictionary} counts: the entropy of a finished change is the
logarithm of the number of distinct trees that count as done. On Task A the landed coding runs ended on 125 different trees and the landed uncoding runs on 21. Weighting each tree by its share of the runs gives the entropy of the ending: 4.6 bits in coding and 0.6 in uncoding on Task A, and 1.1 and 0.1 on Task B. The count of trees is a lower bound, and the entropy reads low at a finite number of runs (\S\ref{sec:m-paths}).

\subsection{A run crosses three basins of the visit potential in order, and cold runs are lost to a loop}
\label{sec:wells}
Ever since Alder and Wainwright ran the first molecular dynamics simulations~\cite{alder1957phase,alder1959studies}, computational chemists and physicists have used the computer as a computational microscope to understand the world~\cite{krenn2022scientific}. Visualizing, in the sense of statistical mechanics, where the runs spend their time is therefore a useful tool for humans to understand the dynamics of a system: its ensembles of trajectories, its kinetics, its states and its traps. This section describes our attempt to do so in the context of agentic coding.

\begin{figure*}[!tp]\centering\includegraphics[width=0.8\textwidth]{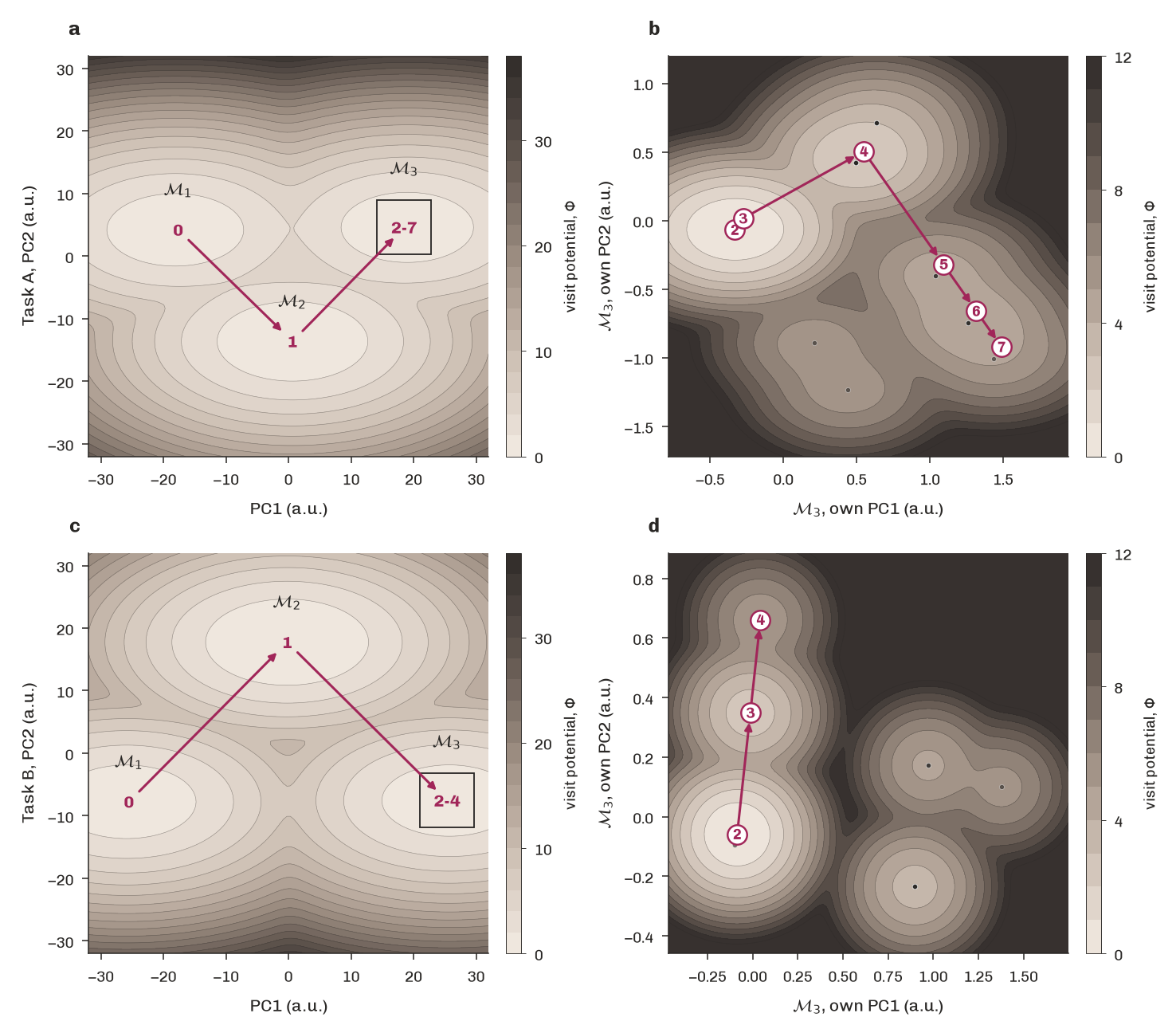}
\caption{A run crosses three basins of the visit potential, and the last one has structure of its own. (a) Task A. The shading is the visit potential, $\Phi$, of the repository states that the runs visited, drawn on the two main axes of those states (PC1, PC2). A lighter shade means that the runs spent more of their edits there. The three light regions are the basins $\mathcal{M}_1$, $\mathcal{M}_2$ and $\mathcal{M}_3$. The arrows follow one coding run that landed, and the numbers count its edits: the run starts in $\mathcal{M}_1$, is in $\mathcal{M}_2$ after one edit, and spends the rest of its edits in $\mathcal{M}_3$, the region inside the square. (b) $\mathcal{M}_3$ of panel a, enlarged and drawn on its own two axes. The dots are the distinct states that the coding runs visited there, and the same run passes through several of them before it lands. (c, d) The same for Task B. A run therefore does not reach its goal in one step: it passes through an intermediate state, and it still has choices to make after it arrives in the last one. The contours are 2 units of $\Phi$ apart in a and c and 1 unit apart in b and d, and the colour bars give $\Phi$. The runs, the counts of states and how the two runs were chosen are in \S\ref{sec:m-map}.}
\label{fig:wells}\end{figure*}

We described every repository state that a run visited by a descriptor vector. Like any representation~\cite{zhang2026molecular}, this one is arbitrary, and we hope that it is a set of descriptors that defines the system. The descriptor vector that we chose is composed of the counts of the syntax elements, such as function definitions, calls and assignments, that the edits of the run had added, less the counts of those that they had removed (\S\ref{sec:m-map}). The first two principal components of these vectors, over all the states of a task, are the two axes of the map. On these axes we drew the \emph{visit potential}, $\Phi = -\ln\rho$, where $\rho$ is the density of the visited states (Fig.~\ref{fig:wells}). The visit potential is low where the runs spent many of their edits. It is built from a density, as are the potential of mean force of a system at equilibrium~\cite{kirkwood1935statistical} and the quasi-potential of a system out of equilibrium~\cite{freidlin1984random,wang2008potential}. It is a pure number, fixed up to a constant, and it depends on the description of a state and on the smoothing of the density. The free energy of
\S\ref{sec:free} is a different quantity, measured in tokens from the work.

On both tasks the visited states gather around three minima of the visit potential. We call the set of repository states around one minimum a \emph{basin}, as the region around a minimum of a potential energy surface is called~\cite{stillinger1982hidden}. In the kinetic networks of molecular simulation, a basin in which a system stays for a long time before it jumps to another is a metastable state~\cite{prinz2011markov}. We number the three along the coding direction: $\mathcal{M}_1$, $\mathcal{M}_2$ and $\mathcal{M}_3$. In the words of a chemist, a coding run starts in the reactant state $\mathcal{M}_1$, passes through the intermediate state $\mathcal{M}_2$ and lands in the product state $\mathcal{M}_3$. An uncoding run crosses them in the opposite order, as a reverse reaction does. Of the 384 runs of the map, 288 landed, and 281 of these crossed the three basins in the order of their direction and never went back to an earlier one (\S\ref{sec:m-map}). Only the first basin held a run for long. A landed run stayed in it for 4 to 10 turns at the median, and then reached the second and the third in a single turn. The first basin is therefore metastable in the sense above, and the second is a place of passage. The run drawn in Fig.~\ref{fig:wells} was in $\mathcal{M}_2$ after one edit, and spent its remaining edits in $\mathcal{M}_3$, where it landed. Within $\mathcal{M}_3$, coding runs visited 13 distinct states on Task A and 7 on Task B.

\begin{figure*}[!tp]\centering\includegraphics[width=0.95\textwidth]{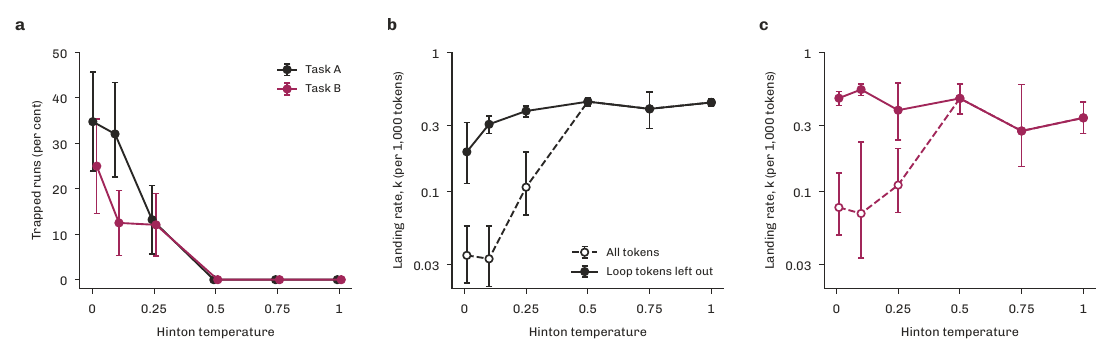}
\caption{Cold runs end in a loop, and the landing rate falls less when the tokens of the loop are left out. (a) The share of the coding runs that end trapped, against the Hinton temperature, for Task A (black) and Task B (rose). A run is trapped when it ends by repeating one command, with no edit between the repeats. No run is trapped from a Hinton temperature of 0.5 up. (b) The landing rate $k$ of the coding runs of Task A, the number of runs that landed divided by the tokens spent, on a logarithmic axis. The hollow points and the dashed line count every token of every run. The filled points and the solid line leave out the tokens that a trapped run spent after it entered its last loop. Where the two lines part, the tokens of the loops are what lowers the rate. (c) The same for Task B. Each point is one condition of 46 to 137 runs, 765 runs in all, and the bars are 90 per cent bootstrap intervals over runs. A run is stopped at 300 turns.}
\label{fig:trapped}\end{figure*}

Not every run reached $\mathcal{M}_3$. At Hinton temperatures of 0.25 and below, 12 to 35 per cent of the coding runs ended in a loop: they repeated one command, with no edit between the repeats, until the limit of 300 turns stopped them (Fig.~\ref{fig:trapped}a; \S\ref{sec:m-unfinished}). We call such a run trapped. From 0.5 to 1 no run was trapped. A trapped run continues to generate tokens after its last code change. We therefore computed the landing rate, the number of landed runs divided by the tokens spent, in two ways: with all the tokens, and without the tokens that a trapped run spent after it entered its last loop (Fig.~\ref{fig:trapped}b,c).

We read these loops turn by turn (\S\ref{sec:m-unfinished}). Nearly all of the runs that the limit stopped repeated one or two search commands that read the repository and returned the same output each time. A run in such a loop receives no new information, and at a low Hinton temperature the model is likely to return the same command for the same context. We therefore assume in LOCA that a stopped run does not land.

The tokens of the loop carry most of the fall of the landing rate at low Hinton temperature on Task A and all of it
on Task B. With all tokens counted, the landing rate of the coding runs was about 13 times lower at a Hinton temperature of 0.01 than at 1 on Task A, and about 4 times lower on Task B. With the tokens of the last loop left out, it was about 2 times lower on Task A, and on Task B it did not fall (\S\ref{sec:m-unfinished}). Loops of this kind are known. A language model that has repeated itself is more likely to repeat again~\cite{holtzman2020curious,xu2022learning}, coding agents repeat their actions in the runs that fail~\cite{yang2024sweagent,yang2025swesmith}, and some harnesses stop a run that repeats one action with the same result~\cite{openhands2026stuck}. Our harness has no such guard, and this comparison is an accounting of the stored runs. A shorter
limit on turns can be read from the same runs, because the agent is not told the limit. A run cut at a given turn is therefore the run that a shorter limit would have produced. No run at a Hinton temperature of 0.25 or below landed after
turn 29. Cut at 40 turns, the stored runs keep every one of these landings. The landing rate of the cold conditions is then 2.6 times lower than that of the warm conditions on Task A and 1.6 times lower on Task B (\S\ref{sec:m-unfinished}).


\subsection{Eighty per cent of the tokens of a landed run are heat, and the work that remains gives the free energy of the change}
\label{sec:free}
\begin{figure*}[!tp]\centering\includegraphics[width=0.9\textwidth]{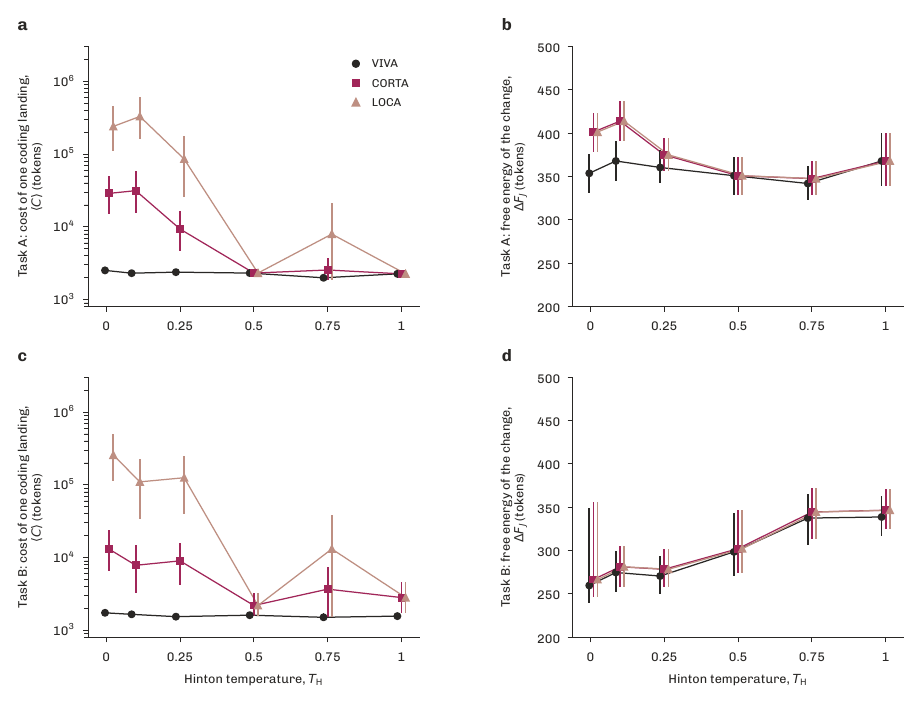}
\caption{Where the tokens go. (a, c) The cost of one coding landing against the Hinton temperature, $T_{\rm H}$, for Task A (a) and Task B (c), under the VIVA approximation (black circles, Eq.~\ref{eq:viva}), the CORTA approximation (rose squares, Eq.~\ref{eq:corta}) and the LOCA approximation (tan triangles, Eq.~\ref{eq:loca}). The vertical axis is logarithmic. Below a Hinton temperature of 0.5 the three approximations stand one to two orders of magnitude apart, because that is where runs fail to land. (b, d) The free energy of the change, $\Delta F_{J}$ (Eq.~\ref{eq:jarzynski}), for Task A (b) and Task B (d), on a linear axis. Under VIVA it is read from the landed coding runs. Under CORTA and LOCA it includes the term for the unfinished runs (Eq.~\ref{eq:dfall}), which is the same in both. The free energy hardly moves between the approximations: the unfinished runs multiply the cost of a landing and change the free energy by at most 13 per cent. The bars are 95 per cent bootstrap intervals over runs in a and c and 90 per cent in b and d. The values are not corrected for the bias of few runs, which lowers the estimates of Task B by 4 to 47 tokens (Test 5 of \S\ref{sec:freetests}). The six Hinton temperatures are 0.01, 0.1, 0.25, 0.5, 0.75 and 1. The twelve conditions hold 1{,}200 runs, of which 1{,}110 landed.}
\label{fig:tokens}\end{figure*}

Ever since Joule measured how much mechanical work produces a given amount of heat~\cite{joule1850mechanical}, thermodynamics has divided the energy that a system exchanges with its surroundings into work and heat. The division is useful because the two parts play different roles: the work of a process bounds the free energy difference between its two end states (Eq.~\ref{eq:secondlaw}), and the heat does not enter that bound. This section divides the tokens of a run in the same way, and then obtains the free energy of the change from the work.

Most of what a landed run generates leaves the state unchanged. Over the 1{,}110 landed runs of Table~\ref{tab:cells} (\S\ref{sec:extended}), 80 per cent
of the tokens were heat in the sense of Eq.~\ref{eq:work}: they were generated in turns that left the state where it
was. The share was between 73 and 85 per cent in every condition and in both directions. A landed run took 9 to 17 turns
on average, and about two of them moved the state. Task A at $T_{\rm H} = 0.5$ illustrates the division between work and heat. A landed coding run cost about 2{,}300 tokens on average, with about 400 tokens of work and 1{,}900 of heat (\S\ref{sec:m-free}).

The work itself varies from run to run, and its spread is what gives the free energy. In thermodynamics the
free energy difference between two states can be read from the work of many repetitions of one process that drives
the system from the first state to the second, however far from equilibrium the process runs. This is Jarzynski's equality~\cite{jarzynski1997nonequilibrium},
\begin{equation}
\left\langle e^{-\beta W}\right\rangle = e^{-\beta\,\Delta F} ,
\label{eq:jarzynski}
\end{equation}
where $W$ is the work of one repetition, the average runs over repetitions, $\beta$ is the inverse temperature and
$\Delta F$ is the free energy of the second state less that of the first. The exponential average gives the most weight to the repetitions that cost the least work, so the equality returns a $\Delta F$ at or below the mean work, as Eq.~\ref{eq:secondlaw} requires. For
work with a Gaussian distribution of variance $\sigma^{2}$ it reduces to
$\Delta F = \langle W\rangle - \beta\sigma^{2}/2$. The equality was first tested by pulling single RNA molecules
open~\cite{liphardt2002equilibrium}, and pulling the same molecule open and letting it close gives the free energy
from both directions~\cite{collin2005verification}.

The equality is proved under assumptions that the runs do not meet, and we use it by analogy. It holds for a system
that starts in equilibrium with a bath at the temperature $1/\beta$ and is then driven along a fixed protocol. The runs of a given condition all start from one stored tree, the instruction to the agent stands in for the protocol, and no bath
sets a temperature. A further difference is in the sign of the work. A molecule that closes returns work to the
apparatus. An agent that removes a block generates tokens as it does so, and Eq.~\ref{eq:work} counts them as
negative because the state moved back. The work of an uncoding run, $W_{R}$, is therefore negative, and the reverse form of the equality, $\langle e^{-\beta W_{R}}\rangle = e^{+\beta\,\Delta F}$, is applied to it (\S\ref{sec:m-free}).

No bath sets $\beta$ in the coding experiment. We estimate it from the work in both directions. In the Gaussian form of Jarzynski's equality each direction gives one equation for
the same free energy, $\Delta F = \langle W_{F}\rangle - \beta\sigma_{F}^{2}/2$ from the coding runs and
$\Delta F = -\langle W_{R}\rangle + \beta\sigma_{R}^{2}/2$ from the uncoding runs. The two equations have two
unknowns, and their solution for the temperature is
\begin{equation}
\beta = \frac{2\left(\langle W_{F}\rangle + \langle W_{R}\rangle\right)}{\sigma_{F}^{2} + \sigma_{R}^{2}} .
\label{eq:betawork}
\end{equation}
The sum in the numerator is the work left over after a cycle, and it is zero for a cycle run reversibly. We call $1/\beta$ the \emph{software temperature} of the condition. It was about 130 to 150 tokens per nat on Task A and about 50 to 90 on Task B (\S\ref{sec:m-free}). This is the relation of Eq.~\ref{eq:betacyc} applied to the work, where the software temperature of the cycle (\S\ref{sec:engine}) applies it to the cost. The Gaussian form is an approximation, and the comparisons at the end of this section test it with the full exponential averages.

With $\beta$ fixed in this way, Eq.~\ref{eq:jarzynski} applied to the coding runs gave a free energy of about 340 to 370 tokens for the change of Task A and about 270 to 340 tokens for that of Task B (\S\ref{sec:m-free}). This free energy is 14 to 22 per cent of the mean cost of a landed coding run. In Task A at $T_{\rm H} = 0.5$, of the 415 tokens of coding work, about 350 are the free energy of the change and about 65 are dissipated work. On Task A the free energy did not move with the Hinton
temperature beyond its intervals. On Task B it rose from about 270 tokens at $T_{\rm H} = 0.1$ to about 340 at $T_{\rm H} = 1$, and the uncoding runs gave a rise of the same size. We cannot distinguish two explanations for Task B. The change may require more work at a higher Hinton temperature, or the estimate may be biased low in the two colder conditions, where 80 per cent of the coding tokens were in unfinished runs.

The unfinished runs can be counted in the free energy. An unfinished run never completes the change, so it adds nothing to the average of Eq.~\ref{eq:jarzynski}, however long it is followed. Only the share of runs that land enters. With $u_F$ and $u_R$ the unfinished shares of the coding and the uncoding runs, the free energy over all runs is~\cite{alemany2012experimental}
\begin{equation}
\Delta F_J^{\rm all} = \Delta F_J^{\rm landed} - \frac{1}{\beta}\,\ln\frac{1-u_F}{1-u_R} .
\label{eq:dfall}
\end{equation}
The second term was at most 48 tokens, 13 per cent of the free energy, and it is the same under CORTA and under LOCA (\S\ref{sec:m-free}). With it, the rise on Task B remains. In the two coldest conditions of Task A, the cost of a coding landing rose from about 2,400 tokens under VIVA to about 30,000 under CORTA and to about 240,000 and 330,000 under LOCA, while the free energy rose by less than 50 tokens. The unfinished runs therefore add heat and almost no free energy.

\subsubsection{Five tests of the free energy}\label{sec:freetests}
As powerful as Jarzynski's equality is, it has limitations. First, it needs a temperature, and the software temperature is one that we fitted from the same runs. Second, its exponential average is dominated by the few runs of lowest work, so it is sensitive to the shape of the work distribution. Third, it overestimates the free energy when the runs are few. These limitations, combined with the fact that we are extending thermodynamics to software, require that we test our estimate of the free energy with care. In this section we therefore carry out five tests. Each test explores one component of the estimate: the direction of the runs, the temperature, the Gaussian approximation for the work, or the number of runs. We list the tests below and what we learn from each.

\paragraph{Test 1. Uncoding gives the same free energy as coding to within 40 tokens.} Jarzynski's equality can be applied to the uncoding runs as well as to the coding runs, and the two must give the same free energy. In the Gaussian approximation they agree by construction, because Eq.~\ref{eq:betawork} is the temperature that makes them equal. We therefore compared the two exponential averages in full, with the same $\beta$ and the same runs. The two directions then differ only through the part of the work distribution that is not Gaussian. In the retained conditions, the uncoding estimate differed from the coding one by less than 20 tokens on Task A, and it was 20 to 40 tokens higher on Task B (\S\ref{sec:m-free}). The part of the work that is not Gaussian therefore moves the free energy by less than 15 per cent.

\paragraph{Test 2. An estimate without a temperature agrees.} The second test uses an estimate that requires no software temperature. If the two directions have the same variance, the two Gaussian equations give
\begin{equation}
\Delta F_{S} = \tfrac{1}{2}\left(\langle W_{F}\rangle - \langle W_{R}\rangle\right) ,
\label{eq:symmetric}
\end{equation}
the half difference of the mean works, in which $\beta$ does not appear. It lay within 2 tokens of $\Delta F_{J}$ on Task A and within 23 tokens on Task B in the nine conditions that we retain, although the variance of the coding work was up to 3 times that of the uncoding work. The free energy therefore changes little when the fitted temperature is taken out of it.

\paragraph{Test 3. The crossing of the two work distributions agrees.} The third test uses neither a temperature nor a Gaussian form. Under Crooks' relation the distribution of the coding work and that of the uncoding work, with its sign changed, cross at the free energy~\cite{crooks1999entropy}. Collin and co-workers read the free energy of the hairpin of Fig.~\ref{fig:cycle}a from this crossing~\cite{collin2005verification}, and Fig.~\ref{fig:cycle}d shows the two distributions for one condition of Task A. The crossing lay within about 20 tokens of $\Delta F_{J}$ on Task A and within about 30 on Task B without its coldest condition, and $\Delta F_{J}$ lay inside the 90 per cent interval of the crossing in each of the twelve conditions. On Task B the crossing rose with the Hinton temperature as $\Delta F_{J}$ did. The rise of the free energy on Task B therefore appears in an estimate that shares neither the temperature nor the Gaussian form.

\paragraph{Test 4. Estimators built for work that is not Gaussian agree.} The exponential average is dominated by the very few runs that carry the least work~\cite{jarzynski2006rare}. Researchers have developed estimators for work distributions that are not Gaussian, and here we apply two of them. The first estimator, by Procacci, fits a mixture of two Gaussians to the coding work and takes the exponential average of the fitted distribution~\cite{procacci2015unbiased}. The second estimator, Bennett's acceptance ratio, uses runs from both directions~\cite{bennett1976efficient}. The mixture of Gaussians lay within 8 tokens of $\Delta F_{J}$ and the acceptance ratio within 12 tokens, both under 4 per cent.  The work of the landed coding runs has no long tail: its skewness was between $-0.3$ and $+0.7$. The shape of the work distribution therefore changes the free energy by a few per cent at most.

\paragraph{Test 5. The number of runs biases Task B more than Task A.} The last test asks how the number of runs affects the estimate. The exponential average gives most of its weight to the runs of lowest work. A small sample rarely contains them, so it overestimates the free energy, and the overestimate falls as runs are added~\cite{gore2003bias}. Palassini and Ritort showed what sets its size: the dissipated work, in units of the temperature, compared with the logarithm of the number of runs~\cite{palassini2011improving}. When the dissipated work is well below that logarithm, the bias is small. Each of the nine conditions that we retain holds 42 to 51 landed coding runs (the three that we leave out hold 29, 32 and 36), and the logarithm of that number is about 3.8. On Task A the dissipated work was 0.3 to 0.6 nats, and the bias was 1 to 5 tokens. On Task B the dissipated work was larger, 0.8 to 2.4 nats, and the bias grew with it: about 5 to 10 tokens in the two warmest conditions and about 20 to 50 tokens in the two coldest. We computed these biases for Gaussian work with the measured variance. With the bias removed, the free energy of Task B rose from about 240 tokens at $T_{\rm H} = 0.1$ to about 330 at $T_{\rm H} = 1$. The crossing of Test 3, which does not use the exponential average, also lay up to 32 tokens below $\Delta F_{J}$ on Task B. The free energy of Task A therefore stands as measured, and that of Task B rises with the Hinton temperature more steeply than the uncorrected estimate shows.

Table~\ref{tab:freetests} sets the estimates side by side. All five tests use the same runs. An independent test would need a temperature measured by another method, which we do not have.

\begin{table*}[!tp]\centering\small
\captionsetup{justification=raggedright,singlelinecheck=false}\caption{The free energy of the change under each test, in tokens. The number in bold is the difference from $\Delta F_{J}$ of the coding runs. The columns are one condition of Task A and the coldest and the warmest quoted conditions of Task B. Test 5 gives two values, one for each of the two variances used for the bias (\S\ref{sec:m-free}). The estimates of the free energy therefore lie within 6 per cent of $\Delta F_{J}$ for Task A and within 17 per cent for Task B.}
\label{tab:freetests}
\begin{tabular}{@{}lccc@{}}
\toprule
estimate & Task A, $T_{\rm H} = 0.5$ & Task B, $T_{\rm H} = 0.1$ & Task B, $T_{\rm H} = 1$ \\
\midrule
$\Delta F_{J}$, coding (Eq.~\ref{eq:jarzynski}) & 351 & 275 & 339 \\
Test 1: $\Delta F_{J}$, uncoding & 334 \textbf{($-$5\%)} & 297 \textbf{($+$8\%)} & 360 \textbf{($+$6\%)} \\
Test 2: $\Delta F_{S}$ (Eq.~\ref{eq:symmetric}) & 352 \textbf{(0\%)} & 295 \textbf{($+$7\%)} & 340 \textbf{(0\%)} \\
Test 3: crossing & 373 \textbf{($+$6\%)} & 243 \textbf{($-$12\%)} & 333 \textbf{($-$2\%)} \\
Test 4: mixture of two Gaussians & 348 \textbf{($-$1\%)} & 275 \textbf{(0\%)} & 338 \textbf{(0\%)} \\
Test 4: Bennett's acceptance ratio & 346 \textbf{($-$2\%)} & 269 \textbf{($-$2\%)} & 327 \textbf{($-$4\%)} \\
Test 5: bias for few runs removed & 348, 348 \textbf{($-$1\%)} & 251, 228 \textbf{($-$9\%, $-$17\%)} & 330, 327 \textbf{($-$3\%, $-$4\%)} \\
\bottomrule
\end{tabular}
\end{table*}

For future measurements we recommend Bennett's acceptance ratio. It uses the runs of both directions at once, and Bennett showed that it has the least variance of the estimators built on the two work distributions~\cite{bennett1976efficient}. Here it lay within 4 per cent of $\Delta F_{J}$ in every quoted condition. We suggest reporting the crossing of the two work distributions beside it, because the crossing needs no temperature. We also suggest enough runs that the logarithm of their number is well above the dissipated work in nats.

\subsubsection{The ratio of the two landing rates does not give the free energy}\label{sec:rates}
There is a second place to look for a free energy, familiar from chemical kinetics: the rates. For a transition and its own reverse, in a system held by one bath, the
logarithm of the ratio of the two rates is $\beta\,\Delta F$; this is local detailed
balance~\cite{seifert2012stochastic}. We therefore asked whether the rates of landing in coding and in uncoding give the free energy that the work gave. They do not. Coding landed more slowly than uncoding by 0.1 to 1.1 nats, in either of two ways of counting the rate, while $\beta\,\Delta F$ was 2.3 to 2.9 nats on Task A and 3.8 to 5.6 on
Task B (\S\ref{sec:m-free}). The two readings therefore stand about 2 nats apart on Task A and 3 to 5 nats apart on Task B.

We think that the rates disagree with the work for the following reason. Local detailed balance compares a transition with its own reverse: the same path, travelled backwards. Uncoding is not coding travelled backwards. An agent that undoes a change does not retrace the steps of the agent that made it. It follows its own instruction and finds its own path. The two are like a road up a mountain and a different road down: the times of the two journeys say little about the height of the mountain. The ratio of the two landing rates therefore compares the costs of two different paths, and the free energy does not set it.

\subsubsection{Uncoding from the ensemble of end trees}\label{sec:endtrees}
A thermodynamic state is an ensemble. In the experiments above, each uncoding run started from a single tree. In some sense, this mirrors the coding runs, which also started from a single tree. A thermodynamic argument can be made that, to really show that coding and uncoding are different, uncoding should start from the ensemble of trees on which the coding runs ended (\S\ref{sec:m-paths}). We made this second experiment on both tasks at $T_H=0.5$ (\S\ref{sec:m-endtrees}), and we describe Task A first.

\begin{figure*}[!tp]\centering\includegraphics{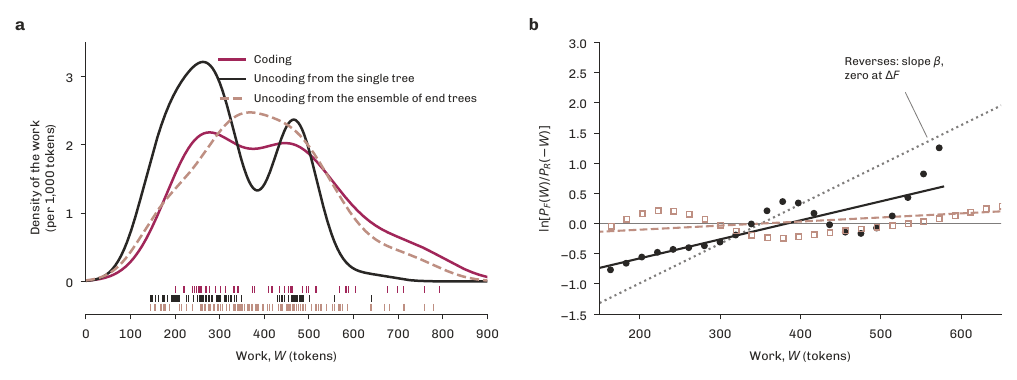}
\caption{Uncoding from the ensemble of end trees, on Task A at $T_H=0.5$. (a) The distribution of the work of the 48 landed coding runs (rose), and of the work of uncoding with its sign changed: 100 runs from the single tree (black) and 100 runs from the ensemble of end trees (dashed). Each tick under the curves is one run. The curves are kernel density estimates (\S\ref{sec:m-endtrees}). (b) The logarithm of Eq.~\ref{eq:crooks} from the single tree (filled circles) and from the ensemble (open squares), with the fitted lines. The dotted line is Crooks' relation, with slope $\beta$ and its zero at the free energy $\Delta F$.}
\label{fig:endtrees}\end{figure*}

If uncoding were the reverse of coding, the work of the two directions would obey Crooks' relation~\cite{crooks1999entropy},
\begin{equation}
\ln\frac{P_F(W)}{P_R(-W)}=\beta\,(W-\Delta F),
\label{eq:crooks}
\end{equation}
where $P_F$ and $P_R$ are the distributions of the work of coding and of uncoding. The logarithm would be a straight line in $W$ with slope $\beta$, which is 0.0065 per token on Task A. The measured slope was 0.0007 per token from the ensemble of end trees and 0.0032 per token from the single tree (Fig.~\ref{fig:endtrees}b). On Task A, Crooks' relation therefore does not hold between coding and uncoding, from either start.

The cost of uncoding did not depend on its starting tree. A landed uncoding run cost about 1,800 tokens from the ensemble and from the single tree, and a landed coding run cost about 2,300 tokens. From the ensemble, the work of uncoding had nearly the same distribution as the work of coding (Fig.~\ref{fig:endtrees}a). The two directions therefore did nearly the same work, and the extra cost of coding was heat.

Task B gave a different answer. There the measured slope was 0.019 per token from the ensemble and 0.022 per token from the single tree, and $\beta$ is 0.018 per token. Each line crossed zero within 13 tokens of the free energy (\S\ref{sec:extended}, Table~\ref{tab:endtrees}). On Task B the work of the two directions is therefore consistent with Crooks' relation. The cost is not: a landed coding run cost about 1,600 tokens and a landed uncoding run about 1,170 tokens, from either start. The coding runs of Task B ended on two trees only, so its ensemble is close to a single tree.

\subsubsection{A landing passes through at least five states}\label{sec:states}
The cost of a landing holds one more piece of information: how regular it is. If a landing were one rare event, like the decay of a nucleus, the cost of a landed run would follow an exponential distribution, and its spread would equal its mean. A landing that needs many steps in a row is more regular, in the way that the wait for ten buses varies less, in proportion, than the wait for one. Aldous and Shepp made this precise~\cite{aldous1987least}. They considered a Markov chain that passes through $n$ states before it stops, and showed that the squared coefficient of variation of its stopping time, the variance over the squared mean, is at least $1/n$. Their bound allows loops and backward steps.

We measured the squared coefficient of variation of the cost of a landed run, with cost in the place of time. It was 0.04 to 0.19 in every condition and direction, where a single rare event gives 1. This range leaves out one landed run of 74{,}298 tokens. With it, the value of its condition is 7.3 in place of 0.08 (\S\ref{sec:m-free}). Under the model of Aldous and Shepp, the process that generates a landing therefore has at least 5 phases in the most variable condition and at least 24 in the most regular. These are phases of the model, and a single run need not pass through all of them.

The count is a lower bound, and it holds under a model. A run keeps its whole history, so it is not a Markov chain, and the bound does not say what a state is (\S\ref{sec:m-free}). We take one thing from it: a landing is a sequence of many steps, as the three basins of \S\ref{sec:wells} already suggested.

\subsubsection{The extra cost of coding is real}\label{sec:shape}
In this section, we do the detective work of finding out whether the extra cost of coding over uncoding is a mirage or a reality. For that, we need statistical tests. We ask two questions. The first is whether the difference between the mean costs is larger than chance would give with our number of runs. The second is whether the extra cost sits in every coding run or in a few expensive ones.

\paragraph{Test 1. Coding costs more than uncoding in 11 of the 12 conditions.}
For the first question we used a permutation test (\S\ref{sec:m-free}). The test shuffles the labels coding and uncoding among the landed runs of a condition many times, and counts how often the shuffled difference of the mean costs is as large as the measured one. We ran one test for each of the 12 conditions. When twelve tests are run, one of them can pass the usual threshold of 0.05 by chance alone. We therefore also report each result after a correction for twelve tests, in which the threshold is divided by twelve. The permutation test found that coding cost more than uncoding in 11 of the 12 conditions, with and without the correction. The exception is Task A at $T_{\rm H} = 0.1$, where the extra cost was the smallest, 12 per cent.

\paragraph{Test 2. The extra cost is spread over all the landed runs.}
For the second question, we compare the distributions of coding and uncoding costs with the Kolmogorov--Smirnov test, which measures their largest gap. We first divided the cost of each landed run by the mean cost of its direction, so that only the spread of the costs around their mean is compared. If every coding run costs more by the same factor, the two divided distributions are the same. The Kolmogorov--Smirnov test found no difference in 10 of the 12 conditions, and in 11 of the 12 after the correction for twelve tests. There were only two exceptions, both on Task A, at $T_{\rm H} = 0.25$ and at $T_{\rm H} = 0.75$. Our hypothesis is that they are statistical fluctuations: among twelve tests at a threshold of 0.05, chance alone gives at least one exception in 46 per cent of cases, and one of the two disappears after the correction. This test has little power against a difference of spread, so we also compared the quantiles of the landed costs directly. The ratio of the cost of coding to that of uncoding was about the same at the 10th percentile, at the median and at the 90th percentile: 1.1 to 1.8, 1.1 to 1.5 and 1.2 to 1.5 over the conditions of Task A, and 1.2 to 1.5, 1.2 to 1.4 and 1.2 to 1.5 over those of Task B. The most expensive tenth of the landed runs held 15 to 19 per cent of the cost of coding and 14 to 21 per cent of the cost of uncoding.

Therefore, we conclude that the extra cost of coding over uncoding is statistically significant, and that it is spread over the landed runs and not carried by a few of them. As our group and other groups start working on software thermodynamics and more measurements become available, this result can be tested on other code bases and in other scenarios.

\subsubsection{At cold Hinton temperatures the tokens go to runs that never land}\label{sec:cold}
It is fitting that the authors of this article and Geoffrey Hinton live in the same cold city in Canada. In this section, we explore the cold Hinton temperatures, the conditions in which the agents enter situations from which they never land.

Every token that an agent spends belongs either to a run that landed or to a run that never did. We look at the two parts in turn, from the warmest condition, $T_{\rm H} = 1$, to the coldest, $T_{\rm H} = 0.01$.

The first part hardly depends on the temperature. A coding run that landed cost about 2{,}000 to 2{,}500 tokens on Task A and about 1{,}500 to 1{,}700 on Task B, whether the agent was warm or cold.

The second part depends strongly on the temperature (Fig.~\ref{fig:tokens}). On Task A, the unfinished runs took at most 22 per cent of the tokens of the coding runs from $T_{\rm H} = 0.5$ up, and 74 to 93 per cent below it. On Task B, they took 26 to 59 per cent from $T_{\rm H} = 0.5$ up, and 79 to 87 per cent below it. Every uncoding run of Task B landed.

The price of a landing is therefore the same in the cold as in the warmth. What the cold adds is the runs that never land, which are the loops that we described in \S\ref{sec:wells}. Like a car that spins its wheels in the snow of a Toronto winter, a looping agent burns tokens and goes nowhere. For this reason the VIVA, CORTA and LOCA approximations for the unfinished runs (Eqs.~\ref{eq:viva} to \ref{eq:loca}) differ most in the cold conditions.
\subsection{A noisy test suite catches no more defects than a file's own tests}
\label{sec:gate}
Software defects are expensive. In 2002, a study for the US National Institute of Standards and Technology put the cost of an inadequate infrastructure for software testing to the US economy at 59.5 billion dollars a year~\cite{nist2002economic}. Twenty years later, the Consortium for Information and Software Quality estimated a broader quantity, the cost of poor software quality in the United States, at 2.41 trillion dollars or more for 2022~\cite{krasner2022cost}, about 9 per cent of the output of the country in that year. If the rest of the world loses the same share of its output, the cost to the world is about 10 trillion dollars a year. This is an estimate of the order of magnitude and no more. To give it a scale, 10 trillion dollars a year is about what the world spent on health in 2021~\cite{who2023spending}, nearly twice the sales of the chemical industry of the world, which were 5.0 trillion euros in 2024~\cite{cefic2025facts}, and more than twice the value that the farms, forests and fisheries of the world add in a year~\cite{worldbank2026wdi}.

A single defect can be costly too. In 2012 the trading firm Knight Capital lost more than 460 million dollars in about 45 minutes, after new code was deployed incorrectly~\cite{sec2013knight}. In July 2024 one faulty update, released by the company CrowdStrike, cost the Fortune 500 companies an estimated 5.4 billion dollars in direct losses~\cite{parametrix2024crowdstrike}. The stakes are now rising, because the agents themselves have become capable attackers. An academic study found in 2024 that a language model exploited 87 per cent of a set of 15 known vulnerabilities~\cite{fang2024llm}, and Anthropic reports that its model Claude Mythos Preview, when a user directs it to, finds and exploits unknown vulnerabilities in every major operating system and every major web browser, some of which had stayed hidden for more than twenty years~\cite{carlini2026mythos}.

The gate, the test suite that decides whether a change may land (\S\ref{sec:defgate}), is the last defence against a defect. Therefore, having robust tests and understanding their thermodynamics is crucial. In this section we measure how robust a gate is. We first describe an experiment in which we plant defects and count how many of them the gate stops. We then show three results: the whole test suite caught as many defects as the tests of the changed file alone, its threshold lay inside its own noise, and calibrating the threshold did not rescue it.

\paragraph{The experiment.} To measure how many defects a gate detects, we planted the defects ourselves. No agent and no model took part. In each of sixty trials, on three machines, we took the clean repository and, with a probability of 50 per cent, planted one defect in one of six files. We then drew one of three gates at random: no gate at all, the tests of the changed file alone, or the whole test suite. A tree landed when its gate passed it.

The whole suite has a difficulty that the other two gates do not have. It fails a few tests even on clean code, and the number of these failures, $N$, changes from one run to the next. Programmers call such tests flaky~\cite{luo2014empirical}. A test can be flaky for many reasons. Some tests wait on a network connection, which may be slow on a given day. Some compete with other processes for the disk or the memory of the machine. Others depend on the order in which the tests run, on the version of a library, or on permissions that differ from one machine to the next. We did not trace the cause of each flaky test in our suite. What matters here is their effect: a gate cannot ask for zero failures. Our harness read $N$ once, on clean code, and passed any tree with at most three times that many failures.

\paragraph{The whole suite caught as many defects as the file's own tests.} One would expect the whole suite, which runs every test of the repository, to catch more defects than the few tests of one file. It did not. Each gate met eight planted defects (\S\ref{sec:extended}, Table~\ref{tab:gates}). The file's own tests stopped three of the eight, and the whole suite also stopped three. The trees that landed tell the same story: 40 per cent of them carried a defect when there was no gate, 29 per cent after the file's own tests, and 31 per cent after the whole suite. Sixty trials cannot tell the two gates apart (\S\ref{sec:m-thermo}), but the larger gate showed no advantage.

\paragraph{The threshold was inside the noise.} The reason is the threshold. Take the machine whose clean run failed 11 tests. Its gate passed any tree with up to 33 failures. With a planted defect the suite failed 8 to 12 tests on that machine, so every defect passed. A scale that wavers by 30 grams cannot weigh a pinch of salt, and this gate could not see one defect. The noise also moved from day to day. The three machines failed 17, 11 and 1 tests on one morning, and 2, 1 and 4 on the next, so one reading of $N$ says little about the next.

\paragraph{Calibrating the threshold did not rescue the gate.} An experimentalist would object that one reading is no calibration, and would be right. We therefore measured the noise with more care. We ran the clean suite twelve times in a row on each of two machines, and we set the threshold so that clean code is refused about 5 per cent of the time. The new thresholds were 23 and 10 failed tests. The gate did not improve. Under the new thresholds the whole suite stopped two of its eight defects, and 35 per cent of the trees that it let land carried a defect, against 29 per cent for the file's own tests. We tried once more on the third machine, with a defect in every trial, and the gate stopped four of eleven.

We conclude that a test suite with noise of its own is a weak gate. With either threshold, the whole suite stopped no more defects than the tests of the changed file. This suggests that the remedy lies in the tests themselves, in removing those that fail on clean code, more than in the threshold.

Now that we have understood the flakiness of the gate as a complex system, in which the test suite interacts with itself and with other factors of the computer on which it runs, we have to remember that gates are code too. Therefore, they can have errors of their own. The next section explores the effect of the errors of the gates themselves.

\subsection{A gate that errs lets broken changes land}
\label{sec:mistake}
\begin{figure*}[!tp]\centering\IfFileExists{../figures/heavy_cost_of_a_mistake_letter.pdf}{\includegraphics[width=0.9\textwidth]{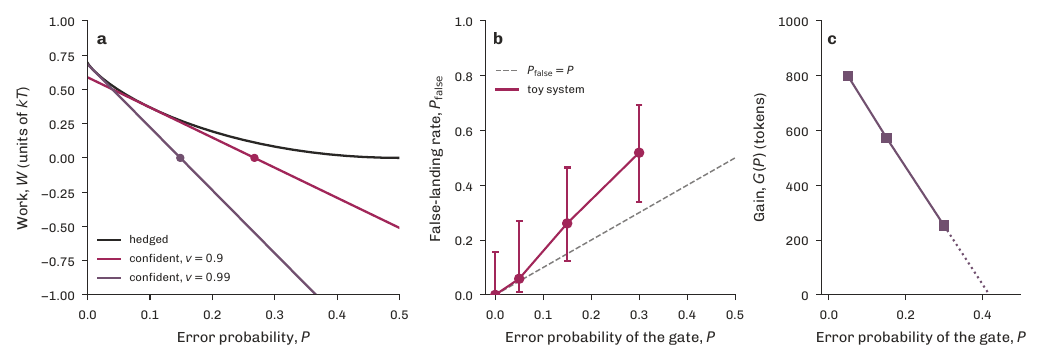}}{\includegraphics[width=0.9\textwidth]{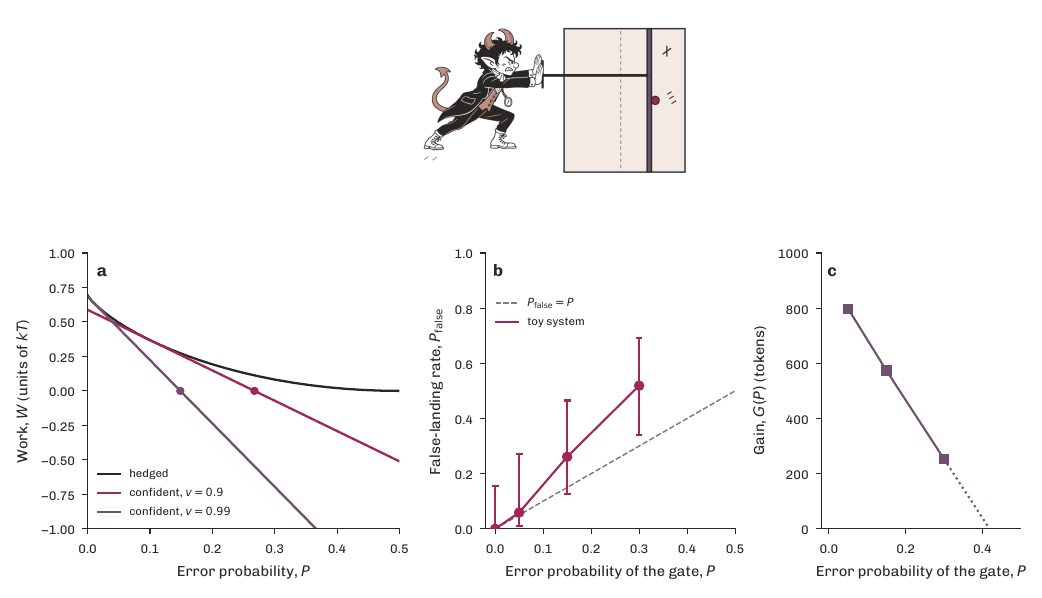}}

\caption{The heavy cost of a mistake. (a) The classical case: Szilard's engine with a faulty measurement. The work that the daemon extracts, in units of $k_BT$, against the error probability $P$ of its measurement. A hedged daemon, which acts only as far as its measurement can be trusted, never loses (black). A confident daemon, which acts as if it were certain, loses work once the error passes a threshold (rose and plum, for two degrees of confidence, $v = 0.9$ and $0.99$; the dots mark where the work crosses zero). (b) Measured on the toy system with its hidden defect: the share of the landed changes that are broken, $P_{\rm false}$, against the error probability that we injected into the gate. From an error of 0.15 up the points lie above the dashed diagonal, on which $P_{\rm false} = P$: broken changes landed more often than the gate erred. The bars are Wilson 95 per cent intervals. (c) The gain of Eq.~\ref{eq:gain} on the same system. It falls as the error rises and stays positive over the measured range. The dotted line is an extrapolation, and we do not use its values. The counts are in \S\ref{sec:m-daemon}.}
\label{fig:mistake}\end{figure*}
We ended the last section on the remark that gates are code, and that code has errors. A gate can err in either direction. It can show green for a broken change or red for a sound one. In the language of \S\ref{sec:intro}, it is a daemon whose measurement sometimes errs~\cite{sagawa2010generalized}. One might think that a gate that errs once in ten lets one broken change in ten land. Physicians know that this is wrong. A test that errs once in ten, used on a ward where nine patients in ten are ill, declares healthy a group of which half are ill. What a wrong verdict costs depends on how many of the attempts are truly broken. In this section we give the gate an error that we control, and we measure three things: the tokens that the error wastes, the share of the landings that are broken, and the temperature that this share gives the gate.

\paragraph{The experiment.} We injected a known error probability $P$ into the gate. Before a verdict was shown to the agent, we flipped it with probability $P$, from green to red or from red to green, and we kept the true verdict beside it. The verdict that the agent sees is the served verdict (\S\ref{sec:defgate}). A false landing is a change that landed on a served green while its true verdict was red. We tested three systems that differed in how often an attempt failed before we flipped its verdict.

\paragraph{When every attempt is sound, the error wastes tokens.} On the first system the agents finished green on every attempt, so a false landing was never possible. The error could only turn a true green into a served red. Of 96 true greens, 7, 12 and 29 were lost in this way at the three errors that we tested, the largest of which was $P = 0.30$. The agents had to try again, and the tokens spent for each true landing rose from about 1{,}700 with no error to about 2{,}400 at $P = 0.30$ (\S\ref{sec:m-daemon}). An error of the gate therefore has a cost even when no broken change can land.

\paragraph{When most attempts are broken, false landings exceed the error.} The second system is a toy that we built with a hidden defect, so that more than half of the honest attempts end red. It is the ward of our opening. With no error, none of 21 landings was false. At $P = 0.30$, 14 of 27 were, which is 52 per cent. From $P = 0.15$ on, the share of false landings was above the error itself (Fig.~\ref{fig:mistake}b). Equation~\ref{eq:pfalse} says why. The share of false landings exceeds the error once more than half of the attempts are truly red, and on the toy 56 to 63 per cent were. We built the toy for this regime. The result is therefore the arithmetic of Eq.~\ref{eq:pfalse} met on a running agent, and it is no general property of agents (Supplementary Note~2).

\paragraph{On the repository, false landings follow the error.} The third system is the repository of this letter. We took the two tasks of \S\ref{sec:cycle} and planted a defect of one line in each, so that a true red existed. We then injected seven errors, from 0.01 to 0.30. Of 669 runs, 550 could be judged, which means that their true verdict could be read (\S\ref{sec:m-daemon}), and 269 of these landed. Here 42 to 60 per cent of the attempts were truly red, on both sides of one half. The share of false landings followed the error: it was 3, 0, 3, 11, 13, 26 and 28 per cent at errors of 1, 2, 5, 10, 15, 20 and 30 per cent (\S\ref{sec:extended}, Table~\ref{tab:verdicts}). For each unit of error added, the share rose by about one unit. The slope was 1.03, with a 90 per cent interval of 0.70 to 1.36.

Equation~\ref{eq:pfalse} therefore describes both systems. The share of false landings is above the error on the toy, where more than half of the attempts are truly red, and close to the error on the repository, where about half are.

\paragraph{The gate has a temperature, and it rises with the error.} In a system held by a bath, a fluctuation that costs an energy $\Delta$ happens with a probability of $e^{-\beta\Delta}$. We read the gate in the same way. A false landing is a fluctuation of the gate that lets a defect through, and the defect costs $\Delta$ tokens to repair. We take $P_{\rm false}$ as the share of the landed changes that are broken, the number that a developer who trusts a landing needs (\S\ref{sec:m-daemon}). The temperature of the gate is the temperature at which such a fluctuation is as frequent as the false landings:
\begin{equation}
\frac{1}{\beta_{\rm gate}} = \frac{\Delta}{-\ln P_{\rm false}} .
\label{eq:gatetemp}
\end{equation}
A gate that almost never errs is cold, and a gate that errs often is hot.

We measured $\Delta$ on the repository. All 36 repairs of the planted defect reached green, at a median of about 1{,}100 tokens (\S\ref{sec:m-daemon}). The temperature of the gate was then about 490, 510, 780 and 830 tokens a nat at errors of 0.10, 0.15, 0.20 and 0.30. It rose with the error. The model, on the same repository, had a temperature of 70 to 147 tokens a nat on Task A and of 51 to 80 on Task B, in the form that counts every token of a run (\S\ref{sec:free}). From an error of 0.10 up, the gate was therefore 3 to 16 times hotter than the model, when the temperature of the model is taken from its landed runs. The gate still removed defects at its hottest: at an error of 0.30, 35 of the 84 judged runs were broken, against 14 of the 50 that landed. The comparison is one of two scales in tokens, and we did not measure a flow of entropy. In the two-reservoir reading the gate should be the colder of the two, and by Clausius's version of the second law (\S\ref{sec:intro}) a gate hotter than the model cannot extract disorder from what the model writes.

\paragraph{What our sample can and cannot show.} A cold gate errs rarely, and a rare event needs many trials to be seen. Repairing a defect costs about a thousand tokens, seven times the highest temperature of the model. A gate as cold as the model would therefore let a broken change through about once in a thousand attempts, and our conditions hold 34 to 50 landings each. With $n$ landings, the lowest temperature that can be read is $\Delta/\ln n$, which is 270 to 299 tokens a nat here, about twice the highest temperature of the model. Our experiment can show a gate hotter than the model. It cannot show one as cold. At the three lowest errors, 0.01, 0.02 and 0.05, there were 1, 0 and 1 false landings among 37, 34 and 39 landings. One false landing sits on this bound, so these conditions cannot be told apart, and we report the condition at 0.05 as an upper limit of 415 tokens a nat.

We also did not measure the temperature of the gate with no error injected. The experiment takes the verdict of the unperturbed suite as the truth, so it cannot see the errors of the suite itself. What we measured is the direction: from an error of 0.10 up, each increase of the error made the gate hotter.

\paragraph{A verdict with an error is still worth having.} We scored the gate with the gain of Eq.~\ref{eq:gain}: what a true green is worth, less what a false landing destroys, each weighted by the error. The gain was positive at every error that we tested (Fig.~\ref{fig:mistake}c). It fell as the error rose, from about 800 tokens to about 250, and it had not reached zero at our largest error, 0.30. The dotted line of Fig.~\ref{fig:mistake}c continues the trend beyond that error as a guide to the eye, and we do not rely on it (\S\ref{sec:m-daemon}).

Therefore, we conclude that an error in a gate costs more than its rate suggests. Sound changes are refused, and their tokens are wasted. Broken changes land. The gate also runs hotter than the model. As agents write and test more of the world's code, we expect gate errors to deserve as much attention as model quality, if not more.

\section{Discussion}\label{sec:discussion}
The Results answered the question of our title. Here we step back and ask what the answers mean, where they may fail, and what they suggest. We do so through six questions that a reader may well put to us.

\paragraph{Is this thermodynamics, or only an analogy?} Thermodynamics has been lent to many fields, and not always with profit. Our answer is that this is an analogy that makes predictions, and that one of them could have failed. We estimated the free energy of a change in five ways that rest on different assumptions, and the estimates agreed within 6 per cent on Task A and within 17 per cent on Task B (Table~\ref{tab:freetests}). An analogy with no content would have had no reason to agree with itself.

Two things keep us from calling it more than an analogy. First, the agent is not in equilibrium with anything. The model proposes without waiting for the gate, a run keeps its whole history, and no detailed balance holds. Our temperatures are therefore operational, and we use Jarzynski's equality by analogy (\S\ref{sec:free}). Second, we have not shown that the free energy is a state function, a quantity fixed by the two end trees however the change is made. The coding and the uncoding estimates lie within 17 tokens of each other on Task A and 21 to 39 tokens apart on Task B. The estimate of Task B rises with the Hinton temperature, and the same task cost different tokens on different days. Uncoding from the ensemble of end trees did not settle the question: the work of the two directions was consistent with Crooks' relation on Task B and was not on Task A (\S\ref{sec:endtrees}).

The title of this letter also promises entropy, and here it has a plain meaning: one change can be made in many ways. The coding runs of Task A ended on 128 different trees, and the uncoding runs on 23 (Fig.~\ref{fig:flow}). A complete account would count these trees in the free energy. We have not done so here, but we plan to investigate this issue further.

\paragraph{How efficient is a coding agent?} Engineers ask of every engine how much of its fuel becomes useful work. The same question can now be put to a harness. The free energy of a change is our estimate of the least that the change can cost on average, and it was 14 to 22 per cent of the cost of a landed coding run. We call this ratio the efficiency of the run. It is a score and no bound of Carnot's kind, and its unit, the token, is shared by every harness and every model. Benchmarks of coding agents report how often an agent succeeds~\cite{jimenez2024swebench}. The efficiency adds what a success cost, set against the least that it could have cost. One could have agents that are efficient in the sense of this letter and inefficient in tokens per watt~\cite{samsi2023words}, or agents that are inefficient in the sense of this letter and very inexpensive in computation. How the other measures of efficiency in the literature correlate with ours, and how this varies across combinations of agent and harness, is a question that future studies from us and others should settle.

\paragraph{Is the heat wasted?} About 80 per cent of the tokens of a landed run left the repository unchanged, and it is tempting to call them waste. We think that most of them are not. In those turns the agent reads files, searches the repository and runs the tests. It is finding out what to change. Szilard and Landauer showed that information has a thermodynamic price~\cite{szilard1929entropieverminderung,landauer1961}, and the heat of a coding run may be, for the most part, the price of the information that the agent needs before it can act. An agent that knew the repository in advance would pay less. Savings are therefore more likely to come from telling the agent what it would otherwise have to discover than from making it read less. We did not measure how much of this heat provided useful information and how much was wasted. We plan to distinguish the two in our next measurement.

\paragraph{Is a colder agent a safer one?} One would expect a colder agent, which gives its most likely reply every time, to be the more reliable one. Our agents were not. Below $T_{\rm H} = 0.5$ they paid the same price for a landing and lost most of their tokens to loops. Counted over all runs, a coding landing cost 2 to 14 times more below $T_{\rm H} = 0.5$ than from 0.5 up. We cannot name a best temperature from one model and two tasks, and a warmer agent may fail in ways that a renaming does not reveal. The loops also point to a remedy other than heat. An agent in a loop receives the same output at every call. What it lacks is information, and we will report on this elsewhere.

\paragraph{What will limit agentic coding, the model or the gate?} Most of the attention in this field goes to the model. Our measurements say that the gate deserves a share of it. Once the gate erred one time in ten, it was 3 to 16 times hotter than the model: the judge was noisier than the author. We injected that error ourselves, and we did not measure the temperature of a test suite as it stands, so we cannot say how hot the gates in use are. That measurement can be made, with an independent reference verdict and enough runs to see one false landing in a thousand, and we think that every harness should make it on its own tests. Therefore, if models improve faster than tests do, test suites will decide the quality of what lands and, eventually, the quality of the software that serves a large part of the internet. More attention to testing is probably deserved, both to the tests of single units and to harnesses built for testing alone.

\paragraph{How far do two renamings reach?} We chose our two tasks because they can be undone exactly, as a molecule can be folded again. Most changes to software cannot. A new feature has no unique reverse, and for such a change the free energy would have to come from the coding runs alone, through the tests of Table~\ref{tab:freetests} that need one direction. Our numbers also come from one model, one repository and one toy system. The extra cost of coding was measured on the runs that landed. Counting the runs that did not land made it larger at the middle Hinton temperature, and at the coldest the matter is open, because the unfinished runs hold most of the tokens.

We therefore state what we expect to hold elsewhere, so that it can be tested. Coding costs more than uncoding. Most of the tokens of a landed run are heat. Cold agents lose tokens to loops. The share of false landings follows Eq.~\ref{eq:pfalse}. The potential energy surfaces of molecules and materials have puzzled scientists for more than a century. We believe that the non-equilibrium kinetic networks of software, which we are only starting to understand through software thermodynamics, will be equally challenging to understand, and will hold as many surprises as their molecular counterparts have shown.

\section{Methods}
\label{sec:methods}
\subsection{System}\label{sec:m-system}
The agent was a language model editing a frozen repository toward a stated goal, one call at a time, through an
instrumentation layer that recorded every call's tokens, every edit and every verdict served. A run was one attempt; a
task was a goal with its reverse; a set of runs shared a model version and a Hinton temperature. A tree's reaction coordinate
$q$ (Eq.~\ref{eq:q}) was the share of the goal's blocks it contained. A block was the renaming of a file or one contiguous group of changed lines, which git
calls a hunk. Task A had $N = 2$ blocks, the renaming of the module's file and one edit to a file that refers to it, so $q$ took the values 0, 1/2 and 1. Task B had $N = 3$ blocks, the renaming and two edits, and $q$ took the values 0, 1/3, 2/3 and 1. A coding run reached its goal at $q = 1$, an uncoding run at $q = 0$, and a run landed when the gate was green there. The gate was the
repository's test suite or a stated subset, served after an edit and once more as the final grade.

Task A and Task B each renamed one module of the repository's own code (Table~\ref{tab:dictionary}), and each was applied once by a script so that the reverse is the exact inverse of the forward. In each task, a file with a single purpose was given a new name in the same folder and every reference to it was updated. The tests that exercised the file served as the gate. The coding (forward) run started from the tree before the renaming and was asked to make it; the uncoding (reverse) run started from the tree after and was asked to undo it. Task A renamed a module that keeps the agent from stalling: when a check takes minutes, such as
the full test suite, it sends the check to the background so that work goes on while it runs. Task B renamed a
module that guards a deployment: it reads back the software image a server is running and confirms that it is the
one that was built.

Each direction of a task was posed as a base
tree, a target tree and a goal in words; each run ended in one of four recorded outcomes: landed; done with the goal unreached; censored at the limit on turns or the wall-time limit; or ungraded, when the grader could not run the tests on the run's last tree. Every landing rate is a share of the runs started. Of the 1{,}200 runs of Table~\ref{tab:cells} (\S\ref{sec:extended}), 1{,}110 landed, 11 ended done with the goal unreached, 78 were censored at the limit on turns and 1 was ungraded. The cost of a run counted the tokens the model generated, which the provider bills as output tokens; the tokens it read were recorded and are not part of the cost.
The repository was a Python codebase of several hundred modules with its test suite, maintained by the authors, frozen
at a stated commit for each set.

\subsection{Model, harness and machines}\label{sec:m-model}
The model was \texttt{deepseek-v4.1-flash}, reached through the commercial gateway Lyceum. Each reply carried the version that the provider served, and on every reply of the seven sets of Table~\ref{tab:cells} (\S\ref{sec:extended}) it was \texttt{deepseek-v4.1-flash-1017}. Each request set the Hinton temperature and switched the model's reasoning mode off. We set no other decoding parameter and no sampling seed, so the provider's defaults applied. A reply could hold at most 65{,}536 tokens, and a run ended at 300 turns. The system prompt was `You change the repository in the working tree with the bash and edit tools. When the task is done, reply DONE with no tool call.' The model had two tools, a shell and a file editor. When a reply called no tool and did not say DONE, the harness answered `If the task is done, reply DONE. Otherwise continue with a tool call.'

The provider ran the model and nothing else. The harness, the working tree, the two tools, the test suite and the grader ran on rented cloud machines, up to eight runs at a time on each. The runs of Fig.~\ref{fig:tokens} used five machines of two cloud providers and two hardware plans, three in Finland and two in Canada. For each set, we recorded the harness commit, the task hash and a seed. The seed fixed the order of the runs and the choice of planted defects, and it did not reach the model.

\subsection{Units and conventions}\label{sec:m-units}
The Hinton temperature $T_{\rm H}$ is the provider's decoding temperature, a dimensionless knob that we set. The software temperatures, $1/\beta_{\rm cyc}$ of the cost around a cycle and $1/\beta$ of the work, are
measured, in tokens per nat. No conversion between the two is assumed. In Fig.~\ref{fig:softT} the vertical axis is the software temperature and the horizontal axis is the Hinton temperature.
Logarithms are natural: an entropy, or a free energy divided by a software temperature, is quoted in nats, the unit in which the
exponents of the fluctuation relations carry no conversion factor; bits are kept for counted information, the bit that a verdict carries at most among them, and one bit is $\ln 2 = 0.693$ nats, so that erasing it costs at least $k_BT\ln 2$ of
heat (Landauer). Table~\ref{tab:intervals} (\S\ref{sec:extended}) gives the kind, the level and the resampled unit of every interval we report. Each bootstrap resampled the runs of one condition and one direction; the repair cost resampled the planted defects first and then the runs of each defect.

\subsection{Unfinished runs}\label{sec:m-unfinished}
Of the 1{,}200 runs, 90 were unfinished. The limit of 300 turns stopped 78 of them, 11 ended by themselves before the tests passed, and one was ungraded: the grader could not run the tests on its last tree. The 78 stopped runs generated 3{,}836{,}090 of the 5{,}846{,}729 tokens of the 1{,}200 runs, 65.6 per cent.

Of the 78 stopped runs, 75 repeated one or two search commands that read the repository and returned the same output each time. All 75 were at $T_{\rm H}$ below 0.5, and none of them changed the repository at any turn. The last new output of such a run came between its 5th and its 31st turn, at the 11th at the median. One further run added a filter to its search at each turn. Two runs at $T_{\rm H} = 0.75$ waited for a test job that never finished. By comparison, a landed run needed 12 turns at the median and at most 50. Among 215 landed runs, the longest streak of repeated commands was 0 in 211, 1 in 3 and 2 in 1.

In the six conditions of the cycle, 25 of the 288 runs were unfinished, and 22 of the 25 were coding runs. The limit stopped 19 of them, all at $T_{\rm H} = 0.1$, and the other 6 ended by themselves before the tests passed. Among the 765 coding runs of Fig.~\ref{fig:trapped}, the share that ended trapped was 35, 32 and 13 per cent on Task A and 25, 13 and 12 per cent on Task B at $T_{\rm H}$ of 0.01, 0.1 and 0.25, and it was zero from 0.5 to 1. With all tokens counted, the landing rate of the coding runs was 12.6 times lower at $T_{\rm H} = 0.01$ than at 1 on Task A and 4.4 times lower on Task B. With the tokens of the last loop left out, it was 2.3 times lower on Task A. With every run cut at 40 turns, it was 2.6 times lower on Task A and 1.6 times lower on Task B.

\subsection{Cost and the cycle temperature}\label{sec:m-cost}
The cost $C$ of a run is its count of generated tokens, a quantity every call reports and a provider bills. A task pairs
a goal with its reverse; coding runs start from the base tree and add the goal's blocks, uncoding runs start from the
target tree and remove them. This construction closes the goal-block coordinate and leaves the
byte-exact tree open, since landed trees carry edits outside the goal's blocks, and we define the cycle's
cost as $Q = \langle C_F\rangle + \langle C_R\rangle$ over landed runs, assuming zero free-energy change on that
coordinate. The asymmetry $\mathcal{A} = \langle C_F\rangle -
\langle C_R\rangle$ is the excess of coding over uncoding. The cycle has one temperature, the model's, and we define it, with no shape fitted, by the relation Gaussian work obeys under the
fluctuation theorems~\cite{jarzynski1997nonequilibrium,crooks1999entropy}, dissipation equal to $\beta\sigma^2/2$:
around the closed cycle, with the mean and the variance $\sigma_{C}^2$ of each direction's cost,
\begin{equation}
\frac{1}{\beta_{\rm cyc}} = \frac{\sigma_{C,F}^2 + \sigma_{C,R}^2}{2\,(\langle C_F\rangle + \langle C_R\rangle)}
\qquad\text{(tokens per nat)}.
\label{eq:betacyc}
\end{equation}

For a process whose work distribution is Gaussian with mean $\langle W\rangle$ and variance $\sigma^2$, the
Jarzynski equality~\cite{jarzynski1997nonequilibrium} $\langle e^{-\beta W}\rangle = e^{-\beta\Delta F}$ evaluates
to $\langle W\rangle - \Delta F = \beta\sigma^2/2$: the dissipated work is fixed by the fluctuations.
Equation~\ref{eq:betacyc} applies this relation to the cost of the cycle, taking the whole cost as dissipated
and $\Delta F = 0$ around the closed cycle. The cost is not Gaussian and the model is not a bath, so the relation
is taken as the definition of the operational temperature. In \S\ref{sec:free} we apply the same relation to
the signed work of Eq.~\ref{eq:work}, one direction at a time (Eq.~\ref{eq:betawork}). The two temperatures differ
because the cost contains the heat and the work does not.

The two tasks (\S\ref{sec:m-system}) were run in both directions, coding (forward) and uncoding (reverse), on four fully instrumented sets (two at temperatures 0.1 and 1.0, two at
0.5), 24 runs launched a direction and condition, 14 to 24 landed, 263 landed runs in all. $Q$ and $\mathcal{A}$ were means
over landed runs; $\mathcal{A}$ carried a 97.5 per cent normal interval, on the standard error of a difference of two means with unequal variances, and a symmetry margin of a tenth of the mean cost. $1/\beta_{\rm cyc}$ was Eq.~\ref{eq:betacyc} per condition. In Fig.~\ref{fig:softT} it is computed under the three approximations on the twelve conditions, and its interval is a 95 per cent percentile bootstrap of 2{,}000 draws over runs, each direction resampled on its own. Where all runs landed, on Task A at $T_{\rm H} = 0.5$ and $T_{\rm H} = 1$, the cycle heat was 3{,}999 and 3{,}670 tokens under all three approximations. On Task A at $T_{\rm H} = 0.1$ it was 4{,}158 tokens under VIVA, 60{,}598 under CORTA and 556{,}271 under LOCA, and on Task B at the same Hinton temperature it was 2{,}745, 17{,}091 and 254{,}492 tokens. 

To test for survivor selection, we ran a separate sample in the two cooler conditions of each task. We pooled runs with and without the harness. We computed the asymmetry first over landed runs and then over all runs, including the tokens spent by unfinished runs. The sample holds twenty runs a direction at the middle Hinton temperature and nine to fourteen at the coldest. At the middle Hinton temperature all runs of the Task A sample landed, and the asymmetry was $+371$ $[+27, +715]$ tokens. That of Task B was $+408$ $[+63, +753]$ on landed runs and $+588$ $[+163, +1013]$ on all runs. At the coldest Hinton temperature, 22 per cent of the runs reached the limit of 300 turns. These ten runs cost 27,000 to 365,000 tokens, compared with a median near 1,800 among landed runs.

\subsection{The force--extension curves of Fig.~\ref{fig:cycle}a}\label{sec:m-curves}
We read the curves of Fig.~\ref{fig:cycle}a from the pixels of Fig.~1 of ref.~\cite{collin2005verification}. The tick marks of its axes gave the scale, 24 pixels per pN and 8.8 pixels per nm. We assigned each coloured pixel to an unfolding or a refolding curve by its colour and left out the drawing of the hairpin. The shaded area is the integral of the force over the extension under the refolding curve that bounds the shaded area of the original, from 355.0 to 380.3 nm and down to zero force. It is 392 pN nm, or 95 $k_BT$ at 298 K. This area belongs to one curve between those two extensions. The mean refolding work of 106.6 $k_BT$ in ref.~\cite{collin2005verification} is taken between the limits of that analysis. As a control we drew a trapezoid of known area in the same colours and read it by the same procedure, which returned the area within 0.2 per cent.

\subsection{The arrow of time of a run}\label{sec:m-arrow}
We read the arrow of time of a run from its output tokens per turn, following the estimates of entropy production from the irreversibility of a recorded series~\cite{parrondo2009entropy,roldan2012kl}. Each landed run gave two numbers that change sign when the run is read backwards: the mean change in tokens from one turn to the next, divided by the run's mean tokens per turn, and the number of rises minus the number of falls, divided by the number of steps. We fitted a logistic classifier without an intercept to tell the runs in recorded order from the same runs reversed, so that its score $f$ changes sign under reversal. On runs kept out of the fit, the Donsker--Varadhan expression $\langle f\rangle - \ln\langle e^{-f}\rangle$ estimates a lower bound on the relative entropy between the recorded and the reversed records, in nats a run. The bound holds for the means over all runs. On a finite sample the expression is an estimate, and it can fall below zero.

We used the 633 landed runs of three of the seven sets of runs behind Table~\ref{tab:cells} (\S\ref{sec:extended}), 326 for the fit and 307 for the estimate, divided by a hash of each run's name. The estimates of \S\ref{sec:engine} are on 55 and 90 of these runs for coding and uncoding on Task A and on 82 and 80 for Task B. They were 0.28 nats a run $[0.02, 0.55]$ for coding on Task A and 0.39 $[0.17, 0.60]$ for uncoding. On Task B they were 0.03 $[-0.18, 0.24]$ and $-0.03$ $[-0.28, 0.22]$. Intervals are 95 per cent bootstraps over the runs of the estimate and leave out the uncertainty of the fit. With the turns of each run shuffled, the estimate was $0.00$ $[-0.03, 0.03]$. The fit did not see the runs of the six cycle conditions; their coding estimates were 0.25 to 0.30 nats on Task A and $-0.06$ to $0.04$ on Task B, on 14 to 24 runs a condition. In the three cycle conditions of Task A the coding estimate, multiplied by the software temperature of the condition, was 19 to 39 tokens, 0.5 to 1.1 per cent of the cycle heat.

\subsection{The free energy from the work}\label{sec:m-free}
\paragraph{The runs.} The two tasks were run in both directions, coding (forward) and uncoding (reverse), by the plain agent at six Hinton temperatures, 0.01, 0.1, 0.25, 0.5, 0.75 and
1, in seven sets on different days. We combined every set and left out one run in which a reply reached the provider's limit of 65{,}536 tokens. This gave 1{,}200 runs, of which 1{,}110 landed (\S\ref{sec:extended}, Table~\ref{tab:cells}). The run that was left out landed, on Task A at $T_{\rm H} = 0.25$ in coding, after 74{,}298 tokens. With it, the mean cost of a landed coding run in that condition is 3{,}904 tokens in place of 2{,}374 and its squared coefficient of variation is 7.3 in place of 0.08, while the free energy changes by 0.1 token.

\paragraph{Work and heat.} The state $q$ after each turn was read from the saved tree, and the work and the heat of a
landed run followed from Eq.~\ref{eq:work}. We included unfinished runs when computing the landed fraction and the unfinished share. We computed mean work from landed runs alone. The mean work of a landed uncoding run was $-288$ to $-316$ tokens on Task A and $-194$ to $-261$ on Task B. The work left over after a cycle, $\langle W_F\rangle + \langle W_R\rangle$, was 91 to 153 tokens on Task A and 150 to 201 on Task B.

\paragraph{The software temperature.} The software temperature of Eq.~\ref{eq:betawork} ranged from 128 to 153 tokens per nat on Task A and from 49 to 90 on Task B (\S\ref{sec:extended}, Table~\ref{tab:cells}). On the same runs the cost form of Eq.~\ref{eq:betacyc} gives 70 to 147 tokens per nat on Task A and 51 to 80 on Task B.

\paragraph{The free energy.} The free energy $\Delta F_{J}$ of the coding runs was 342 to 368 tokens on Task A and 271 to 339 on Task B (\S\ref{sec:extended}, Table~\ref{tab:cells}). On Task B it was 275 $[253, 298]$ tokens at $T_{\rm H} = 0.1$ and 339 $[318, 362]$ at $T_{\rm H} = 1$. In Task A at $T_{\rm H} = 0.5$, 351 of the 415 tokens of coding work were the free energy and 64 were dissipated work, about 0.4 nats at the temperature of that condition.

\paragraph{The free energy as a fraction of the cost.} The free energy as a fraction of the cost of a landed run is
$\Delta F_{J}$ of the coding runs over the mean of the two directions' mean costs. It is 17 to 19 per cent on Task A and 19 to 25 per cent on Task B in the conditions that are not left out. Over the mean cost of a landed
coding run alone it is 15 to 17 per cent on Task A and 17 to 22 per cent on Task B, and over that of a landed
uncoding run it is 19 to 24 and 22 to 28 per cent.

\paragraph{The heat of a landed run.} At $T_{\rm H} = 0.5$ a landed coding run of Task A cost 2{,}305 tokens on average and a landed uncoding run cost 1{,}710. The turns that changed the state account for 415 and 288 of those tokens, and the other 1{,}890 and 1{,}422 tokens, 82 and 83 per cent, were heat. On Task B the two directions cost 1{,}606 and 1{,}205 tokens, of which 410 and 227 were work and 74 and 81 per cent were heat (Supplementary Fig.~\ref{fig:cycleB}d).

\paragraph{The estimators.} Per condition, $\beta$ was Eq.~\ref{eq:betawork} with the sample variances of the two directions; the free energy was
$\Delta F_{J} = -(1/\beta)\ln\langle e^{-\beta W_F}\rangle$ over the landed coding runs, its reverse form
$(1/\beta)\ln\langle e^{-\beta W_R}\rangle$ over the landed uncoding runs, and $\Delta F_{S}$ of
Eq.~\ref{eq:symmetric}. In the retained conditions, the uncoding estimate less the coding one was between $-17$ and $+15$ tokens on Task A and between $+21$ and $+39$ on Task B. $\Delta F_{S}$ lay within 1.4 tokens of $\Delta F_{J}$ in the conditions of Task A and within 23 tokens in the retained conditions of Task B. In the cold condition of Task B that we leave out, $T_{\rm H} = 0.01$, the two differ by 51 tokens (\S\ref{sec:extended}, Table~\ref{tab:cells}). The variance of the coding work was 1.2 to 1.9 times that of the uncoding work on Task A and 1.1 to 3.2 times on Task B.

\paragraph{Intervals.} Intervals are 90 per cent percentile bootstraps of 2{,}000 draws over runs, each direction
resampled on its own and $\beta$ fitted again in every draw. The shift for few runs was the mean of $\Delta F_{J}$
over 400 random subsets of 5, 10 and 20 landed coding runs, less its value on all of them, at the condition's $\beta$.

\paragraph{The crossing.} The crossing was the work, between the two mean works, at which Gaussian kernel density estimates of $W_F$ and of $-W_R$ were equal, the middle one when there were several. Its interval is the 90 per cent percentile bootstrap of 1{,}000 draws over runs, each direction resampled on its own. The bandwidth of each density was set by Scott's rule, and a crossing existed in all twelve conditions. In 0.4 to 24 per cent of the draws on Task A, and in at most 0.5 per cent on Task B, the two estimates did not cross between the mean works, and the interval is over the draws that did. The crossing less $\Delta F_{J}$ was between $-16$ and $+22$ tokens on Task A and between $-32$ and $+1$ on Task B without its coldest condition, where the crossing was 49 tokens above $\Delta F_{J}$. On Task B the crossing was 243 tokens at $T_{\rm H} = 0.1$ and 333 at $T_{\rm H} = 1$.

\paragraph{Estimators for work that is not Gaussian.} The mixture of two Gaussians was fitted to the landed coding work by expectation maximisation, and its free energy is the exponential average of the fitted density at the $\beta$ of the condition. It lay within 5 tokens of $\Delta F_{J}$ on Task A and within 8 on Task B. Bennett's acceptance ratio less $\Delta F_{J}$ was between $-7$ and $-2$ tokens on Task A and between $-12$ and $+10$ on Task B. A single Gaussian was 9 to 19 tokens below $\Delta F_{J}$ on Task A and 4 to 32 below on Task B. The excess kurtosis of the landed coding work was between $-1.2$ and $+0.2$.

\paragraph{The bias for few runs.} The bias for few runs was simulated for Gaussian work, 40{,}000 samples of the condition's number of landed coding runs, with the variance of $\beta W$ set in two ways: to the sample variance, and to the mean squared distance below the median of the runs below the median, which is the Gaussian tail of Palassini and Ritort~\cite{palassini2011improving}. Their parameter $\lambda$, the logarithm of the number of runs over the dissipated work of the tail, was 5 to 16 on Task A and 0.8 to 2.9 on Task B; their closed forms hold for $\lambda$ below 1, so we simulated. At $T_{\rm H} = 0.1$, 0.25, 0.5, 0.75 and 1 on Task B the bias was 23, 17, 12, 4 and 9 tokens with the first variance and 47, 32, 17, 8 and 12 with the second. With the bias removed the free energy of Task B was 251 or 228 tokens at $T_{\rm H} = 0.1$ and 330 or 327 at $T_{\rm H} = 1$. On planted Gaussian work of 48 runs with a dissipated work of $\ln 48$ nats the simulated bias was 0.68 nats, against $\ln 2$ in their scaling limit. Subsets of 20 landed coding runs gave an estimate above that of all the runs by at most 3 tokens on Task A and 7 on Task B.

\paragraph{The free energy and the cost of a landing.} Over the twelve conditions, $\Delta F_{J}$ was 14 to 22 per cent of the mean cost of a landed coding run. With the tokens of the unfinished runs counted (CORTA), a coding landing cost 7{,}798 to 31{,}359 tokens in the conditions below $T_{\rm H} = 0.5$ and 2{,}183 to 3{,}665 from 0.5 up.

\paragraph{The landed fraction.} The term for the landed fraction was $-(1/\beta)\ln(\phi_F/\phi_R)$ with $\phi$ the landed
fraction of each direction~\cite{alemany2012experimental}. It was between 0 and 48 tokens in the twelve conditions. With it, the free energy of Task B was 281 tokens at $T_{\rm H} = 0.1$ and 347 at $T_{\rm H} = 1$. In the two coldest conditions of Task A, the cost of a coding landing under LOCA was 240{,}000 and 329{,}000 tokens, and the free energy rose from 354 to 401 tokens and from 368 to 414. In the nine conditions of the quoted ranges the term was at most 0.15 nats and at most 15 tokens, less than the half width of the 90 per cent interval of the free energy, which was 18 to 35 tokens. In the two coldest conditions of Task A it was 0.30 and 0.40 nats, 46 and 48 tokens, about twice the half width of 22 tokens. The derivation of the term assumes that each state is in equilibrium within itself. An unfinished run here is one stopped by the limit of 300 turns, which a folding molecule does not have.

\paragraph{The two rates.} The ratio of costs was the mean cost of a landed coding
run over that of a landed uncoding run. We give the rate of landing in two ways. The first is the reciprocal of the mean cost of a landed run. A landed coding run cost 1.12 to 1.45 times a landed uncoding run over all twelve conditions, so this rate is lower in coding by 0.11 to 0.37 nats. The second is the number of landed runs divided by all the tokens of the condition, the rate of Fig.~\ref{fig:trapped}, which also counts the tokens of the runs that did not land. In the six conditions from $T_{\rm H} = 0.5$ to $T_{\rm H} = 1$, where no run was trapped, this second rate is lower in coding by 0.30 to 0.48 nats on Task A and by 0.59 to 1.12 nats on Task B. With the first rate the two readings stand 2.0 to 2.5 nats apart on Task A and 3.5 to 5.2 on Task B. With the second rate, in its six conditions, they stand 1.9 to 2.5 nats apart on Task A and 2.6 to 4.7 on Task B.

\paragraph{The tests on the cost.} The mean costs of the landed runs of the two directions were compared in each condition by a two-sided permutation test of 20{,}000 relabellings on the difference of the means. The $p$ value was below 0.001 in 11 of the 12 conditions and 0.053 on Task A at $T_{\rm H} = 0.1$. The shape of the two directions' costs was
compared by a permutation test of 2{,}000 relabellings on the Kolmogorov--Smirnov distance between the costs, each
divided by its direction's mean. Each relabelling pooled the scaled costs of the two directions, dealt them into two groups of the original sizes and divided each group by its own mean again before the distance was taken, as was done for the data. The correction for twelve tests was Bonferroni's, a threshold of 0.05/12. We estimated what the test can detect by simulation with gamma-distributed costs, 48 a direction, the means of the two directions a factor of 1.3 apart, 400 pairs for each case and 500 relabellings a pair. The test rejected at 0.05 in 4 per cent of pairs of equal shape. It rejected in 20 per cent of pairs with squared coefficients of variation of 0.10 and 0.05, in 32 per cent with 0.10 and 0.20, and in 56 per cent with 0.10 and 0.033.

\paragraph{The regularity of the cost and the bound on the states.} We calculated the squared coefficient of variation by dividing the sample variance of landed-run costs by the square of their mean. The bound on the number of states is that of Aldous and Shepp~\cite{aldous1987least} for the time a continuous-time Markov chain on $n$ states takes to reach an absorbing state. Tokens are counted in whole numbers and a run retains its history, so the chain is a model of the run and the bound is read under it. The bound is reached only when the states are visited one after another at equal rates. A state of the model is not a turn. Relations between the regularity of a first-passage time and the dissipation that pays for it are proved for continuous-time Markov processes~\cite{barato2015thermodynamic,gingrich2017fundamental,pal2021thermodynamic}, and we did not apply them to the runs.

\paragraph{Conditions left out.} A condition was left out of the quoted ranges when more than 85 per cent of the tokens of either direction were in
unfinished runs, because in such a condition the landed runs are the subset that stopped early, and the limit on turns shapes what is averaged. The rule removed three conditions, which Table~\ref{tab:cells} (\S\ref{sec:extended}) shows
with the rest. The mean cost of a landed coding run was 1{,}502 to 1{,}732 tokens on Task B and 1{,}986 to 2{,}510 on Task A. In the three conditions that the rule removed, 63 to 75 per cent of the coding runs landed, and they hold 7 to 13 per cent of the coding tokens. With every condition included, the free energy of Task A stays within 342 to 368 tokens and that of Task B spans 260 to 339. The share of the tokens of the coding runs that the unfinished runs took was, from the coldest condition to the warmest, 91, 93, 74, 0, 22 and 0 per cent on Task A and 87, 79, 83, 26, 59 and 44 per cent on Task B.

\paragraph{Checks on planted data.} The estimators were checked on planted data: Gaussian work obeying the work
relations with a known temperature and free energy, which they returned within 2 per cent and 1 token; costs of known
regularity; two samples of one shape with means 1.35 times apart, which the shape test called the same, and two of
different shape, which it told apart; wide Gaussian work, on which the shift for few runs was positive and fell
as runs were added; and Gaussian work of 48 runs a direction obeying Crooks' relation with a free energy of 350 tokens, on which the crossing was 365 $[334, 389]$ tokens.

\subsection{Uncoding from the ensemble of end trees}\label{sec:m-endtrees}
We repeated the uncoding of each task at $T_H=0.5$ from two starts, with the model, the harness, the instruction and the limits of the cycle (\S\ref{sec:m-system}, \S\ref{sec:m-model}). One start was the single tree from which every uncoding run of the cycle started. The other was the ensemble of trees on which the coding runs ended.

\paragraph{Starts.} On Task A the 48 landed coding runs ended on 27 distinct trees, identified as in \S\ref{sec:m-paths}. For each uncoding run of the ensemble, we drew one of the 48 runs with equal probability and with replacement, and rebuilt its end tree from the starting tree and the change that the run had stored. Every rebuilt tree had the hash that the coding run had recorded. The 100 draws gave 24 distinct trees, and the most frequent tree was drawn 39 times. Outside the files of the change, these trees differed from the single tree in at most one file and four lines. We made 100 further uncoding runs from the single tree. The 200 runs were made together, in random order, and all of them landed. On Task B the 45 landed coding runs ended on two distinct trees, which held 33 and 12 of the runs, and both were drawn. These trees differed from the single tree in the files of the change only. We again made 100 runs from each start, together and in random order, and all 200 landed. In one run of Task B the provider was asked twice for one reply; the run is counted with the tokens that were metered.

\paragraph{Cost and work.} Table~\ref{tab:endtrees} (\S\ref{sec:extended}) gives the cost and the work of a landed uncoding run from each start, with 90\% bootstrap intervals over runs (2,000 resamples). The costs of the two starts differed by $-27$ tokens $[-194, 127]$ on Task A and by $-27$ tokens $[-111, 57]$ on Task B. On both tasks the cost from the single tree agrees with the uncoding cost of the cycle (\S\ref{sec:extended}, Table~\ref{tab:cells}).

\paragraph{Crooks' test.} We estimated the density of the work of coding and of the work of uncoding with its sign changed, each with a Gaussian kernel whose width follows Scott's rule. We fitted a straight line to the logarithm of their ratio over the works at which both densities exceeded one tenth of their peaks. Table~\ref{tab:endtrees} (\S\ref{sec:extended}) gives the slope of the line and the work at which it crosses zero. On simulated Gaussian works that obey Eq.~\ref{eq:crooks}, the same procedure returned 94\% of the true slope. We fixed this analysis before we read the runs.

\subsection{Paths and the count of end trees}\label{sec:m-paths}
A run's path is the sequence of trees it passed through. We recorded the tree at the start of the run, after each code change and at the end of the run. We identified each tree by a hash of its files, taken after comments and white space were removed from each file, so two trees that differ only in layout or in comments count as one. The tree includes any files the run added.
We merged consecutive repeats of one tree. Each step in the resulting path represents one change of the code. The end tree of a
run is the last tree of its path. Figure~\ref{fig:flow} pools the six Hinton temperatures and draws the runs of one
task and one direction in one panel.

Among landed runs, we counted distinct end trees and calculated the entropy of the end-tree distribution, $-\sum_i p_i \log_2 p_i$, where $p_i$ is the fraction of landed runs ending on tree $i$. The counts pool the six Hinton temperatures. On Task A the 310 coding runs that landed ended on 125 trees, and the commonest of them held 45 per cent of the runs. The 343 uncoding runs that landed ended on 21 trees, and one tree held 94 per cent. On Task B the landed runs ended on 7 trees in coding and 4 in uncoding. The count is a lower bound, since a tree that no run reached is not counted. The entropy is the plug-in estimate, which reads low on average at a finite number of runs~\cite{paninski2003estimation}. The first-order correction, $(K-1)/(2N\ln 2)$ bits for $K$ trees seen in $N$ runs, is 0.29 bits for the coding runs of Task A and less than 0.05 bits for each of the other three sets. With 114 of 125 trees reached once, this correction is itself too small for the coding runs of Task A, and we report the uncorrected values.

The figure draws every run, landed or not. Its colours rank the end trees by the number of runs that end on them; trees with equal numbers are ranked in the order of the run table, which carries no meaning.
The figure holds more than 1{,}500 runs. For 192 of them the intermediate trees had not been stored. We rebuilt these trees by replaying each run's stored commands on its starting tree, without calling the model. The replay reproduced the recorded tree identifiers at every step in 190 of the 192 runs. The two other runs ended unsolved and are drawn with the trees that could be read. In seven unsolved runs some intermediate trees could not be read, and each such tree is drawn as a tree of its own. The four hues of the figure were checked for separation under
the common forms of colour blindness.

The counts of the four panels are over all the runs of a panel, landed or not. Panel a, the coding runs of Task A, holds 354 runs, of which 310 landed, on 151 different paths that end on 128 different trees; the five commonest end trees hold 138, 42, 16, 15 and 6 runs, and 137 runs end on other trees. Panel b, the coding runs of Task B, holds 411 runs, of which 360 landed, on 55 paths that end on 14 trees, with 233, 139, 27, 2 and 1 runs on the five commonest and 9 on other trees; seven of its runs go on beyond step 6, to between 9 and 65 changes, and are drawn to step 6. Panel c, the uncoding runs of Task A, holds 357 runs, of which 343 landed, on 29 paths that end on 23 trees, with 323, 13, 1, 1 and 1 runs on the five commonest and 18 on other trees. Panel d, the uncoding runs of Task B, holds 410 runs, all of which landed, on 13 paths that end on 4 trees, with 407 runs on one tree and one run on each of the other three. The colours are assigned within a panel and carry no meaning across panels.

\subsection{The map of visited states}\label{sec:m-map}
The map of Fig.~\ref{fig:wells} uses 192 runs of the plain agent for each task, 24 coding and 24 uncoding at each of the Hinton temperatures 0.1, 0.5, 1 and 1.5. At 0.1, 0.5 and 1 these are the runs of the six cycle conditions, which are also among the runs of Table~\ref{tab:cells} (\S\ref{sec:extended}). The temperature 1.5 is not among its conditions. A state is the starting tree
together with the edits the run had made so far. We read the edits from the run's edit commands and rebuilt a file
renamed through the shell as an edit; other changes made through the shell are not read, so a run that changed the
code only in that way shows no movement on the map. We described each edit by the counts of the node types of the
Python syntax tree of the text it added, less those of the text it removed, kept for each file and hashed into 4{,}096
bins. The descriptor of a state is the sum over its edits, with the signed square root taken of each bin. We shifted
the states of the uncoding runs by the mean end state of the landed coding runs, so the two
directions share one map.

The principal components were fitted to all states of a task. The density is a Gaussian
kernel estimate on the first two components. The kernel has one covariance for both tasks, the mean of the two tasks'
covariance matrices multiplied by $n^{-1/3}$, where $n$ is the mean number of states of a task. Each run has a total
weight of one, and the starting state is not counted. The map is the visit potential, $\Phi = -\ln(\rho/\rho_{\max})$, with $\rho_{\max}$ the
highest density of the two tasks.

We chose three basins after seeing the map and assigned each repository state to one of them by $k$-means clustering with three centres on the two components. $\mathcal{M}_1$ is the one that holds the start of the coding runs and $\mathcal{M}_3$ the one in which most landed coding runs ended. A run crossed them in order if it visited all three, ended in the last one of its direction and never moved to one it had left. Of the 64 landed coding runs on Task A, 62 crossed them in order, and 66 of 67 did on Task B. Of the landed uncoding runs, 70 of 74 on Task A and 83 of 83 on Task B crossed the same three in the opposite order. The seven other landed runs stayed in one basin on the map, which reads a run's edit commands and its renames and no other change of a file. One run that did not land, of the 384, went back to an earlier basin.
No barrier height is read from the map. We measured how long a landed run stayed in each basin. The median stay in the first was 10 turns in coding and 7 in uncoding on Task A, and 5 and 4 on Task B. The second was reached in the same turn as the third in every landed run, because the edits of the change come in one reply of the model. The stay of 4 to 5 turns in the third is set by the rule that ends a run shortly after it reaches its goal. No landed run returned to a basin that it had left. The map reads only edits, so the turns in which a run reads or tests do not move it, and we did not measure a rate of jumps.

Panels b and d of Fig.~\ref{fig:wells} draw $\mathcal{M}_3$ on its own two principal components, fitted to the states that the coding runs visited inside it: 114 visits to 13 distinct states on Task A, and 100 visits to 7 distinct states on Task B. The run drawn in panels a and b was in $\mathcal{M}_2$ after one edit, stayed in $\mathcal{M}_3$ from edit 2 to edit 7 and passed through six states there. The run drawn in panels c and d entered $\mathcal{M}_3$ at edit 2 and passed through three states there before it landed at edit 4. Each of the two is the landed run of its task with the most distinct states among those that never returned to an earlier basin. Panels a and c use the same density scale. We calculated the principal components separately for each task.

\subsection{The thermometer}\label{sec:m-thermo}

The daemon reads a verdict on each change, green or red and at most one bit, from a test suite
that on a clean tree fails $N$ tests by its own noise. Our harness read $N$ once and set the gate's threshold at three times that reading. One reading is a number. The instrument is the distribution of $N$, and a threshold is a quantile of that distribution with a
stated false-alarm rate. Among trials that land, a defect survives a gate of false-green probability $P$ with
probability $P$ and a clean trial passes with probability $1-\alpha$, so with defect rate $d$ before the gate the
defect rate among survivors is
\begin{equation}
d' = \frac{dP}{dP + (1-d)(1-\alpha)},
\label{eq:survivor}
\end{equation}
a selection on landing of Berkson's kind~\cite{berkson1946limitations}: it vanishes as $P \to 0$ and equals $d$
at $P = 1$ and $\alpha = 0$.

\paragraph{The trials.} We ran sixty trials with no model, twenty on each of three machines. Each trial drew, with probability one half, a planted
single-site defect in one of six files that the gate reaches. It also drew a gate: none, unit (the file's own tests) or whole.
The whole gate passed a tree whose failure count did not exceed $3N_1$, with $N_1$ from one clean run.

\paragraph{The calibrated threshold.} The calibrated threshold was the 95th percentile of a negative binomial distribution fitted to the failure counts of twelve consecutive clean runs on each of two machines, 23 and 10 failed tests. The third machine had no such runs, and its threshold, 20 failed tests, came from one fit to the counts of the two machines pooled.
The dispersion of a count is its variance over its mean. It is 1 for a Poisson count, which is what failures that are independent and equally likely in every run would give. A likelihood-ratio test rejected the Poisson distribution on both machines ($p < 0.0001$ on each), and we therefore fitted the negative binomial. A defect was caught
when its count exceeded the threshold.

\paragraph{The defects that survived.} Of the trials that landed, the share carrying a defect was 0.40 with no gate (8 of 20), 0.29 with the file's own tests (5 of 17) and 0.31 with the whole suite (5 of 16). With the realised $d = 0.40$ and the measured $P$ of 5 in 8, Eq.~\ref{eq:survivor} gives 0.29. The counts of failures on clean code were overdispersed on both machines, with variance over mean of 3.4 and 4.9, and the set of spuriously failing tests was still growing at the twelfth run. The calibrated thresholds had a false-alarm rate near 0.05. Under them, 6 of the 17 landed trials of the whole suite carried a defect, a share of 0.35.

\paragraph{Twelve further trials.} We ran twelve further trials on the third machine with a defect in every trial, under a threshold of three times the median of three clean readings (17, 3 and 1 failures), 9 failed tests. One trial never ran and eleven produced a failure count. The sixty trials were judged under both thresholds. The association between defect and gate among the 53 trials that landed under the one-reading threshold is the Pearson correlation of defect, 0 or 1, with the gate ranked none, unit, whole. It was $-0.08$, with a 90 per cent bootstrap interval over trials of $-0.31$ to $0.15$. Running the test suites for these trials took about thirty host-hours.

\subsection{The daemon's mistake and the injected error}\label{sec:m-daemon}

Maxwell's daemon, reading one bit and acting on it, extracts at most $k_BT\ln 2$ of
work~\cite{szilard1929entropieverminderung,landauer1961,bennett1982}; one that misreads with probability $P$ and
acts as if certain loses work on every cycle once $P$ passes a threshold set by how far it
acts~\cite{sagawa2010generalized,horowitz2011thermodynamic}. In Fig.~\ref{fig:mistake}a the confident daemon expands the gas to a fraction $v = 0.9$ or $0.99$ of the box as if its measurement were certain, and its work crosses zero at $P = 0.27$ and $0.15$; the hedged daemon stops at $1-P$. Our daemon serving green on a red tree makes the same mistake. Let $s$ be the share of attempts that are truly red. The information that a served verdict carries about the true one is the mutual information of a binary channel that flips its input with probability $P$,
\begin{equation}
I(s,P) = h\bigl(s(1-P) + (1-s)P\bigr) - h(P),
\label{eq:info}
\end{equation}
in bits, with $h$ the binary entropy. It is largest at $s = 1/2$, where it equals $1-h(P)$, and it reaches one bit only when in addition $P = 0$. In our experiments $s$ was 0.56 to 0.63 on the toy system and 0.42 to 0.60 on the repository, so a served verdict carried at least 93 per cent of $1-h(P)$ at every injected error.

\subsubsection{Error injection}

A small program placed between the test suite and the agent flipped each served verdict with probability $P$ by a seeded draw and recorded
the true verdict beside it, about 5.6 verdicts a run; a false landing was a final grade served green whose true
verdict was red. The first system was the repository's own cycle panel, 96 runs at each of $P = 0, 0.05, 0.15, 0.30$. The second was a toy with a hidden defect in a file the visible tests never touch, graded by a hidden test the agent
never sees, 21, 17, 23 and 27 landings at the four rates, of which 0, 1, 6 and 14 were false. We did not measure the cost of a repair on the toy, so we give its gate no temperature. The third was the cycle panel again with a one-line defect
from a fixed list planted by seed, at $P = 0.01, 0.02, 0.05, 0.10, 0.15, 0.20, 0.30$, the full suite on the reduced repository
as the gate. Every served and true verdict was recorded. Each condition held 48 runs with the defect planted in half and 72 further planted runs under a second seed, six
to eight runs at a time on each machine.

\subsubsection{Unjudgeable runs}

We classified runs as unjudged when the limit on turns, a wall-time limit, an undecodable tree or a provider error prevented us from reading the true verdict. This happened in 119 of the 669 planted runs: 79 of the 477 in the five conditions from 0.05 up, and 40 of the 192 in the two conditions at 0.01 and 0.02. Each of the seven conditions was planned with 96 planted runs, 672 in all, and three of them are absent from our records, which leaves 669. Refusals by the provider cut short 56 runs, and five more failed for local reasons. We repeated all 61 under the same plan and seed and excluded the interrupted attempts. An unjudged run counts among the planted runs and enters no false-landing rate. We flagged runs when the host logs showed that load exceeded the number of cores or that the memory limit had killed a process. We analysed the results with and without the flagged runs.

\subsubsection{False-landing rates}

A green is served either because a true green was left
alone or because a true red was flipped, so the false-landing rate among landings is
\begin{equation}
P_{\rm false} = \frac{sP}{(1-s)(1-P) + sP},
\label{eq:pfalse}
\end{equation}
which exceeds $P$ exactly when $s > 1/2$, in a model of one verdict per run. A run of the experiment saw about 5.6 verdicts, so we use the equation as a guide and set it beside the counts in Supplementary Note~2. The false-landing rate per condition is the false landings over the judged planted runs whose final grade was served green, the quantity of Eq.~\ref{eq:pfalse}. Its slope against $P$ is the
least-squares slope over the seven conditions. The interval of the slope is a bootstrap of 4,000 draws that resamples the judged runs within each condition. The slope estimates how much the broken share of the landed changes grows per unit of injected error.

On the first system the tokens spent per true landing rose from 1,712 at $P = 0$ to 2,355 at $P = 0.30$. On the repository the false-landing rate was 1 of 37, 0 of 34, 1 of 39, 4 of 35, 5 of 39, 9 of 35 and 14 of 50 landings, or 0.03, 0, 0.03, 0.11, 0.13, 0.26 and 0.28, at the seven errors. The Wilson 90 per cent interval of each rate contains the error of its condition. The slope of the rate against the error, which we take as the gate's response, is 1.03 false landings per unit error over the seven conditions (90 per cent bootstrap interval $[0.70, 1.36]$ over runs within conditions), and 1.04 $[0.56, 1.50]$ over the five conditions from 0.05 up. Setting aside the runs that carried a load or memory flag leaves 5 to 12 landings a condition and rates of 1 of 9, 0 of 8, 1 of 8, 0 of 8, 1 of 5, 5 of 12 and 5 of 10, rising overall, at sample sizes that fix no slope.

\subsubsection{Gate temperature}

\paragraph{Definition.} In physics, a system at a temperature $T$ makes a fluctuation that costs an energy $\Delta$ with probability $e^{-\Delta/k_BT}$. The hotter the system, the more often the fluctuation happens. We read the gate in the same way. Its fluctuation is a false landing: a broken change that the gate lets through. The cost $\Delta$ is the number of tokens that it takes to repair that change. We take $P_{\rm false}$ to be the share of the landed changes that are broken, we set $P_{\rm false} = e^{-\beta_{\rm gate}\Delta}$, and we call $1/\beta_{\rm gate}$ the temperature of the gate. A gate that rarely lets a broken change land is cold, and a gate that often does is hot.

We chose this share because it answers the question of a developer who is about to trust a landed change: how likely is it to be broken? The answer depends on the gate, and also on how many of the attempts that reach the gate are broken. One could ask instead what share of the broken changes land, which is a different number and gives a different temperature. We use the first share everywhere in this article. The cost of a repair, $\Delta$, is counted as the cost of a run is: every output token that the run generated up to the turn of the repair.

\paragraph{Cost of a repair.} We measured $\Delta$ on the third system with 36 repair runs over ten planted defects, three runs on each of eight defects and six on each of the other two. For each run, the experiment's grader evaluated the saved trees in editing order until the first green verdict. Two controls came first: an untouched planted tree read red, and a tree with the plant reverted read green at the first turn. $\Delta$ is the median of the output tokens spent up to the turn of the first green. All 36 repairs reached green, at a median of 1,056 output tokens (90 per cent interval 913 to 1,225), and the first edit was the fix in 35 of them. Runs on one defect are not independent, so the interval is a two-stage bootstrap of 4,000 draws, which resamples the ten defects and then the runs of each drawn defect.

\paragraph{Limits of the definition.} A condition with no false landing has no temperature. A condition of $n$ landings cannot show a temperature between zero and $\Delta/\ln n$, the value that a single false landing gives.

\paragraph{Temperatures and their intervals.} The temperature of the gate was 487, 514, 778 and 830 tokens a nat at errors of 0.10, 0.15, 0.20 and 0.30, with 90 per cent intervals of 302 to 698, 335 to 731, 531 to 1,133 and 594 to 1,146. With the other share of the Definition, the broken changes that land, the temperature at an error of 0.30 would be about 1,150 tokens a nat (14 of the 35 broken changes landed) in place of 830 (14 of the 50 landed changes were broken). Each interval resamples the repair runs in the two stages above, together with the judged planted runs of the condition. A draw with no false landing has no temperature, and we counted it as zero. Such draws were under 2 per cent of the draws at the errors from 0.10 up and 37 per cent at 0.05, and for that condition we therefore give the 95th percentile of the draws as an upper limit. At 0.05 one false landing among 39 gives 288 tokens a nat, the lowest temperature that a condition of that size can show. At 0.01 one false landing among 37 gives 292 tokens a nat in the same way, and the condition at 0.02 had no false landing and therefore no temperature.

\paragraph{Comparison with the model.} We compared the gate with the cost form of the model's temperature and not with the work form, because we did not separate the tokens of a repair into work and heat. By Clausius's version of the second law (\S\ref{sec:intro}), a gate hotter than the model cannot serve as the cold side of the two-reservoir reading.

\subsubsection{Estimating the gain}

\paragraph{Definition.} The gain of serving a
verdict at error $P$ is the work a true green is worth less the work a false landing destroys, each weighted by
the injected error,
\begin{equation}
G(P) = (1-P)\,W_{\rm true}(0) - P\,D(P),
\label{eq:gain}
\end{equation}
with $W_{\rm true}(0)$ the mean number of output tokens per true landing at $P = 0$ on the toy system, taken as the worth of a green, and $D(P)$ what a false landing destroys: the mean cost of the runs that falsely landed at $P$, plus the worth of the green that each of them stands in for. Both are in tokens. The error at which $G$ crosses zero is where the gate stops paying for itself.

\paragraph{Gain at three errors.} The gain was 798, 573 and 253 tokens at errors of 0.05, 0.15 and 0.30. A straight extrapolation puts a crossing of zero near 0.42, a value we do not rely on. Weighting by $P$ makes the gain a score. We did not measure the return directly.

\paragraph{Weighting by the measured rate.} Weighting by the measured
$P_{\rm false}$ instead of $P$ lowers $G$ wherever $P_{\rm false} > P$. Re-weighting by the measured $P_{\rm false}$ in place of $P$, a sensitivity analysis on the same records (90 per cent bootstrap over runs), gives $+779$ $[+512, +1136]$, $+330$ $[-20, +714]$ and $-221$ $[-584, +133]$ tokens at the three errors. At 0.30 the point estimate is negative, with 85 per cent of draws below zero and an interval that includes zero, so the second weighting does not establish a crossing either.

\subsection{Symmetry rule and controls}\label{sec:m-symmetry}
We call a condition symmetric when the interval of its asymmetry lies inside the margin, and otherwise report whether the interval excludes zero. Each experiment carried a model-free positive
control, a planted defect the gate had to catch or a planted mutant the set had to grade red, and the free-energy script carried the planted controls described above. The model calls of the experiments of this letter cost about 515 US dollars, over 2{,}880 priced runs. This is a lower bound, because some stages recorded no cost for the runs that ended in an error.

\subsection{Data and code availability}\label{sec:m-data}
The analysis scripts, the figure scripts and the timed analysis records will be released with this article, once the text of the private repository that the records carry (names of test files, labels of edits and names of hosts) has been removed from them. The records give one row per run, with its task, direction, Hinton temperature, set, outcome and number of turns, the tokens generated at each turn and the tokens read in the run. They suffice to repeat the division into work and heat, the free energies, the dispersion of the costs, their quantiles and the test of time order. They do not hold what the map, the loops, the paths, the rates of the gate and the cost of a repair require, which are the edits, the commands, the identifiers of the trees and the verdicts of the gate. The code of the harness and the transcripts of the runs are not released, because the harness and the repository on which the agents worked belong to a private development repository. No text from them appears in this article. In follow-up work we will repeat the measurements on a public benchmark.

\section{Conclusions, outlook and practical recommendations for harness optimization}
\label{sec:conclusions}
\begin{table*}[!tp]\centering\small
\captionsetup{justification=raggedright,singlelinecheck=false}\caption{Where the tokens go: five questions and the answers of this letter. The last column gives the section of the Results that holds the evidence.}
\label{tab:findings}
\begin{tabular}{@{}>{\raggedright\arraybackslash}p{0.25\textwidth}p{0.57\textwidth}p{0.08\textwidth}@{}}
\toprule
question & answer & section \\
\midrule
\textbf{\textit{What does it cost to make a change and then undo it?}} & The cycle returns the requested edits to their start, and it still costs tokens. It is a hysteresis loop at one temperature, like that of a molecule pulled open and folded again, and all of its tokens are lost. & \S\ref{sec:cycle}, \S\ref{sec:engine} \\\addlinespace
\textbf{\textit{How much of a run is heat, and what must a change cost at the least?}} & About 80 per cent of the tokens of a landed run are heat. The free energy of a change, our estimate of the least that the change can cost, is 14 to 22 per cent of the cost of a landed coding run. Coding and uncoding are two driven processes, and their rates do not measure that free energy. & \S\ref{sec:free} \\\addlinespace
\textbf{\textit{Where do the tokens go when the agent is cold?}} & Most of them go to runs that never land, because the agent is caught in a loop. The price of a landing stays the same. The lost runs add heat and almost no free energy. & \S\ref{sec:wells}, \S\ref{sec:cold} \\\addlinespace
\textbf{\textit{How good an instrument is the test suite?}} & It is a noisy one. The number of tests failed by the suite varies from one clean run to the next, so a threshold set on one run fixes no false-alarm rate. In sixty trials the whole suite caught no more planted defects than the tests of the changed file. & \S\ref{sec:gate} \\\addlinespace
\textbf{\textit{What does an error of the gate cost?}} & More than its rate suggests. A verdict is a measurement with an error, and sound changes are refused while broken ones land. The share of false landings exceeded the error on the toy system, where more than half of the attempts were truly red. It stayed close to the error on the repository, where about half were. & \S\ref{sec:mistake} \\
\bottomrule
\end{tabular}
\end{table*}
Table~\ref{tab:findings} collects the conclusions of this letter as five questions and their answers.

In this work, we showed that the token records of a coding agent can be read as an experimentalist reads the work records of a single molecule that is pulled open and let fold again. A change to a repository has a free energy, the least that it can cost, and most of the tokens that a run spends above it are heat. Coding costs more than uncoding, and most of what an agent spends is heat. The test suite that judges a change is a measuring instrument, with noise and with errors that have a cost of their own. Thermodynamics therefore gives agentic coding a common unit, the token, and a floor against which a harness can be compared.

By building the simplest harness that we could, we allowed ourselves to explore the raw thermodynamic behaviour of agentic coding. Our analysis therefore found a behaviour that we were not looking for originally: the cost of the runs had a long tail. To treat that tail, we had to decide how to count the runs that never landed, which in turn led us to introduce the VIVA, CORTA and LOCA approximations. By then inspecting the traces of those runs, we found that cold agents repeat themselves until the limit on turns stops them (\S\ref{sec:wells}). Afterwards, by bibliographic search, we found that loops of this kind are known and that several production harnesses already guard against them~\cite{yang2024sweagent,yang2025swesmith,openhands2026stuck}. Our list of citations here is not exhaustive. Very likely, those guards were found heuristically, by humans who inspected the trajectories, found the loops and developed heuristic algorithms to stop them. The thermodynamic analysis led us to the same fault without previous knowledge of that work. In addition, with the proper statistics we can state what this fault costs in heat, which identifies it as perhaps one of the biggest issues of agentic coding. In our experiment, the limit of 300 turns stopped 78 of the 1{,}200 runs, and those 78 runs generated 66 per cent of all the tokens, most of it, if not all, heat (\S\ref{sec:m-unfinished}).

\subsection{Practical recommendations}

Our analysis suggests six practical recommendations, R1 to R6, for optimizing a harness:
\paragraph{R1. Avoid sampling the model cold.} A cold agent repeats itself, and a run caught in a loop spends tokens without changing the code. Which temperature is best we cannot say from one model and two tasks (\S\ref{sec:wells}, \S\ref{sec:cold}, \S\ref{sec:m-unfinished}).

{\bfseries\itshape What does thermodynamics say about this?}\enspace A system that sits in a metastable state leaves it only by a fluctuation, and Arrhenius and Kramers showed that the rate of escape falls exponentially as the temperature falls~\cite{arrhenius1889reaktionsgeschwindigkeit,kramers1940brownian}. The Hinton temperature sets the size of the fluctuations in what the model writes. A cold model gives its most likely reply every time, so a reply that left the repository unchanged is followed by the same reply. The agent then behaves like a liquid cooled into a glass: it has not reached its goal, and it no longer moves. We measured no barrier and no rate of escape in the agent. The analogy is with the role that temperature plays in an escape; the Hinton temperature is a setting of the sampler, and it is a different quantity from the software temperature.

\paragraph{R2. Get a stuck agent unstuck.} Detect loops, and stop or restart the run. In the theory of stochastic resetting, a restart lowers the mean time to completion when the standard deviation of that time exceeds its mean, for a process that begins afresh at each restart~\cite{luby1993optimal,evans2011diffusion,reuveni2016optimal}. Our stored runs show what a shorter limit on turns would have kept. A restart or a hint changes what the agent does next, and each needs its own experiment. Choose the limit on turns from the turn at which successful runs finish (\S\ref{sec:wells}, \S\ref{sec:cold}, \S\ref{sec:m-unfinished}).

{\bfseries\itshape What does thermodynamics say about this?}\enspace A run caught in a loop resembles a system held in a metastable state. Its tokens keep flowing and the repository does not change, so by our definition all of them are heat. Waiting for a fluctuation to free the run costs heat for as long as the wait lasts. Statistical physics offers two cheaper remedies. The first is to reset: when the time to completion varies by more than its mean, starting again shortens the mean time~\cite{evans2011diffusion,reuveni2016optimal}. The second is information. Szilard showed that one bit about the state of a system is worth work~\cite{szilard1929entropieverminderung}, and an agent in a loop receives the same output at every call, which brings it no new bit.

\paragraph{R3. Report the cost of every run as work and heat.} Include the failed runs, which are the most expensive. Heat is the largest part of the cost of a landed run, and a harness can lower it by telling the agent what it would otherwise have to discover (\S\ref{sec:defwork}, \S\ref{sec:cycle}, \S\ref{sec:discussion}).

{\bfseries\itshape What does thermodynamics say about this?}\enspace The first law divides the energy that a system receives into work and heat. Our bookkeeping divides the tokens by what a turn does: the tokens of a turn that makes the requested edits are work, and all the others are heat. The second law sets the least work that a change requires, which is its free energy, and everything spent above it is dissipated. A single total of tokens hides this division, as the fuel bill of an engine hides its efficiency. A run that fails leaves no change that is kept, so nearly all of its tokens are heat.

\paragraph{R4. Treat the test suite as an instrument.} Measure its noise on clean code and repair the tests that fail there. A larger suite is not necessarily a better gate (\S\ref{sec:gate}, \S\ref{sec:m-thermo}).

{\bfseries\itshape What does thermodynamics say about this?}\enspace A thermometer whose reading jumps from one moment to the next tells a physician little about the patient. The test suite is the instrument through which the daemon looks at each change, and Sagawa and Ueda showed that a daemon can gain no more work than its measurement gives it in information~\cite{sagawa2010generalized}. We use their result as an analogy, since we measured no work that a gate extracts. An experimentalist calibrates such an instrument before trusting it: she reads it many times on a sample whose answer she knows, and learns how much it fluctuates. For a test suite, the sample with a known answer is clean code.

\paragraph{R5. Make the gate more reliable as the task gets harder.} When most attempts fail, the broken changes that land outnumber what the gate's error rate alone would suggest (\S\ref{sec:mistake}, \S\ref{sec:m-daemon}).

{\bfseries\itshape What does thermodynamics say about this?}\enspace Clausius's version of the second law says that heat does not pass by itself from a colder body to a hotter one~\cite{clausius1854,clausius1856}. In our reading the gate is the cold side. It takes out of the repository the disorder that the agent puts in, as a refrigerator takes heat out of what it holds. A gate that errs is warmer, and a gate warmer than the agent works like a refrigerator that heats: it lets disorder in. The harder the task, the more broken attempts arrive at the gate, and the colder the gate has to be to keep them out. This is an analogy. We compared two scales in tokens, and we did not measure a flow of entropy (\S\ref{sec:mistake}).

\paragraph{R6. Benchmark a harness with a cycle of coding and uncoding.} Its start and end states are known, and every token that it spends is lost (\S\ref{sec:defcycle}, \S\ref{sec:cycle}, \S\ref{sec:engine}).

{\bfseries\itshape What does thermodynamics say about this?}\enspace A cycle returns the system to the state in which it began. Energy and free energy are functions of the state, so their change over a cycle is zero, and everything that the cycle spent has been dissipated~\cite{clausius1865}. When a molecule is pulled open and let fold again, the two curves of force against extension enclose a loop, and the area of that loop is the work that was lost~\cite{collin2005verification}. Our cycle has no such curve to draw, so we measure the same loss by subtraction: the work that coding costs, minus the work that uncoding gives back. A benchmark that scores whether a task was solved records the end state and nothing about the path. A cycle of coding and uncoding measures the dissipation of a harness directly, without an estimate of the free energy.

\subsection{Outlook}

\paragraph{One kind of change.} These recommendations rest on one kind of change. The software change that we employed is similar to the hairpin of the pulling experiments (\S\ref{sec:intro}) in the sense that the inherent structure of the code does not change very much. What we did in each task was rename a module that stays in place within the large codebase. Therefore, our system has two well-defined states, and it also does not add anything new logically to the repository.

\paragraph{Larger changes.} In future work, we will explore more complex scenarios where coding and uncoding will lead to large transformations in the source code, and then the cycle will no longer be a comparison between two similar states. We expect the free energy of such changes to be a larger share of their cost, and the faults that we will find will be harder for a test suite to detect.

\paragraph{Memory and swarms.} We are also extending this analysis to agents with memory systems and to swarms of agents that work on the same repository. A memory acts as a second daemon, which selects what the agent reads before it acts, and a swarm shares one gate among many agents. We expect the same bookkeeping of work, heat and error to apply to both. Repeating the measurements with other models and harnesses will show how general these recommendations are.

\section*{Acknowledgements}
During their adventures in vibe coding, A.A.-G. and V.B. started to discuss the thermodynamic implications of coding: in other words, how many tokens were lost. A.A.-G. was inspired by talks of Christopher Jarzynski that he had heard as a graduate student at the University of California, Berkeley, and again recently as a professor at the University of Toronto. A long period of research followed, with discussions between A.A.-G., V.B. and the swarm, about sixteen research and coding agents, each named after a famous Mexican wrestler. The agents were a mix of Claude Code, Codex and ZCode coding agents. A.A.-G. and V.B. thank Anders G. Fr\o{}seth for his generous support. The design of this paper is inspired by Latin American colours, in particular those of the architect Luis Barrag\'an. The Maxwell's daemons in this article were drawn by Codex in dialogue with the authors and Claude, in a style inspired by the cartoons of Quino, the renowned Latin American cartoonist and creator of Mafalda. The typefaces are by Latin American designers: Chivo and Archivo by H\'ector Gatti, Faustina by Alfonso Garc\'ia and Bitter by Sol Matas, released under the SIL Open Font License.

\section*{Funding}
This work was supported by the Defense Advanced Research Projects Agency (DARPA) under Agreement No.~HR0011262E022. This work was also supported by the AI2050 program of Schmidt Sciences. A.A.-G. and V.B. acknowledge the generous support of the University of Toronto's Acceleration Consortium, which receives funding from the CFREF-2022-00042 Canada First Research Excellence Fund.

\bibliographystyle{unsrt}
\bibliography{references}
\makeatletter\setlength{\@fptop}{0pt}\setlength{\@fpsep}{14pt}\setlength{\@fpbot}{0pt plus 1fil}\setlength{\@dblfptop}{0pt}\setlength{\@dblfpsep}{14pt}\setlength{\@dblfpbot}{0pt plus 1fil}\makeatother
\newpage
\section{Supplementary Information and Extended Data}\label{sec:extended}
\subsection*{Supplementary Note 1: the run-length null for the cycle temperature}
\label{app:cycle}
The cycle temperature has a null under which it reads only how the lengths of runs scatter. Let a run take $n$ turns, and let each turn cost $m$ tokens on average, with variance $\sigma_m^2$, independently of the other turns and of $n$. The cost $C$ of the run is then a sum of a random number of terms, with mean $m\langle n\rangle$ and variance $\langle n\rangle\sigma_m^2 + m^2\,\mathrm{var}(n)$. For one direction the variance of the cost over twice its mean is
\[
\frac{\sigma_C^2}{2\langle C\rangle} = \frac{m F_n}{2} + \frac{\sigma_m^2}{2m},
\qquad F_n = \frac{\mathrm{var}(n)}{\langle n\rangle}.
\]
The dispersion $F_n$ of the turn count is its variance divided by its mean, the Fano factor, which is one for a Poisson count. Equation~\ref{eq:betacyc} adds the variances of the two directions and adds their means. The cycle temperature is therefore the average of the two directions' ratios, each weighted by its direction's share of the summed mean cost:
\[
\frac{1}{\beta_{\rm cyc}} = w_F\,\frac{\sigma_{C,F}^2}{2\langle C_F\rangle} + w_R\,\frac{\sigma_{C,R}^2}{2\langle C_R\rangle},
\]
with $w_F = \langle C_F\rangle/(\langle C_F\rangle + \langle C_R\rangle)$ and $w_R = 1 - w_F$. When the turns of a run cost about the same, $\sigma_m^2 \ll m^2 F_n$, each ratio is $mF_n/2$, and the temperature reads how run lengths scatter in units of a turn. 

\subsection*{Supplementary Note 2: the survivor rate and the false-landing rate}
\label{app:mistake}
Equation~\ref{eq:survivor}: with defect rate $d$ before the gate, a defective trial lands with probability $P$ and a
clean one with probability near one, so the landed set carries $dP$ defective trials per $dP + (1-d)$ landed. With the threshold based on one clean run, the measured $P$ was 5 of 8 under either gate. With the observed $d = 0.40$ and $\alpha = 0$, the equation gives $d' = 0.29$. The observed survivor defect rates were 0.29 and 0.31. Under the calibrated thresholds the whole suite passed 6 of its 8 defects, for which the equation gives 0.33 against the observed 0.35. Equation~\ref{eq:pfalse} follows from the two ways
a green can be served; with each error probability's $s$ it predicts 0.08, 0.21 and 0.38 at $P = 0.05, 0.15, 0.30$ against the observed 0.06, 0.26 and 0.52. Each prediction lies inside the Wilson 95 per cent interval of the observed proportion: $[0.010, 0.270]$, $[0.125, 0.465]$ and $[0.340, 0.693]$. On the repository, with each condition's $s$, it predicts 0.04, 0.14, 0.17, 0.23 and 0.23 at the five errors against the observed 0.03, 0.11, 0.13, 0.26 and 0.28. Each prediction again lies inside the Wilson 95 per cent interval of the observed proportion: $[0.005, 0.132]$, $[0.045, 0.260]$, $[0.056, 0.267]$, $[0.142, 0.421]$ and $[0.175, 0.417]$. The agent received about 5.6 verdicts per run,
so flips in both directions compound, which a per-run model would carry.

\setcounter{figure}{0}\renewcommand{\figurename}{Supplementary Fig.}
\makeatletter\@ifundefined{theHfigure}{}{\renewcommand{\theHfigure}{S\arabic{figure}}}\makeatother
\begin{figure*}[!htbp]\centering\includegraphics[width=\textwidth]{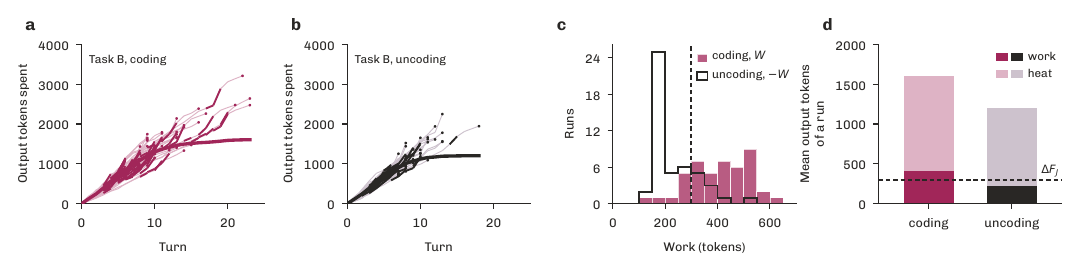}
\caption{A change made and undone on Task B at Hinton temperature 0.5, drawn as panels b to e of Fig.~\ref{fig:cycle} are for Task A. (a) The 45 landed coding runs (rose), in which the agent made the change (forward). The other three of the 48 coding runs stopped at the limit of 300 turns and are not drawn. Each thin line is one run: the output tokens it had spent after each turn, ending in a dot at the turn where it landed. A turn that changed the state $q$ is drawn dark, and its tokens are work. A turn that left $q$ unchanged is drawn pale, and its tokens are heat. The thick line is the mean over the runs, each run held at its final cost after it landed. (b) The 48 landed uncoding runs (black), in which it undid the change (reverse), on the same axes. (c) The work of the runs of a and b, in bins of 50 tokens: $W$ for coding (filled) and $-W$ for uncoding (outline). The dashed line is the free energy of the change, $\Delta F_J$, 299 tokens. (d) The mean output tokens of a run in each direction, divided into work (dark) and heat (pale). The dashed line is $\Delta F_J$.}
\label{fig:cycleB}\end{figure*}

\begin{figure*}[!htbp]\centering\includegraphics[width=0.9\textwidth]{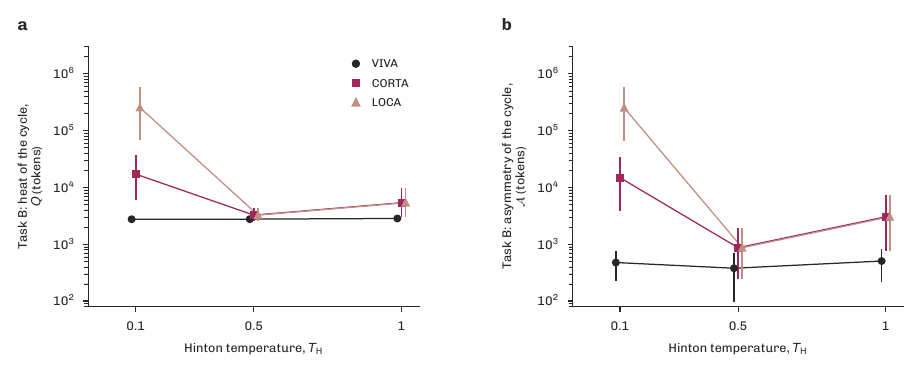}
\caption{The heat and the asymmetry of the cycle of Task B, counted in three ways, drawn as Fig.~\ref{fig:threeways} is for Task A. (a) The heat of the cycle, $Q$, against the Hinton temperature, $T_{\rm H}$, at 0.1, 0.5 and 1. (b) The asymmetry of the cycle, $\mathcal{A}$. Black circles employ the VIVA approximation (Eq.~\ref{eq:viva}), rose squares the CORTA approximation (Eq.~\ref{eq:corta}) and tan triangles the LOCA approximation (Eq.~\ref{eq:loca}). The vertical axes are logarithmic. Bars are 95 per cent bootstrap intervals over runs, and a bar that reaches the bottom of an axis extends below it. The three conditions of Task B hold 144 runs, of which 132 landed.}
\label{fig:threewaysB}\end{figure*}

\begin{figure*}[!htbp]\centering\includegraphics[width=0.48\textwidth]{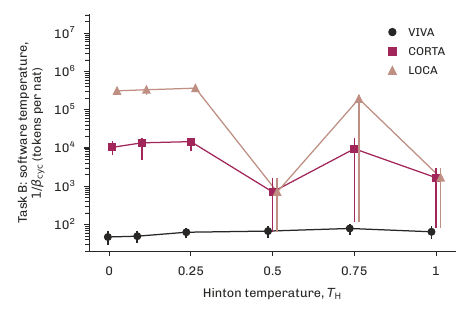}
\caption{The software temperature of the cycle, $1/\beta_{\rm cyc}$, in tokens per nat, against the Hinton temperature, $T_{\rm H}$, for Task B, drawn as Fig.~\ref{fig:softT} is for Task A: the VIVA approximation (black circles), the CORTA approximation (rose squares) and the LOCA approximation (tan triangles). The vertical axis is logarithmic. Bars are 95 per cent bootstrap intervals over runs. The six conditions of Task B hold 602 runs, of which 564 landed.}
\label{fig:softTB}\end{figure*}

\begin{table*}[!htbp]\centering\small
\caption{The cycle in each of its six conditions: two tasks at three Hinton temperatures, with 24 runs started in each direction. $Q$ is the heat of the cycle and $\mathcal{A}$ its asymmetry, both in tokens and over the landed runs. The asymmetry is given with its 97.5 per cent normal interval, and the next column says whether zero lies inside or outside that interval. $1/\beta_{\rm cyc}$ is the software temperature of the cycle, in tokens a nat, computed on the 288 runs of these six conditions; Fig.~\ref{fig:softT} computes it on the 1{,}200 runs of the twelve conditions, which include them, and its values differ accordingly. The last column gives the number of coding (forward) and of uncoding (reverse) runs that landed. A condition is marked h, for incomplete, when more than five per cent of the runs of a direction did not land.}
\label{tab:cycle}
\begin{tabular}{llrrlrl}
\toprule
task & $T_{\rm H}$ & $Q$ & $\mathcal{A}$ [97.5\%] & zero & $1/\beta_{\rm cyc}$ & landed F/R \\
\midrule
Task A &  0.1 & 4,158 & $+117$ $[-289, +523]$ & inside & 63 & 14/21 h \\
Task A &  0.5 & 3,999 & $+427$ $[-17, +870]$ & inside & 118 & 24/24 \\
Task A &  1.0 & 3,670 & $+818$ $[+367, +1268]$ & outside & 132 & 24/24 \\
Task B &  0.1 & 2,745 & $+475$ $[+191, +759]$ & outside & 56 & 18/24 h \\
Task B &  0.5 & 2,757 & $+377$ $[+53, +701]$ & outside & 85 & 22/24 h \\
Task B &  1.0 & 2,835 & $+505$ $[+178, +832]$ & outside & 80 & 20/24 h \\
\bottomrule
\end{tabular}
\end{table*}

\begin{table}[!htbp]\centering\small
\caption{Counts of distinct end trees among landed runs of Task A, by direction, at five Hinton temperatures. On Task B the counts are 2 to 4 in each direction at every temperature.}
\label{tab:trees}
\begin{tabular}{lrrrrr}
\toprule
Hinton temperature, $T_{\rm H}$ & 0.01 & 0.25 & 0.5 & 0.75 & 1 \\
\midrule
coding (forward) & 24 & 32 & 29 & 39 & 19 \\
uncoding (reverse) & 3 & 5 & 2 & 10 & 5 \\
\bottomrule
\end{tabular}
\end{table}

\begin{table*}[!htbp]\centering\small
\caption{Uncoding from the ensemble of end trees, on both tasks at $T_H=0.5$ (\S\ref{sec:endtrees}): the cost and the work of a landed uncoding run, and the slope and the zero crossing of Crooks' test, for the two starts. Intervals are 90\% bootstrap intervals. On Task A the 48 landed coding runs cost 2,305 tokens and did 415 tokens of work, $\beta$ is 0.0065 per token and the free energy is 351 tokens. On Task B the 45 landed coding runs cost 1,606 tokens and did 410 tokens of work, $\beta$ is 0.0178 per token and the free energy is 299 tokens.}
\label{tab:endtrees}
\begin{tabular}{llrrllll}
\toprule
Task & Start & Runs & Trees & Cost (tokens) & Work (tokens) & Slope (per token) & Crossing (tokens) \\
\midrule
A & single tree & 100 & 1 & 1,804 $[1{,}709, 1{,}907]$ & 317 $[297, 337]$ & 0.0032 $[0.0012, 0.0052]$ & 383 $[352, 450]$ \\
A & ensemble & 100 & 24 & 1,831 $[1{,}726, 1{,}954]$ & 397 $[374, 422]$ & 0.0007 $[-0.0012, 0.0021]$ & 350 $[159, 620]$ \\
B & single tree & 100 & 1 & 1,153 $[1{,}100, 1{,}206]$ & 212 $[200, 226]$ & 0.0221 $[0.0160, 0.0321]$ & 286 $[271, 309]$ \\
B & ensemble & 100 & 2 & 1,179 $[1{,}120, 1{,}247]$ & 222 $[212, 232]$ & 0.0190 $[0.0138, 0.0276]$ & 304 $[288, 327]$ \\
\bottomrule
\end{tabular}
\end{table*}

\begin{table*}[!htbp]\centering\small
\caption{The twelve conditions of the free-energy measurement: two tasks at six Hinton temperatures $T_{\rm H}$, 1{,}200 runs of the plain agent, of which 1{,}110 landed. The coding (forward) and uncoding (reverse) columns each give landed runs over runs started. The unfinished share is the percentage
of a direction's tokens spent in runs that did not land. Cost is the mean cost of a landed run, in tokens, and the heat
share is the percentage of the landed runs' tokens spent in turns that did not move the state. $1/\beta$ is the
temperature of the work in tokens per nat (Eq.~\ref{eq:betawork}), $\Delta F_{J}$ the free energy from the coding
runs (Eq.~\ref{eq:jarzynski}) and $\Delta F_{S}$ the half difference of the mean works (Eq.~\ref{eq:symmetric}), in
tokens; brackets are 90 per cent bootstrap intervals over runs. The last column gives the mean cost of a landed coding run divided by the mean cost of a landed uncoding run. A condition marked $\circ$ has more than 85 per cent of a direction's tokens in unfinished runs and is left out of
the ranges quoted in the text.}
\label{tab:cells}
\setlength{\tabcolsep}{5pt}
\begin{tabular}{llrrrrrrrrrr}
\toprule
 & $T_{\rm H}$ & \multicolumn{2}{c}{landed/runs} & unfinished & \multicolumn{2}{c}{cost} & heat & $1/\beta$ & $\Delta F_{J}$ & $\Delta F_{S}$ & cost \\
 & & coding & uncoding & share, \% & coding & uncoding & share, \% & & & & ratio \\
\midrule
Task A & 0.01$\,\circ$ & 29/46 & 50/53 & 91/66 & 2{,}510 & 2{,}030 & 83/85 & 118 $[76, 181]$ & 354 $[332, 375]$ & 367 & 1.24 \\
Task A & 0.1$\,\circ$ & 32/48 & 43/48 & 93/79 & 2{,}297 & 2{,}060 & 82/84 & 156 $[87, 332]$ & 368 $[346, 390]$ & 373 & 1.12 \\
Task A & 0.25 & 46/53 & 52/54 & 74/63 & 2{,}374 & 1{,}883 & 83/83 & 141 $[93, 233]$ & 360 $[343, 379]$ & 362 & 1.26 \\
Task A & 0.5 & 48/48 & 48/48 & 0/0 & 2{,}305 & 1{,}710 & 82/83 & 153 $[108, 234]$ & 351 $[330, 371]$ & 352 & 1.35 \\
Task A & 0.75 & 51/53 & 51/51 & 22/0 & 1{,}986 & 1{,}580 & 80/82 & 147 $[99, 228]$ & 342 $[324, 361]$ & 341 & 1.26 \\
Task A & 1 & 48/48 & 48/48 & 0/0 & 2{,}249 & 1{,}551 & 80/81 & 128 $[82, 203]$ & 368 $[340, 399]$ & 367 & 1.45 \\
Task B & 0.01$\,\circ$ & 36/48 & 47/47 & 87/0 & 1{,}732 & 1{,}312 & 75/85 & 24 $[15, 35]$ & 260 $[240, 348]$ & 311 & 1.32 \\
Task B & 0.1 & 42/48 & 48/48 & 79/0 & 1{,}650 & 1{,}171 & 76/83 & 49 $[32, 70]$ & 275 $[253, 298]$ & 295 & 1.41 \\
Task B & 0.25 & 50/58 & 52/52 & 83/0 & 1{,}534 & 1{,}250 & 75/84 & 53 $[36, 73]$ & 271 $[251, 293]$ & 294 & 1.23 \\
Task B & 0.5 & 45/48 & 48/48 & 26/0 & 1{,}606 & 1{,}205 & 74/81 & 56 $[38, 79]$ & 299 $[271, 342]$ & 319 & 1.33 \\
Task B & 0.75 & 49/53 & 56/56 & 59/0 & 1{,}502 & 1{,}193 & 73/78 & 90 $[58, 141]$ & 337 $[307, 364]$ & 336 & 1.26 \\
Task B & 1 & 43/48 & 48/48 & 44/0 & 1{,}559 & 1{,}201 & 73/79 & 70 $[47, 101]$ & 339 $[318, 362]$ & 340 & 1.30 \\
\bottomrule
\end{tabular}
\end{table*}

\begin{table*}[!htbp]\centering\small
\caption{The sixty gate trials by gate. Planted is the number of trials that carried a planted defect, caught the number of those the gate refused, and clean passed the trials without a defect that the gate let land. The share is the landed trials that carried a defect over all landed trials. The whole suite is judged twice on the same 19 trials: under a threshold of three times one clean reading, and under the calibrated threshold (\S\ref{sec:m-thermo}).}
\label{tab:gates}
\begin{tabular}{lrrrrrr}
\toprule
gate & trials & planted & caught & clean passed & landed with a defect & share \\
\midrule
none & 20 & 8 & 0 & 12 of 12 & 8 of 20 & 0.40 \\
the file's own tests & 21 & 8 & 3 & 12 of 13 & 5 of 17 & 0.29 \\
whole suite, one reading & 19 & 8 & 3 & 11 of 11 & 5 of 16 & 0.31 \\
whole suite, calibrated & 19 & 8 & 2 & 11 of 11 & 6 of 17 & 0.35 \\
\bottomrule
\end{tabular}
\end{table*}

\begin{table}[!htbp]\centering\small
\caption{True and served final verdicts of the judged planted runs on the repository, at each injected error
probability $P$. A false landing is a run whose final grade was served green while its true verdict was red. $s$ is
the share of judged runs that were truly red, and the false-landing rate is the false landings over all runs served
green.}
\label{tab:verdicts}
\begin{tabular}{lrrrrrrr}
\toprule
 & & \multicolumn{2}{c}{truly green} & \multicolumn{2}{c}{truly red} & & false-landing \\
$P$ & judged & served green & served red & served green & served red & $s$ & rate \\
\midrule
0.01 & 76 & 36 & 1 & 1 & 38 & 0.51 & 1/37 \\
0.02 & 76 & 34 & 0 & 0 & 42 & 0.55 & 0/34 \\
0.05 & 75 & 38 & 3 & 1 & 33 & 0.45 & 1/39 \\
0.10 & 79 & 31 & 1 & 4 & 43 & 0.60 & 4/35 \\
0.15 & 81 & 34 & 3 & 5 & 39 & 0.54 & 5/39 \\
0.20 & 79 & 26 & 10 & 9 & 34 & 0.54 & 9/35 \\
0.30 & 84 & 36 & 13 & 14 & 21 & 0.42 & 14/50 \\
\bottomrule
\end{tabular}
\end{table}

\begin{table*}[!htbp]\centering\small
\caption{The intervals of this article: the quantity, where it is reported, the kind of interval, its level and the unit that was resampled. A bootstrap is a percentile bootstrap, and its draws are in the last column. Counts and shares without bars (Figs.~\ref{fig:cycle}, \ref{fig:flow} and \ref{fig:wells}; Tables~\ref{tab:trees}, \ref{tab:gates} and \ref{tab:verdicts}) carry no interval.}
\label{tab:intervals}
\begin{tabular}{p{4.3cm}p{3.0cm}p{3.2cm}rp{3.4cm}}
\toprule
quantity & reported in & kind & level & resampled unit, draws \\
\midrule
asymmetry $\mathcal{A}$ & Table~\ref{tab:cycle} & normal, unequal variances & 97.5\% & none \\
mean cost of a landed run, temperature of the work $1/\beta$, free energies $\Delta F_J$ and $\Delta F_S$ & Fig.~\ref{fig:tokens}b, d, Table~\ref{tab:cells} & bootstrap & 90\% & runs, each direction on its own; 2{,}000 \\
cost of one landing, heat, asymmetry and software temperature of the cycle, under the three approximations & Figs.~\ref{fig:threeways}, \ref{fig:softT} and \ref{fig:tokens}a, c; Supplementary Figs.~\ref{fig:threewaysB} and \ref{fig:softTB} & bootstrap & 95\% & runs; 2{,}000 \\
crossing of the two work distributions & \S\ref{sec:free} & bootstrap & 90\% & runs, each direction on its own; 1{,}000 \\
trapped share and landing rates & Fig.~\ref{fig:trapped} & bootstrap & 90\% & runs of a condition; 4{,}000 \\
bound on the relative entropy of a run and its reverse & \S\ref{sec:engine} & bootstrap & 95\% & runs kept out of the fit \\
false-landing rate, toy system & Fig.~\ref{fig:mistake}b, Supplementary Note 2 & Wilson & 95\% & none \\
false-landing rate, repository & \S\ref{sec:mistake}, Supplementary Note 2 & Wilson & 90\% and 95\% & none \\
slope of the false-landing rate against the error & \S\ref{sec:mistake} & bootstrap & 90\% & judged runs within conditions; 4{,}000 \\
repair cost $\Delta$ & \S\ref{sec:mistake} & two-stage bootstrap & 90\% & planted defects, then runs of each defect; 4{,}000 \\
gate temperature & \S\ref{sec:mistake} & two-stage bootstrap & 90\% & repair runs as for $\Delta$, with the judged runs of the condition; 4{,}000 \\
gain re-weighted by the false-landing rate & \S\ref{sec:mistake} & bootstrap & 90\% & runs \\
correlation of defect and gate among the landed trials & \S\ref{sec:m-thermo} & bootstrap & 90\% & trials \\
upper limit of the gate temperature at an error of 0.05 & \S\ref{sec:m-daemon} & 95th percentile of the draws of the two-stage bootstrap & 95\%, one-sided & as for the gate temperature \\
\bottomrule
\end{tabular}
\end{table*}

\end{document}